\nonstopmode

\documentclass[10pt,letterpaper]{article}
\usepackage{amsthm, amssymb, amsmath, bm, bbm, mathtools, physics, dsfont}
\usepackage[scr=boondoxo]{mathalfa}
\usepackage[usenames,dvipsnames,svgnames,table]{xcolor}
\usepackage[pdftex, xetex]{graphicx}
\usepackage{float, colortbl, tabularx, longtable, multirow, subcaption, environ, wrapfig, textcomp, nicefrac,booktabs,colortbl,pdfpages,fontawesome,soul,comment, pgf, tikz,framed,tikz-cd,adjustbox,blindtext,setspace,multicol}
\usepackage{caption}
\usepackage[normalem]{ulem}
\usepackage[shortlabels,inline]{enumitem}
\usetikzlibrary{arrows,positioning,automata,shadows,fit,shapes,cd}
\usepackage{anyfontsize}

\usepackage[T1]{fontenc}
\usepackage{textcomp}
\usepackage[USenglish]{babel}
\usepackage[final]{microtype}

\usepackage[allcolors=Cerulean,colorlinks=true]{hyperref}
\usepackage{bookmark}
\usepackage{url}
\usepackage{tablefootnote}
\usepackage{makecell}
\usepackage[perpage]{footmisc}

\definecolor{myred}{RGB}{153,51,51}
\definecolor{mygray}{RGB}{102,102,102}
\definecolor{darkgreen}{RGB}{60,160,60}
\definecolor{lightblue}{RGB}{60,60,200}
\definecolor{darkred}{RGB}{70,10,10}
\definecolor{darkblue}{RGB}{1,1,255}
\definecolor{yellow}{RGB}{255,255,0}
\definecolor{purple}{RGB}{127,0,255}
\definecolor{dark}{RGB}{1,1,1}
\definecolor{lightgray}{RGB}{230,230,230}
\definecolor{ucla_gold}{RGB}{255,232,0}
\definecolor{ucla_blue}{RGB}{50,132,191}
\definecolor{lightyellow}{rgb}{1.0, 1.0, 0.88}

\usepackage[capitalize,nameinlink]{cleveref}
\crefformat{section}{Sec.\@ #2#1#3}
\crefmultiformat{section}{Sec.\@ #2#1#3}{ and~#2#1#3}{, #2#1#3}{, and~#2#1#3}
\crefformat{equation}{(#2#1#3)}
\renewcommand{\Cref}[1]{\cref{#1}}

\newcommand{\aed}[1]{\begin{aligned} #1 \end{aligned}}
\newcommand{\beq}[1]{\begin{equation}#1\end{equation}}

\newcommand{\inner}[2]{\left\langle #1,#2 \right\rangle}
\newcommand{\rbr}[1]{\left(#1\right)}
\newcommand{\sbr}[1]{\left[#1\right]}
\newcommand{\cbr}[1]{\left\{#1\right\}}

\providecommand\f[2]{\ensuremath \frac{#1}{#2}}

\DeclareMathOperator*{\argmin}{\text{argmin}}
\DeclareMathOperator*{\argmax}{\text{argmax}}

\DeclareMathOperator*{\E}{\mathbb{E}}

\newtheoremstyle{mystyle}%                % Name
  {}%                                     % Space above
  {}%                                     % Space below
  {\upshape}%                                     % Body font
  {}%                                     % Indent amount
  {\bfseries}%                            % Theorem head font
  {.}%                                    % Punctuation after theorem head
  { }%                                    % Space after theorem head, ' ', or \newline
  {\thmname{#1}\thmnumber{ #2}\thmnote{ (#3)}}%        v

\theoremstyle{mystyle}

\theoremstyle{mystyle}

\makeatletter

\makeatletter

\definecolor{color_skyblue}{rgb}{0.01,0.39,0.75}

\def \th {\theta}
\def \a {\alpha}

\def \l {\lambda}

\def \Om {\Omega}

\def \FF {\mathcal{F}}

\def \reals {\mathbb{R}}

\usepackage{chngpage,tablefootnote,authblk,lineno}
\usepackage[title]{appendix}

\usepackage[margin=1.25in]{geometry}
\usepackage[small,compact]{titlesec}

\usepackage{newtxtext}

\usepackage[autocite=superscript, backend=bibtex, sortcites=true, sorting=none, bibstyle=nature, citestyle=numeric-comp, maxbibnames=15]{biblatex}
\renewbibmacro{in:}{}

\makeatletter
\newcommand{\printfnsymbol}[1]{%
  \textsuperscript{\@fnsymbol{#1}}%
}
\makeatother

\usepackage[color=gray!30,textsize=tiny]{todonotes}

\date{}

\graphicspath{{./figs/}}%,{./old_figs/}}

\begin{document}
%\listoftodos
%\clearpage

\title{\Large
\textbf{Adapting Machine Learning Diagnostic Models to New Populations Using a Small Amount of Data: Results from Clinical Neuroscience}
}
\author[1,2,3]{\normalsize Rongguang Wang}
\author[2,3]{Guray Erus}
\author[1,4]{Pratik Chaudhari\printfnsymbol{1}\printfnsymbol{2}}
\author[1,2,3,5]{Christos Davatzikos\footnote{Corresponding authors. Email: \href{mailto:pratikac@seas.upenn.edu}{pratikac@seas.upenn.edu}; \href{mailto:christos.davatzikos@pennmedicine.upenn.edu}{christos.davatzikos@pennmedicine.upenn.edu}. Address: 200 S 33rd Street, Department of Electrical and Systems Engineering, University of Pennsylvania, Philadelphia, PA 19104, USA; 3700 Hamilton Walk, 7th Floor, Center for Biomedical Image Computing and Analytics, University of Pennsylvania, Philadelphia, PA 19104, USA.}
\thanks{These authors contributed equally to this work.}
}
\author[ ]{\footnote{For the iSTAGING~\autocite{habes2021brain}{} and PHENOM~\autocite{chand2020two}{} consortia, and for the ADNI~\autocite{jack2008alzheimer}{}.}}
\affil[1]{\small Department of Electrical and Systems Engineering, University of Pennsylvania}
\affil[2]{Center for AI and Data Science for Integrated Diagnostics, University of Pennsylvania}
\affil[3]{Center for Biomedical Image Computing and Analytics, University of Pennsylvania}
\affil[4]{Department of Computer and Information Science, University of Pennsylvania}
\affil[5]{Department of Radiology, Perelman School of Medicine, University of Pennsylvania}
\maketitle

% !TEX root = ../main.tex

\begin{abstract}

Machine learning (ML) is revolutionizing many areas of engineering and science, including healthcare. However, it is also facing a reproducibility crisis, especially in healthcare. ML models that are carefully constructed from and evaluated on data from one part of the population may not generalize well on data from a different population group, or acquisition instrument settings and acquisition protocols. We tackle this problem in the context of neuroimaging of Alzheimer's disease (AD), schizophrenia (SZ) and brain aging. We develop a weighted empirical risk minimization approach that optimally combines data from a source group, e.g., subjects are stratified by attributes such as sex, age group, race and clinical cohort to make predictions on a target group, e.g., other sex, age group, etc. using a small fraction (10\%) of data from the target group. We apply this method to multi-source data of 15,363 individuals from 20 neuroimaging studies to build ML models for diagnosis of AD and SZ, and estimation of brain age. We found that this approach achieves substantially better accuracy than existing domain adaptation techniques: it obtains area under curve greater than 0.95 for AD classification, area under curve greater than 0.7 for SZ classification and mean absolute error less than 5 years for brain age prediction on all target groups, achieving robustness to variations of scanners, protocols, and demographic or clinical characteristics. In some cases, it is even better than training on all data from the target group, because it leverages the diversity and size of a larger training set. We also demonstrate the utility of our models for prognostic tasks such as predicting disease progression in individuals with mild cognitive impairment. Critically, our brain age prediction models lead to new clinical insights regarding correlations with neurophysiological tests. In summary, we present a relatively simple methodology, along with ample experimental evidence, supporting the good generalization of ML models to new datasets and patient cohorts.

\end{abstract}
% !TEX root = ../main.tex

\section{Introduction}

Machine learning has been adopted by the medical imaging research community for building diagnostic models of neurological disorders such as Alzheimer's disease~\autocite{qiu2022multimodal, wang2023applications}{}, schizophrenia~\autocite{rozycki2018multisite}{}, major depression~\autocite{gao2018machine}{}, autism~\autocite{heinsfeld2018identification}, as also for other applications such as brain cancer prognostication~\autocite{macyszyn2015imaging} and brain age estimation~\autocite{dinsdale2021learning, leonardsen2022deep}{}. These models are very promising. They achieve, or even surpass, expert human-level accuracy in some cases~\autocite{rajpurkar2022ai}{}. They can enable prognostication, precision diagnostics and disease subtyping, and guidance for therapy in situations when it may not be feasible to obtain human expert interpretations. These models are also relatively inexpensive to deploy compared to human clinicians; this can help reduce the wide healthcare disparities prevalent in our society today.

%\newtext{
These promises of machine learning-based diagnosis are tempered, however, by the difficulty of building models that can work robustly and accurately across broad populations. The key reason for this inadequacy is that clinical data is highly heterogeneous. For neurological disorders such as Alzheimer’s disease, the heterogeneity stems not only from diverse anatomies, overlapping clinical phenotypes, or genomic traits of different subjects, but also from operational, demographic and social aspects such as data acquisition devices and protocols of different hospitals~\autocite{pomponio2020harmonization, rajpurkar2023current}{}, and paucity of data for minorities~\autocite{davatzikos2019machine}{}. As a consequence, machine learning models often exhibit poor reproducibility~\autocite{geirhos2020shortcut,ghaffari2022adversarial,larrazabal2020gender, degrave2021ai, kanakasabapathy2021adaptive, howard2021impact, benkarim2022population,li2022cross,seyyed2021underdiagnosis,daneshjou2022disparities}{}. %Machine learning is fundamentally about identifying salient features from data. But the one-fits-all paradigm that it employs works poorly for problems with heterogeneous data~\autocite{davatzikos2019machine, kopal2023end}{}.
%}

%\newtext{
There are a number of approaches in the machine learning literature to tackle the above issue. Each of these approaches entails a specific trade-off, e.g., learning representations that are invariant to the variability in the data does lead to robust predictions across the population but it can have poor accuracy~\autocite{moyer2021harmonization}{}. In simple words, the best solution on average for the population is best for none. Techniques like transfer, meta or self-supervised learning can be used to building representations that can be adapted to new domains. But, in practice, these techniques are often ineffective because the adaptation procedure for complicated neural architectures and hyper-parameter regimes is very fragile~\autocite{dhillon2019a,li2019rethinking, greene2022brain, marek2022reproducible, dhamala2023one}{}. In such approaches, the inadequacy of the one-fits-all~\autocite{davatzikos2019machine} model pre-trained on all data is traded off against the risk of the variability of the adaptation procedure using (often) samples from the new domain. Both these kinds of trade-offs are fundamental and there is no way to obviate them in general. As a consequence, it is difficult for even an expert practitioner to select an approach that is ideal for a specific problem.
%}

%\newtext{
We take a first-principles perspective on addressing heterogeneity.
A machine learning model is a statistic of the training data, the model predicts accurately if this is a sufficient statistic for the task.
By the data processing inequality, any statistic discards some information about the data, e.g., a representation invariant to age loses information about pathology induced atrophy because of correlations in age and atrophy.
If the statistic obtained from the source data is not sufficient for the target data, then we cannot predict well on the target data. For very young or very old subjects in the target group, the statistic that is invariant to age built using the source group would work well. But for target subjects that are in the intermediate ages, such a statistic has to lose some precision.
We formalize this argument mathematically in the Methods section. \footnote{As a precursor: given a dataset $D$, a model $f(x; D)$ makes predictions on the test datum $x$ using a learned statistic $f$. Formally, this is equivalent to $\argmin_{f \in \FF} f(x; D)$ where we search over some set of functions $\FF$ before making every inference. The former leads to efficient inference. But the latter we argue is more effective for addressing heterogeneity in the data.}
Therefore, the key technical idea of this paper is as follows. Instead of summarizing the source data as a statistic that may or may not be ideal for the target, we could use the source data directly, together with the target data. Predictions on the target data via this approach have to be at least as good as those obtained by adapting a source-trained model to the target data.
%}

We demonstrate this approach for domain adaptation in neuroimaging data in the context of three applications: diagnosis of Alzheimer’s disease (AD) and schizophrenia (SZ), and prediction of brain age, which is a frequently used marker for general brain health. We use multi-source data (magnetic resonance imaging features, demographic, clinical variables, genetic factors, and cognitive scores) of 15,363 individuals from several large-scale multi-modal neuroimaging datasets.  We focused on building automated diagnostic models that are trained using data from a source group, e.g., subjects stratified into groups based on attributes such as sex, age group, race and clinical cohort, and ensuring that these models can make accurate predictions on another target group, e.g., other sex, age group etc. 
%\newtext{
As a baseline, we first show that ensembles built using bagging, boosting and stacking and trained using careful data pre-processing, hyper-parameter tuning and model selection, can predict accurately on target groups, which is significantly better than deep neural networks trained on the same data; these results are also substantially better than existing domain generalization techniques such as IRM~\autocite{arjovsky2019invariant}{}, DANN~\autocite{ganin2016domain}{}, JAN~\autocite{long2017deep}{}, JDOT~\autocite{courty2017joint}{}, TENT~\autocite{wang2020tent}{}, SHOT~\autocite{liang2020we}{}, DALN~\autocite{chen2022reusing}{}, and TAST~\autocite{jang2022test}{}. Then, we develop a theoretical argument for a weighted empirical risk minimization (ERM) objective that optimally combines all data from the source group and a small fraction of the data from the target group (in practice, as small as 10\%), to further improve the predictive ability on target groups. This weighted-ERM approach can be thought of as the simplest possible instantiation of few-shot learning, i.e., adapting a model trained on a source group to a new group using few samples. Surprisingly, the performance of this approach is often comparable to, and in many cases it is even better than, that of training on \emph{all} data from the target group.
%}
Overall, on all target groups, our approach obtains area under curve greater than 0.95 for AD classification, area under curve greater than 0.7 for SZ classification and mean absolute error less than 5 years for brain age prediction, while offering brain age residuals that carry clinical value.

We further show that our diagnostic models can be used for predictive tasks, e.g., our models trained only for AD classification can be used to predict which subjects with mild cognitive impairment will progress to AD, remain stable, or revert to cognitively normal, in future follow-ups. Improvements in AD classification performance on the target group using a small amount of data also result in improvements in performance on these predictive tasks. In part due to the improved predictive ability of our brain age model, our results offer new evidence about the correlation between the brain age residual and neurophysiological tests. In particular, we found stronger correlations of brain age residuals with mini-mental state examination, digital symbol substitution test, digit span forward/backward, and trial making tests than those reported in the literature. In contrast to results in existing literature, we found that correlations between brain age residuals, cardiovascular features and lifestyle factors such as systolic and diastolic blood pressure, and body mass index were not statistically significant, potentially reflecting the fact that in this study we did not use brain measurements that more directly reflect effects of cardiovascular risk factors, such as infarcts and microbleeds.

Our results suggest that one can build machine learning models that are able to  effectively ameliorate the heterogeneity of neuroimaging data and also provide accurate predictions on diverse groups of populations, even if some of them are very under-represented (data disadvantaged). Furthermore, this might indicate that diagnostic models of neurological disorders may be ready to be evaluated in clinical settings---after necessary checks and balances are met.
% !TEX root = ../main.tex

\section{Results}
\label{s:results}

\subsection{Quantifying the differences in the distribution of data from different groups}

\begin{table}
\caption{\textbf{Summary of the data from the iSTAGING consortium (Alzheimer's disease and brain age) and the PHENOM consortium (schizophrenia) used in this study.}
}
\label{tab:data}
\begin{center}
\begin{footnotesize}

\resizebox{0.7\linewidth}{!}{
\begin{tabular}{p{0.15\linewidth}rrrrrrr}
\toprule
\textbf{Alzheimer’s} && ADNI-1 & ADNI-2/3 & PENN & AIBL & OASIS & Total \\
\textbf{Disease} && (17.37\%) & (23.29\%) & (25.44\%) & (10.07\%) & (23.82\%) \\
\cmidrule{3-8}
% & Subjects & 364 & 488 & 533 & 211 & 499 & 2095  \\
Subjects\\
& Control & 173 & 261 & 228 & 119 & 276 & 1057  \\
& Patient & 191 & 227 & 305 & 92 & 223 & 1038  \\
Sex (\%)\\
& Female & 8.31 & 11.93 & 16.28 & 6.16 & 12.70 & 55.37 \\
& Male & 9.07 & 11.36 & 9.16 & 3.91 & 11.12 & 44.63 \\
Age (\%, years)\\
& 0--65 & 1.15 & 2.67 & 4.82 & 1.38 & 4.92 & 14.94 \\
& 65--70 & 1.48 & 6.21 & 5.35 & 1.96 & 5.11 & 20.10 \\
& 70--75 & 5.54 & 5.30 & 5.20 & 2.86 & 4.53 & 23.44 \\
& 75--80 & 4.96 & 5.25 & 5.01 & 1.96 & 4.82 & 22.00 \\
& > 80 & 4.25 & 3.87 & 5.06 & 1.91 & 4.44 & 19.52 \\
Race (\%)\\
& White & 16.13 & 12.94 & 19.14 & 5.58 & 19.57 & 73.37 \\
& Black & 0.91 & 0.67 & 5.11 & - & 4.06 & 10.74 \\
& Asian & 0.24 & 0.43 & 0.48 & - & 0.19 & 1.34 \\
\bottomrule\\
\end{tabular}
}

\vspace*{2em}
\resizebox{0.75\linewidth}{!}{
\begin{tabular}{p{0.15\linewidth} rrrrrrr}
\toprule
\textbf{Schizophrenia} &  & Penn & China & Munich & Utrecht & Melbourne & Total \\
& & (22.28\%) & (13.94\%) & (29.64\%) & (20.12\%) & (14.03\%)\\
\cmidrule{3-8}
Subjects\\
%& Subjects & 227 & 142 & 302 & 205 & 143 & 1019 \\
& Control & 131 & 76 & 157 & 115 & 84 & 563 \\
& Patient & 96 & 66 & 145 & 90 & 59 & 456 \\
Sex (\%)\\
& Female & 11.87 & 6.77 & 7.75 & 6.97 & 4.02 & 37.39 \\
& Male & 10.40 & 7.16 & 21.88 & 13.15 & 10.01 & 62.61 \\
Age (\%, years)\\
& 0--25 & 5.79 & 4.91 & 9.42 & 10.11 & 5.89 & 36.11 \\
& 25--30 & 6.28 & 2.36 & 7.26 & 4.12 & 2.16 & 22.18 \\
& 30--35 & 3.53 & 2.16 & 5.99 & 2.85 & 1.37 & 15.90 \\
& > 35 & 6.67 & 4.51 & 6.97 & 3.04 & 4.61 & 25.81 \\
Race (\%)\\
& Native & 10.50 & - & - & - & - & 10.50 \\
& Asian & 7.36 & - & - & - & - & 7.36 \\
\bottomrule\\
\end{tabular}
}

\vspace*{2em}
\resizebox{\linewidth}{!}{
\begin{tabular}{p{0.13\linewidth}rrrrrrrrrrrr}
\toprule
\textbf{Brain Age} && BIOCARD & BLSA-1.5T & BLSA-3T & CARDIA & SHIP & SPRINT & UKBB & WHIMS & WRAP & lookAHEAD & Total \\
&& (2.12\%) & (1.05\%) & (7.82\%) & (5.97\%) & (28.54\%) & (5.54\%) & (34.79\%) & (9.40\%) & (2.09\%) & (2.68\%) \\
\cmidrule{3-13}
Subjects\\
& Control & 246 & 122 & 907 & 693 & 3311 &  643 & 4036 & 1090 & 242 & 311 & 11601 \\
Sex (\%)\\
& Female & 1.31 & 0.46 & 4.37 & 2.97 & 14.63 & 1.96 & 18.12 & 9.40 & 1.46 & 1.91 & 56.56 \\
& Male & 0.81 & 0.59 & 3.45 & 3.01 & 13.91 & 3.59 & 16.67 & - & 0.63 & 0.78 & 43.44 \\
Age (years)\\
& Mean & 57.61 & 67.87 & 64.06 & 51.00 & 52.81 & 68.58 & 62.81 & 69.59 & 63.57 & 58.05 & 60.14 \\
& Min & 21 & 48 & 22 & 42 & 21 & 50 & 45 & 64 & 50 & 44 & 21 \\
& Max & 86 & 85 & 92 & 61 & 90 & 91 & 79 & 79 & 78 & 74 & 92 \\
Race (\%)\\
& White & 2.09 & 0.96 & 5.15 & 3.35 & - & 3.62 & 33.79 & 8.63 & 1.99 & 1.96 & 61.54 \\
& Black & 0.02 & 0.09 & 1.97 & 2.62 & - & 1.74 & 0.21 & 0.40 & 0.04 & 0.60 & 7.69 \\
& Asian & - & - & 0.49 & - & - & 0.05 & 0.41 & 0.13 & 0.01 & - & 1.09 \\
\bottomrule\\
\end{tabular}
}
\end{footnotesize}
\end{center}
\end{table}

We focused our analysis on groups stratified by four attributes: sex, age-group, race and clinical study, for three different applications: classification of Alzheimer’s disease, classification of schizophrenia, and prediction of brain age. For all three problems, we used multi-source data including imaging measures, demographic and clinical variables, genetic factors, and cognitive scores. \cref{tab:data} summarizes the data used in our study. For a number of groups, there is a large imbalance in the number of subjects, e.g., in the iSTAGING consortium for AD, 73.37\% of the subjects are European Americans (White), 10.74\% African Americans (Black), and only 1.34\% are Asians; for the PHENOM dataset for SZ, 37.4\% participants are Female while the rest (62.6\%) are Male. Before we develop methods to adapt machine learning-based diagnostic models to subjects from different groups, in order to understand the data better, we developed a procedure based on a two-sample test to quantify the distribution shift between data from different groups. This involves building a multi-layer perceptron (MLP)-based model that classifies subjects as belonging to different groups corresponding to each of the four attributes (e.g., sex: male vs.\@ female, age: < 65 vs.\@ 65--70 vs.\@ etc., race: White vs.\@ Black vs.\@ Asian, and clinical study: ADNI-1 vs. ADNI-2/3 vs. PENN vs. etc.). The empirical two-sample test statistic on these learned features can be used to quantify the differences in the data distribution across groups; we use the maximum-mean discrepancy (MMD)~\autocite{gretton2012kernel} and elaborate upon it in~\cref{s:two_sample_test}. This procedure is conducted independently for each of the three applications.

\begin{figure}[htpb]
\centering
\begin{subfigure}[b]{\linewidth}
\centering
\includegraphics[width=\linewidth]{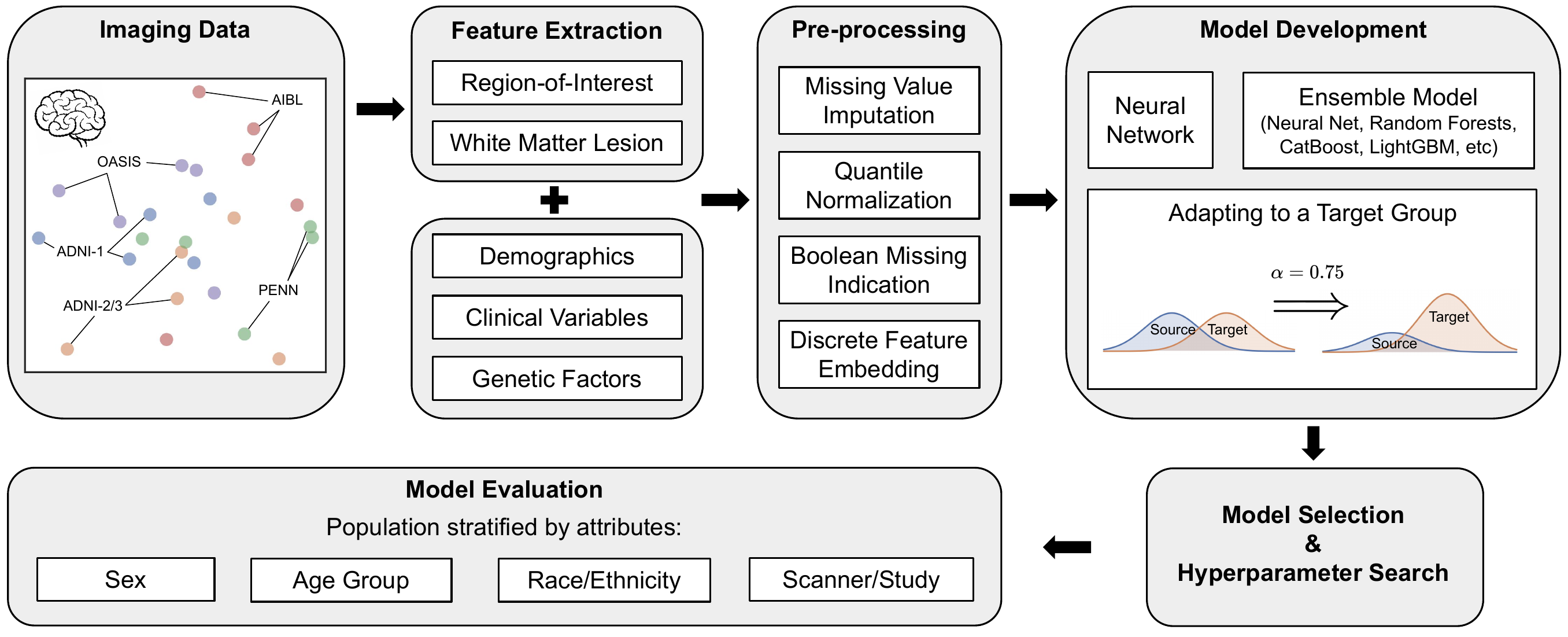}
\caption{}
\label{fig:overview}
\end{subfigure}
\begin{subfigure}[b]{0.3\linewidth}
\centering
\includegraphics[width=0.9\linewidth]{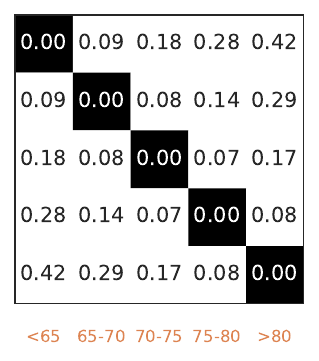} % dendrogram_ad
\caption{}
\label{fig:dendrogram_ad}
\end{subfigure}
\begin{subfigure}[b]{0.69\linewidth}
\centering
\includegraphics[width=0.49\linewidth]{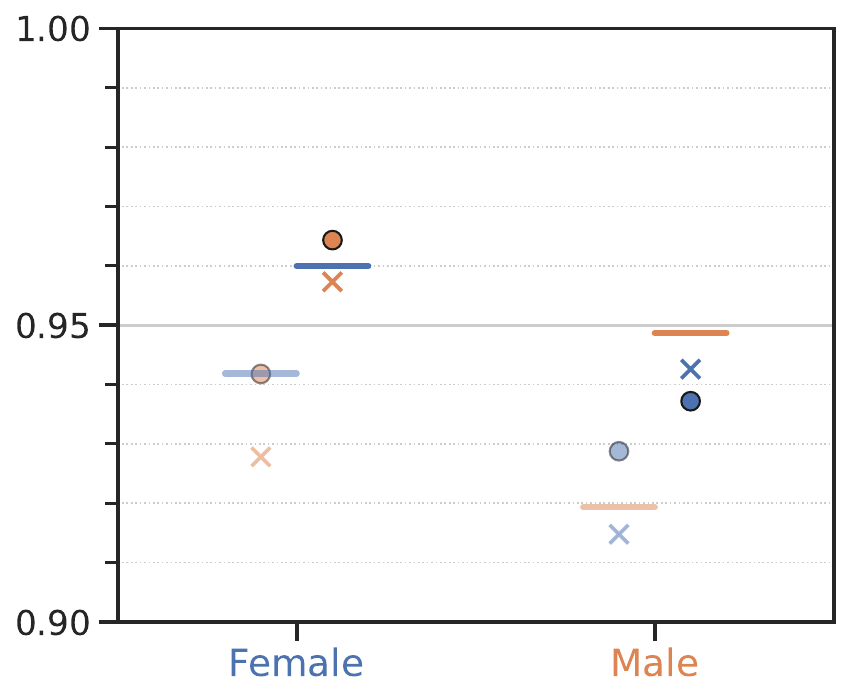}
\includegraphics[width=0.49\linewidth]{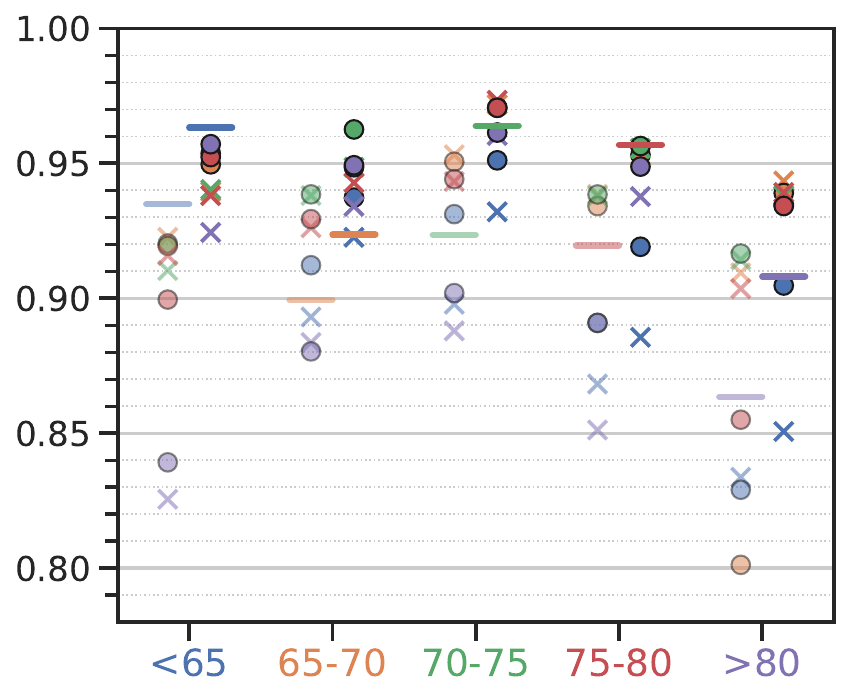}\\
\includegraphics[width=0.95\linewidth]{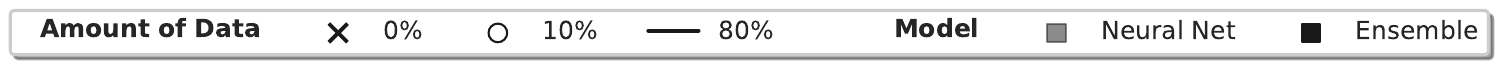}
\caption{}
\label{fig:ad_nn}
\end{subfigure}
\caption{
\textbf{Automated and robust diagnosis of neurological disorders using machine learning models.}
\textbf{(a)} A schematic of the framework for data pre-processing, model development, optimization, and evaluation employed in this paper to build machine learning models that can predict accurately on different groups for heterogeneous neurological disorders using MR images, demographic and clinical variables, genetic factors, and cognitive scores.
\textbf{(b)} %Distance between leaves of this dendrogram indicates the 
Pairwise MMD statistic between learned features of pairs of groups, e.g., distributional discrepancy between Male-Female groups is 0.17, while the distributional discrepancy between < 65 years and > 80 years, or between ADNI-1 and ADNI-2/3, is larger (0.42 and 0.26 respectively). See~\cref{s:two_sample_test} for details of the MMD calculation. \cref{fig:app:distance_ad,fig:app:dendrogram_ad} provide more details of the numerical statistics.
\textbf{(c)} Average AUC of Alzheimer’s disease classification for sex and age attributes computed using five-fold nested cross-validation; see~\cref{fig:app:dendrogram_nn_ensemble_ad} for other attributes. For both sex and age, we trained machine learning models, a deep neural network (translucent markers) and an ensemble using boosting, bagging and stacking (bold markers), using data from different source groups (different colors) and evaluated this model (cross marks) on data from different target groups (X-axis); circles denote model fitted using our $\a$-weighted ERM procedure with access to 10\% data from the target group; horizontal lines denote models that are directly trained on the target group using 80\% of data (the rest for testing). All models use data from multiple sources, namely structural measures, demographic, clinical variables, genetic factors, and cognitive scores. In general, (i) the AUC of ensemble models is higher than that of the neural network in all cases ($p$ < 0.01), (ii) AUC of a model trained on a source group remains remarkably high when evaluated on the target group (crosses), (iii) in most cases, it further improves when one has access to a small fraction of data from the target group (circles are higher than crosses), and (iv) often times even beyond the AUC of a model directly trained on the target group (circles above the horizontal lines).
}
\label{fig:dendrogram_nn_ensemble_ad}
\end{figure}

In~\cref{fig:dendrogram_ad}, for each attribute, we have shown a dendrogram obtained using the pairwise maximum-mean discrepancy (MMD) statistic~\autocite{gretton2012kernel} across different groups.  For all attributes, the hypothesis that different groups have the same distribution does not hold ($p< 10^{-4}$). In our visualization, the angular distance between groups is proportional to the MMD statistic which quantifies the difference between their data distributions. Specifically, in the Alzheimer’s disease data, the difference between distributions of features of Female and Male subjects has a large MMD statistic (0.17). Broadly, the distance between distributions of different age groups increases gradually as the age gap grows. Distribution shifts due to different ethnicities and clinical studies are smaller than those across sex and age groups. \cref{fig:app:dendrogram_scz,fig:app:dendrogram_age} show similar dendrograms for schizophrenia and brain age data. For schizophrenia data, there is a large difference in the distribution for groups stratified by sex and clinical study while the differences in the distributions of groups due to age is relatively small. For brain age data, there is a large heterogeneity in the data corresponding to subjects from different clinical studies (e.g., SHIP and BLSA-3T are quite different, but UKBB and CARDIA are quite similar); in general, WHIMS and SHIP data are different from all other clinical studies considered here. Our technique to evaluate the MMD statistic between two distributions is a precise way to understand the differences between data from different groups. It also provides context while interpreting the results, as we will do often in the sequel. If two groups have similar data distributions, then---for a well-trained model---we should expect a small deterioration when it is trained on one group and evaluated on the other; and we should similarly expect small improvements when such a model is adapted using a few samples from the target group.

\subsection{Even if machine learning models can predict very accurately on the groups that they were trained on, they do not generalize as well to data from other groups}

We trained machine learning models using data from each group in our study using five-fold nested cross-validation and evaluated these models on data from other groups in the population. We report area under the receiver operating curve (AUC) for classification tasks (AD and SZ) and mean absolute error (MAE) for regression tasks (brain age prediction). We have developed elaborate data pre-processing techniques and performed extensive hyper-parameter search and model selection while fitting machine learning models, which are detailed in~\cref{s:methods}, in particular~\cref{s:models,s:pipeline}. As a sanity check, for all three problems, neural networks trained on this pre-processed data can predict  accurately on all groups, with accuracy varying depending on the task's difficulty; for AD the AUC is at least 0.92, for SZ the AUC is at least 0.7, for brain age prediction the MAE is at most 5.58 years. These results are slightly better than existing results in the literature~\autocite{wen2020convolutional, rozycki2018multisite, bashyam2020mri}{}.

Comparing the above numbers to the translucent markers in~\cref{fig:ad_nn}, there is a drop in the performance when neural networks trained on data from one group using the same training procedure are evaluated on other groups. For example, in AD classification, the neural network trained on Males achieves 0.928 ± 0.001 AUC on Females, which is less than training directly on Females (0.942 ± 0.010) at a significance level $p = 4\times 10^{-4}$. Training only on subjects with age $\leq$ 65 years has much higher AUC (0.935 ± 0.019) compared to training on $\geq$ 80 years old (0.825 ± 0.013) and evaluating on $\leq$ 65 year olds; $p = 1.3\times 10^{-5}$. For all three tasks, for all attributes, we find that when a model trained on data from another group is evaluated on data from a target specific group, it achieves a worse performance as compared to training on data from only the target group ($p < 0.01$). There are five exceptions: Asian as target group in AD, and Female, 30--35 years age, Native Americans and Asian as target group in SZ. This suggests that even if, with appropriate pre-processing and model selection, neural networks trained on multi-source data can predict accurately, they do not generalize to data from groups that are different from the ones that were used for training.

We next constructed ensembles of machine learning models (neural networks and different kinds of random forests) using classical techniques such as bagging, boosting and stacking. For all three problems, for all groups, our ensembles (bold markers in~\cref{fig:ad_nn,fig:app:ad_nn}) predict more accurately than the corresponding neural networks (translucent markers) ($p < 10^{-3}$). For AD classification, the ensemble trained on Males has an AUC of 0.957 ± 0.001 on Females whereas the neural network trained has an AUC of 0.928 ± 0.001; in some cases the differences are extremely large, for example, models trained on Asian show AUC 0.793 ± 0.003 (ensemble) and 0.365 ± 0.003 (neural net) on White separately. Similarly, for SZ classification (\cref{fig:app:scz_nn}), the neural network trained directly on data from Females has 0.620 ± 0.096 AUC whereas the ensemble has an AUC of 0.740 ± 0.015; the ensemble (0.674 ± 0.068) predicts more accurately than the neural network (0.567 ± 0.057) for Asians. However, the ensemble also does not generalize to groups outside of the training set. For brain age prediction (\cref{fig:app:age_nn}), the ensemble trained on Asians predicts less accurately (6.24 ± 0.0 MAE) on Whites, compared to directly training the model on the target sub–group (3.99 ± 0.1 MAE). Similarly, the ensemble trained on data from the WHIMS study has a much smaller error (2.67 ± 0.1 MAE) than that of a model trained on data from CARDIA study (15.91 ± 0.0 MAE). This suggests that even if ensembles can lead to improvements in predictive performance as compared to neural networks, their AUC still deteriorates when evaluated on data from groups outside of the training set.

\subsection{Generalization to groups outside of the training set can be improved with access to a small amount of labeled samples}

There are many methods in the current literature to enable machine learning models to generalize better to data outside of the training set. These methods are based on the idea that one can learn representations that are invariant to the group and thereby, ensuring that predictions of the model are robust to different groups~\autocite{ganin2016domain, arjovsky2019invariant, zhao2020training, moyer2020scanner, dinsdale2021deep}{}. When data is heterogeneous, it is difficult to learn representations that are invariant and yet have good predictive ability. Indeed it has been argued before that the accuracy of output predictions of an invariant representation is limited by the domain with the least amount of information to begin with~\autocite{moyer2021harmonization}{}.

We have developed a mathematical argument in~\cref{s:theory} that elucidates inevitable trade-offs in the predictive ability when invariant representations are learned using heterogeneous data. It has two key implications. First, if we seek good performance on the target group, it is necessary to explicitly train the model using a small number of labeled samples from the target group in addition to data from the source group. Second, in order to effectively adapt the model to new data from the target group, we also need access to the data that it was trained upon, i.e., data from the source group; just having parameters of the pre-trained model is not sufficient. Based on this theory, we have developed a method that uses a weighted-ERM objective to adapt to new groups. Suppose we have $n$ labeled samples from the target group and $m$ samples from the source group. We construct a joint training set $\cbr{(x_i, y_i)}_{i=1}^{m+n}$ and fit a model with parameters $\th$ to minimize the weighted objective:
\beq{
    -\f{(1-\a)}{m} \sum_{i=1}^m \log p_\th(y_i \mid x_i) -\f{\a}{n} \sum_{i=m+1}^{m+n} \log p_\th(y_i \mid x_i);
    \label{eq:weighted_erm}
}
for a hyper-parameter $\a \in [0,1]$ that depends on $m$ and $n$. The theory gives guidelines to choose this hyper-parameter $\a$, e.g., if data from the target group has a similar distribution as that of the data from the source group and $m \sim n$, then $\a \sim 0.5$; if $m \gg n$, then we have few samples from the target group and should therefore use an objective with $\a \ll 1$ to predominantly use the samples from the source group for training. In this paper, we treat $\a$ as a hyper-parameter and search for its optimal value using nested cross-validation; the domain of the search is chosen using these guidelines. For all three problems, we trained both the neural network and the ensemble using this $\a$-weighted ERM objective for two settings: when we have 10\% data from the target group, and when we are allowed to use all data from the target group (in practice, this amounts to 80\% of the data because we treat the remainder as the held-out test set). The first setting corresponds to realistic situations when it is reasonable to procure some data from the target group for the purposes of building better diagnostic models; the second setting when adaptation is performed using all target data can be thought of as an upper bound on the performance.

\begin{figure}
\centering
\includegraphics[width=0.49\linewidth]{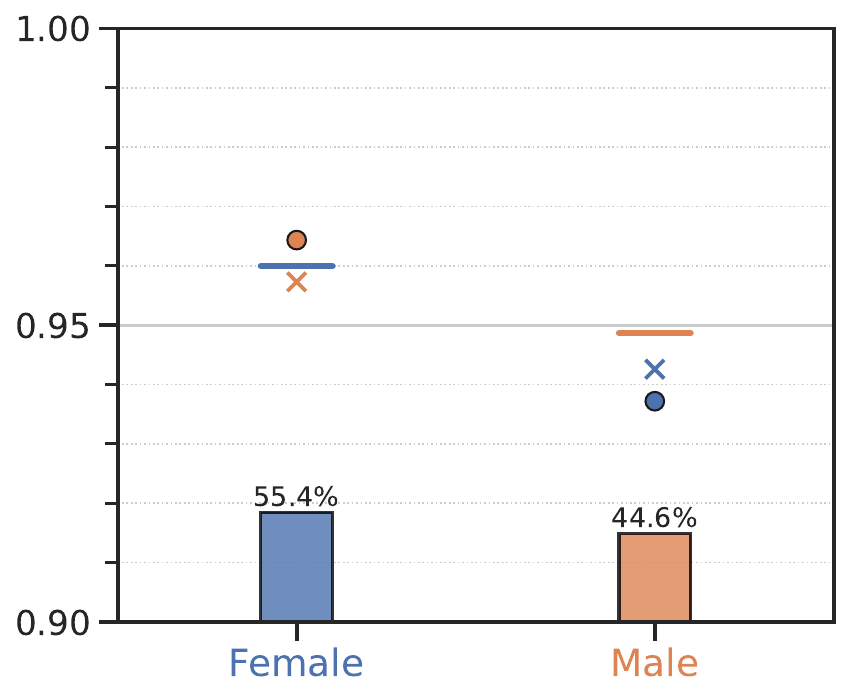}
\includegraphics[width=0.49\linewidth]{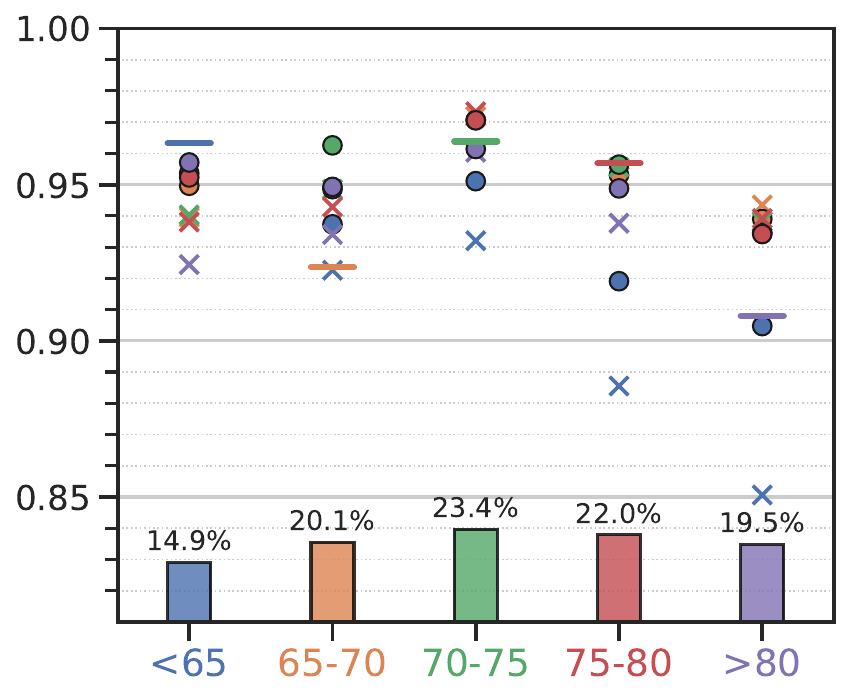}\\
\includegraphics[width=0.49\linewidth]{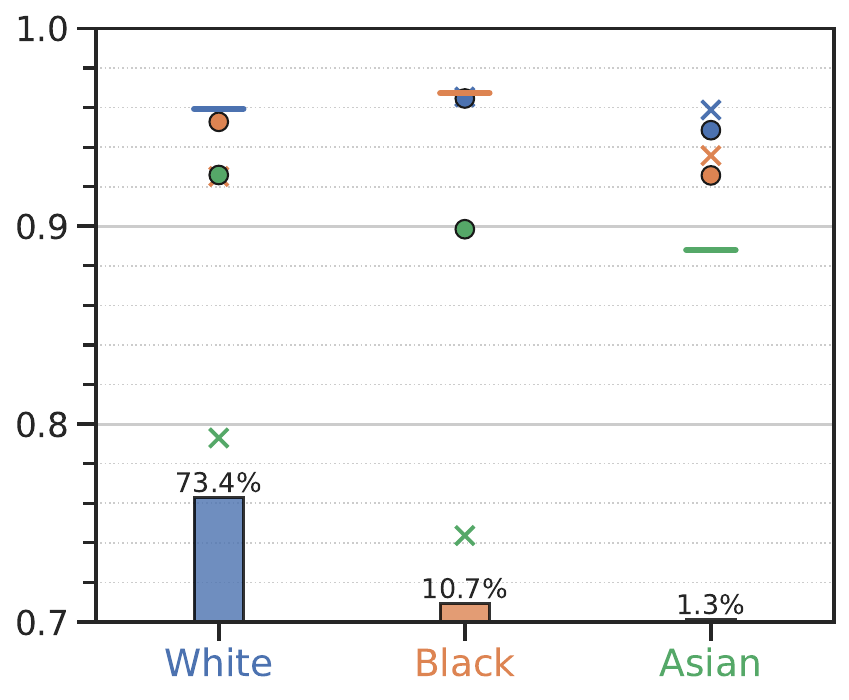}
\includegraphics[width=0.49\linewidth]{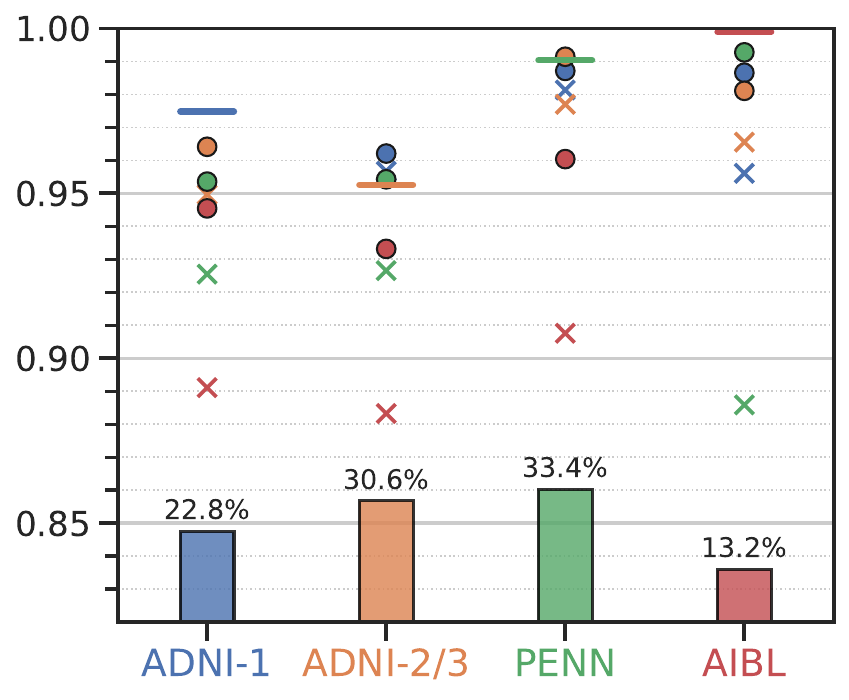}\\
\includegraphics[width=0.55\linewidth]{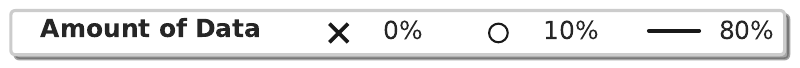}\\
\includegraphics[width=0.24\linewidth]{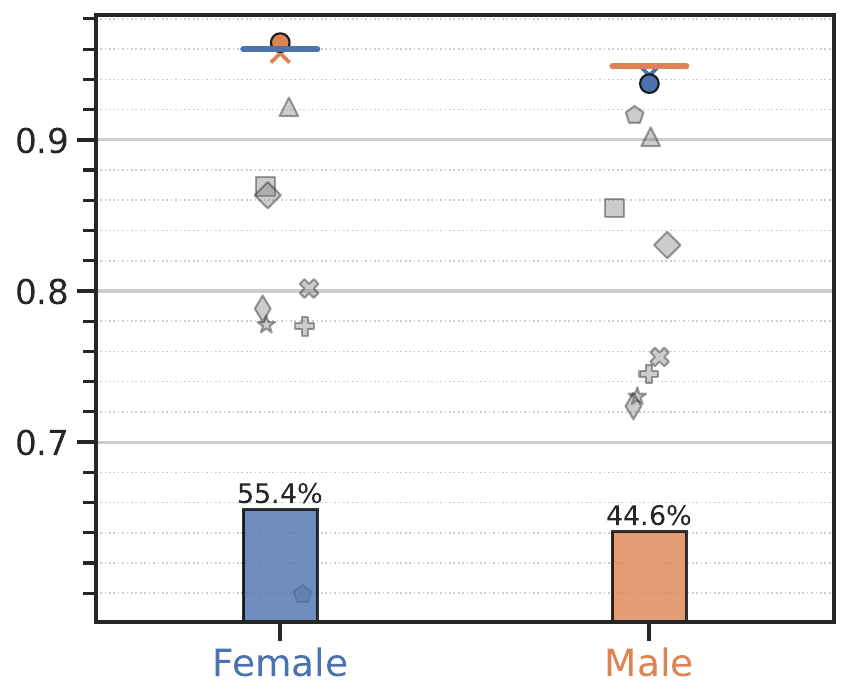}
\includegraphics[width=0.24\linewidth]{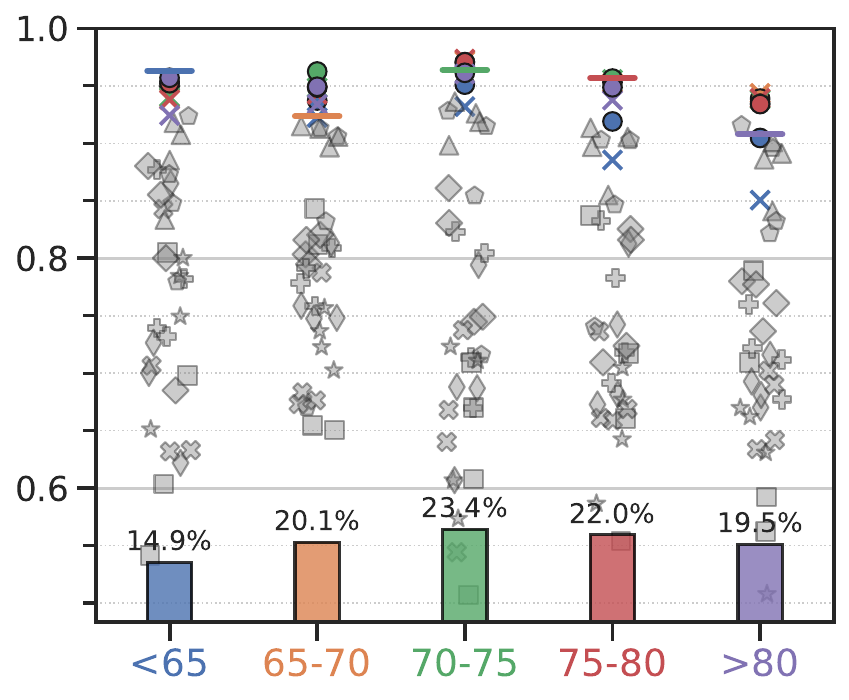}
\includegraphics[width=0.24\linewidth]{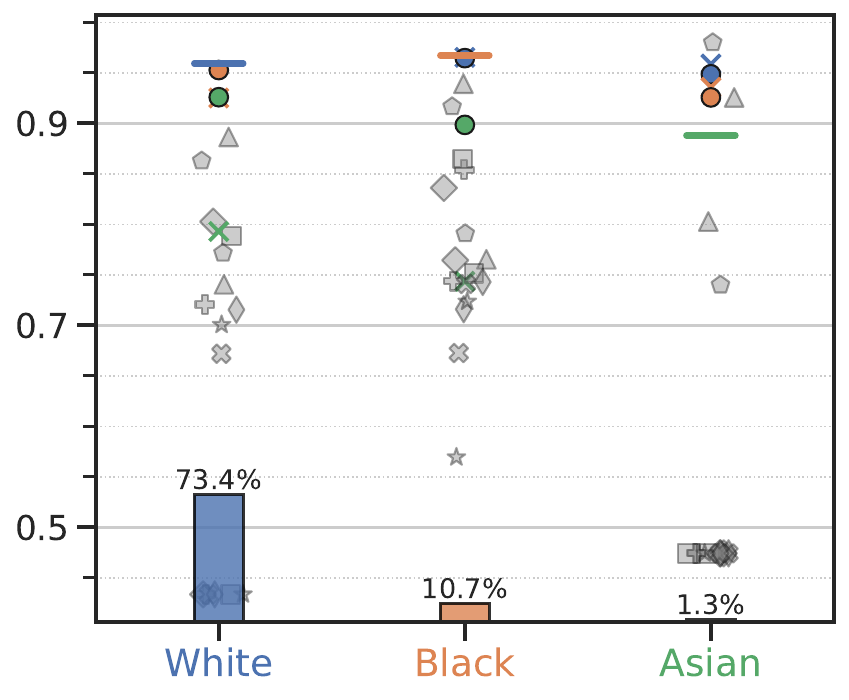}
\includegraphics[width=0.24\linewidth]{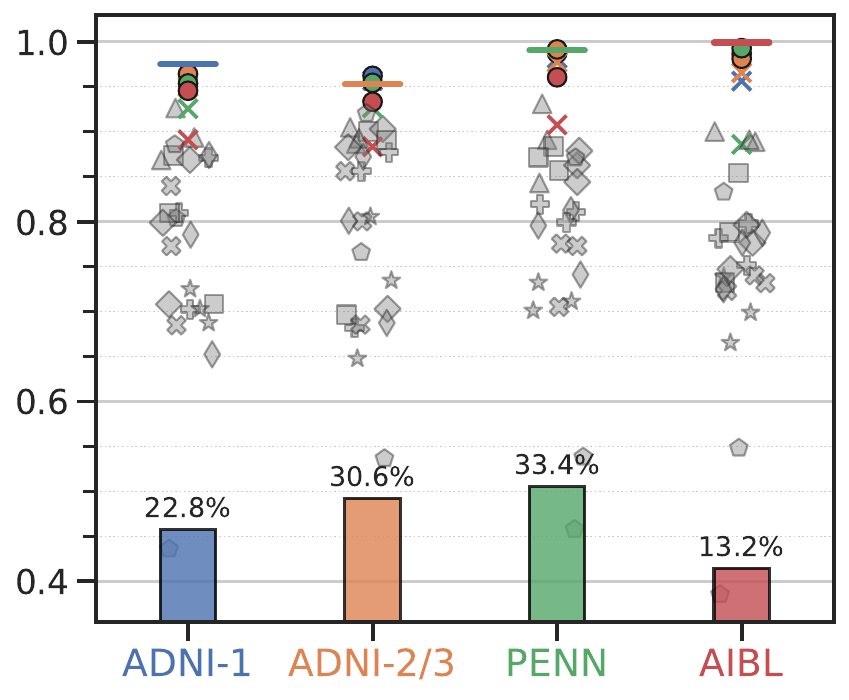}
\includegraphics[width=0.8\linewidth]{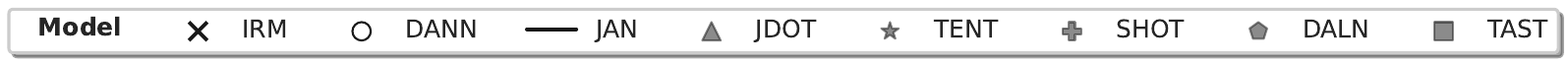}
\caption{
\textbf{Alzheimer's disease classification (see~\cref{tab:ad_table} for numerical data).} Markers denote the average AUC on the target group computed using five-fold nested cross-validation for models trained only on data from the target group (e.g., Female subjects, denoted by the blue horizontal line), only on data from the source group (e.g., trained on all Male subjects and evaluated on Female subjects is denoted by the orange cross), and trained on all data from the source group and 10\% data from the target group (orange circle). Panels denote groups stratified by one of the four attributes, namely sex, age group, race and clinical study. Bar plots denote the proportion of subjects in these groups in our study. All models are ensembles trained using features derived from structural measures, demographic and clinical variables, genetic factors, and cognitive scores. In spite of imbalances in the proportion of data in different groups, the AUC of the ensemble is consistently high (above 0.85 in all cases except when transferring from models built from Asians). The gap in predictive performance of a model trained on only target data (horizontal lines) and a model trained only on source data (crosses) can be improved with access to as little as 10\% data from the target group (circles) for Male, < 65 years, > 80 years, Asian, ADNI-1, ADNI-2/3, PENN and AIBL, when transferring from any of other groups ($p$ < 0.005). The improvement in AUC using 10\% target data is not statistically significant for the other groups; in one case (Female) we also see deterioration after including the target data perhaps due to confounding factors. We observe that the AUC for the > 80 years subgroup is low compared to other age groups even for models directly trained on this group. This might be due to the strong normal aging effects which make it difficult to distinguish cognitively normal individuals from  AD patients.
%\newtext{
In the lower panel, we also compare the proposed model with 8 representative domain adaptation/generalization techniques including IRM~\autocite{arjovsky2019invariant}{}, DANN~\autocite{ganin2016domain}{}, JAN~\autocite{long2017deep}{}, JDOT~\autocite{courty2017joint}{}, TENT~\autocite{wang2020tent}{}, SHOT~\autocite{liang2020we}{}, DALN~\autocite{chen2022reusing}{}, and TAST~\autocite{jang2022test}{} as shown in grey markers. See~\cref{s:baselines} for details of these methods.}
%}
\label{fig:ad}
\end{figure}

\begin{figure}
\centering
\includegraphics[width=0.49\linewidth]{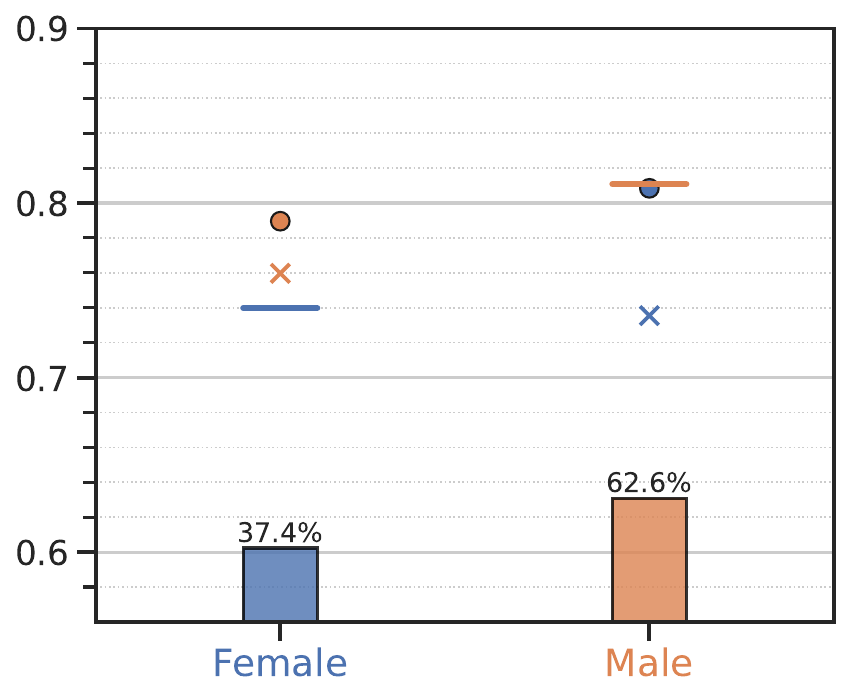}
\includegraphics[width=0.49\linewidth]{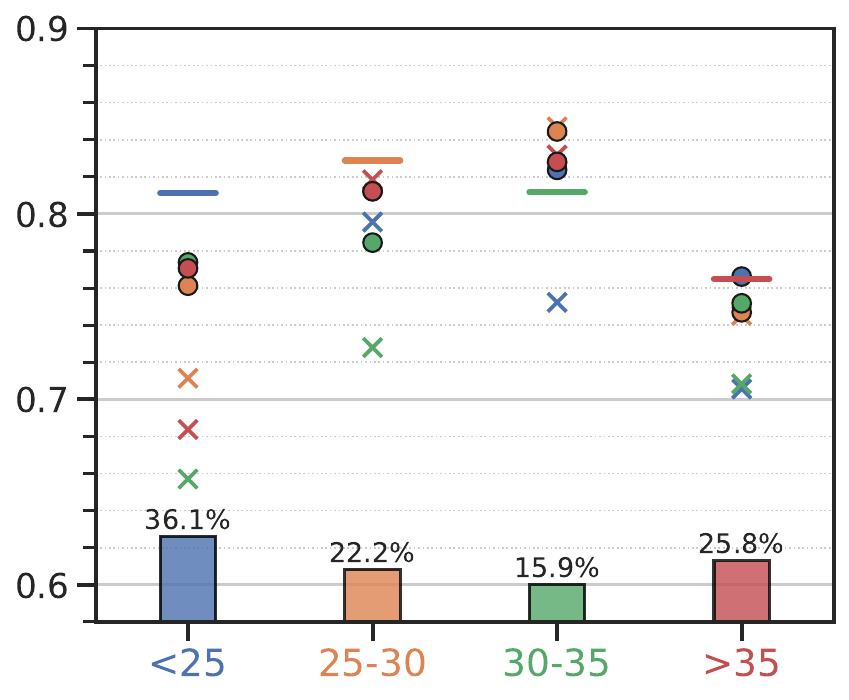}\\
\includegraphics[width=0.49\linewidth]{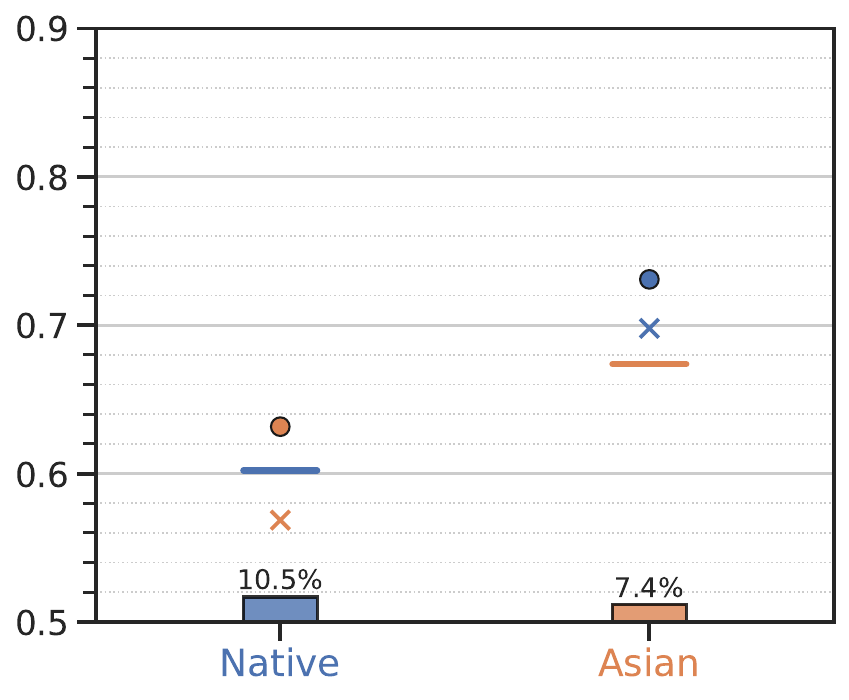}
\includegraphics[width=0.49\linewidth]{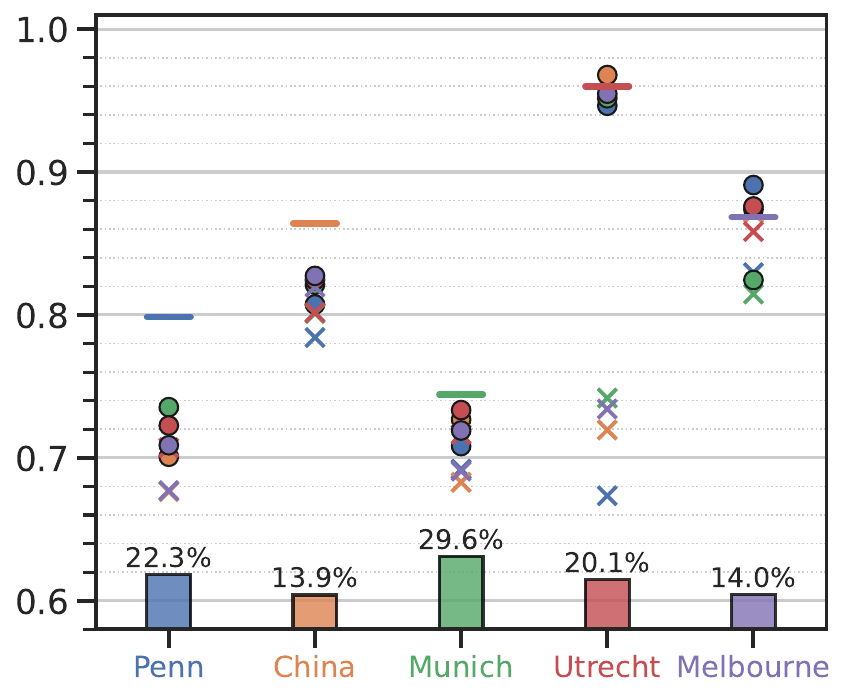}\\
\includegraphics[width=0.55\linewidth]{legend}\\
\includegraphics[width=0.24\linewidth]{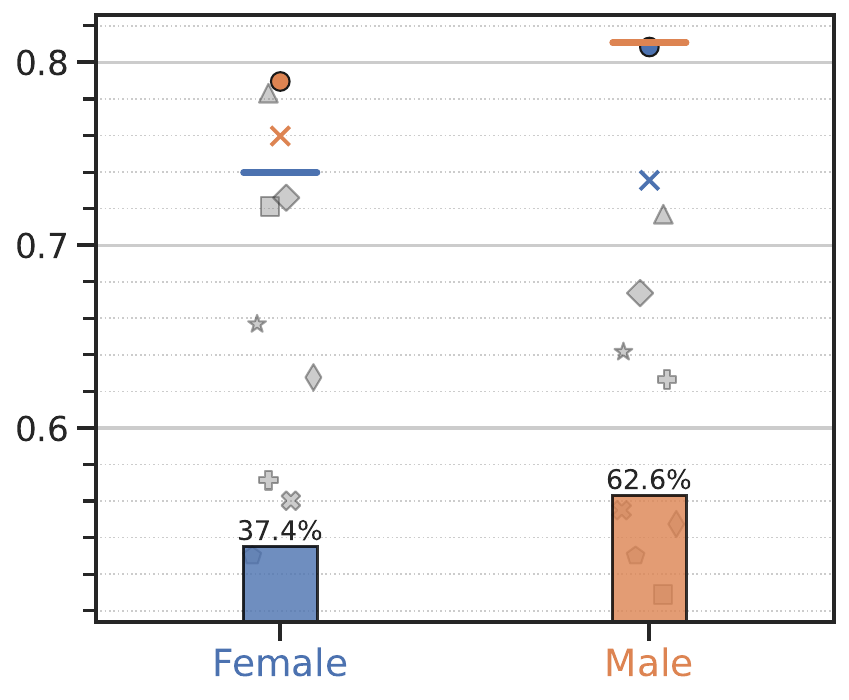}
\includegraphics[width=0.24\linewidth]{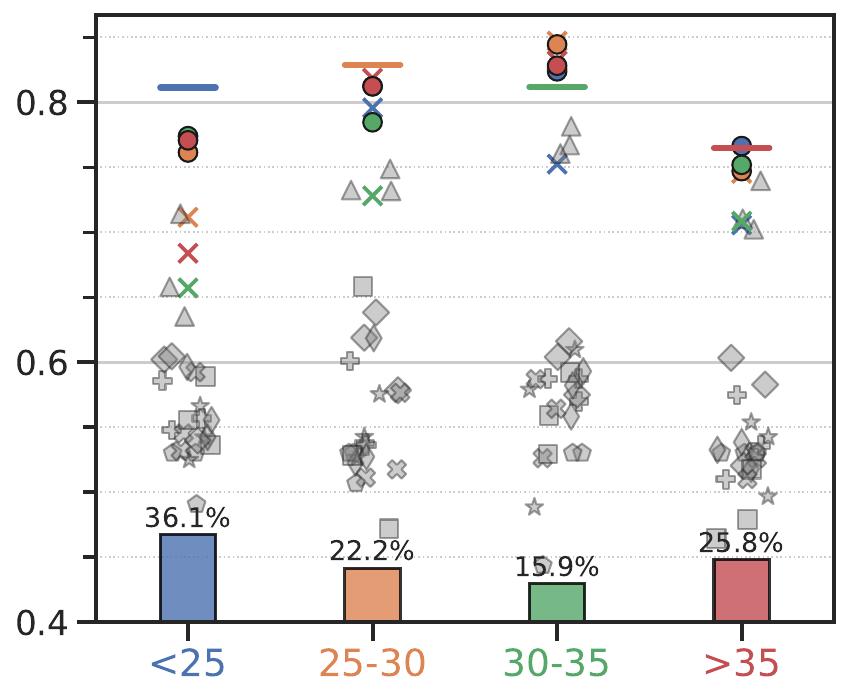}
\includegraphics[width=0.24\linewidth]{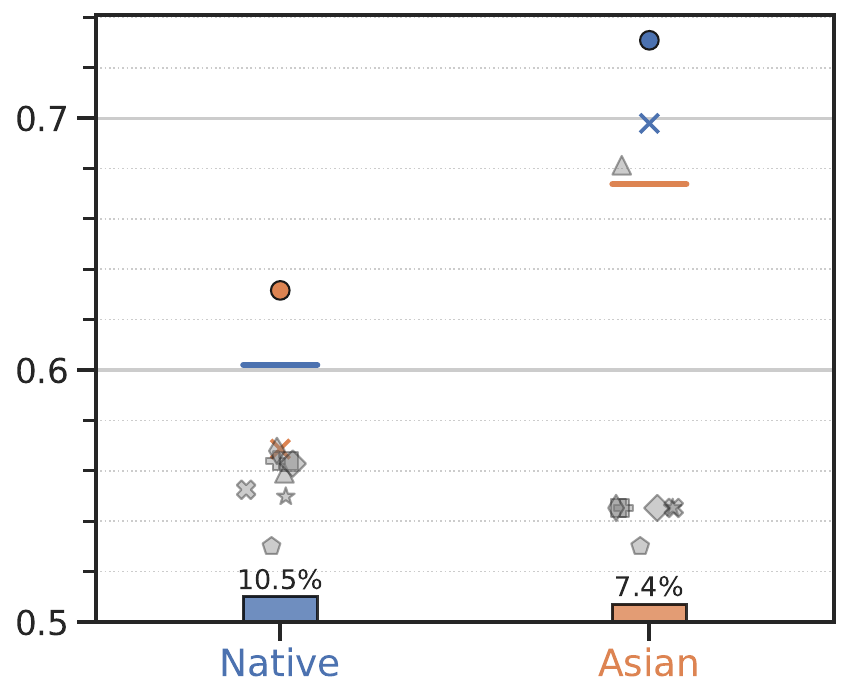}
\includegraphics[width=0.24\linewidth]{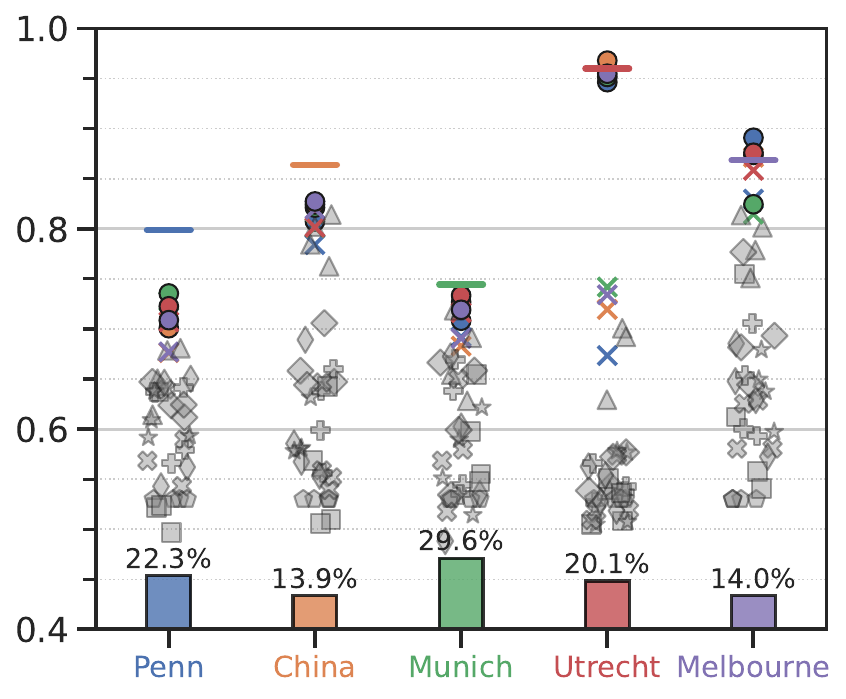}
\includegraphics[width=0.8\linewidth]{legend_bg}
\caption{
\textbf{Schizophrenia classification (see~\cref{tab:scz_table} for numerical data).}
Markers denote the average AUC of the ensemble on the target group computed using five-fold nested cross-validation for models trained only on data from the target group (e.g., Female subjects, denoted by the blue horizontal line), only on data from the source group (crosses), and trained on all data from the source group and 10\% data from the target group (circles). Compared to~\cref{fig:ad}, the AUC for schizophrenia classification is lower in general, as expected based on respective prior literature. We find that $\a$-weighted ERM using 10\% data from the target group improves the AUC of the ensemble (circles are above crosses of the same color) in all cases except two: 25-30 years old and 30-35 years old. In most cases, models adapted from source groups using 10\% data from the target group perform better than those trained on all target data, except when target groups are Male, > 35 years old, Munich and Utrecht, when the difference is statistically insignificant. We observe large performance discrepancies between different clinical studies. Besides scanner and acquisition protocols variations, disease severity might be playing a role here. For example, the AUC of China cohort is large perhaps because on-site clinical cases are usually relatively more severe clinically, largely due to cultural factors influencing who and when will seek hospitalization.
%\newtext{
In the lower panel, we also compare the proposed model with 8 representative domain adaptation/generalization techniques including IRM~\autocite{arjovsky2019invariant}{}, DANN~\autocite{ganin2016domain}{}, JAN~\autocite{long2017deep}{}, JDOT~\autocite{courty2017joint}{}, TENT~\autocite{wang2020tent}{}, SHOT~\autocite{liang2020we}{}, DALN~\autocite{chen2022reusing}{}, and TAST~\autocite{jang2022test}{} as shown in grey markers. See~\cref{s:baselines} for baseline method details.}
%}
\label{fig:scz}
\end{figure}

\begin{figure}
\centering
\includegraphics[width=0.49\linewidth]{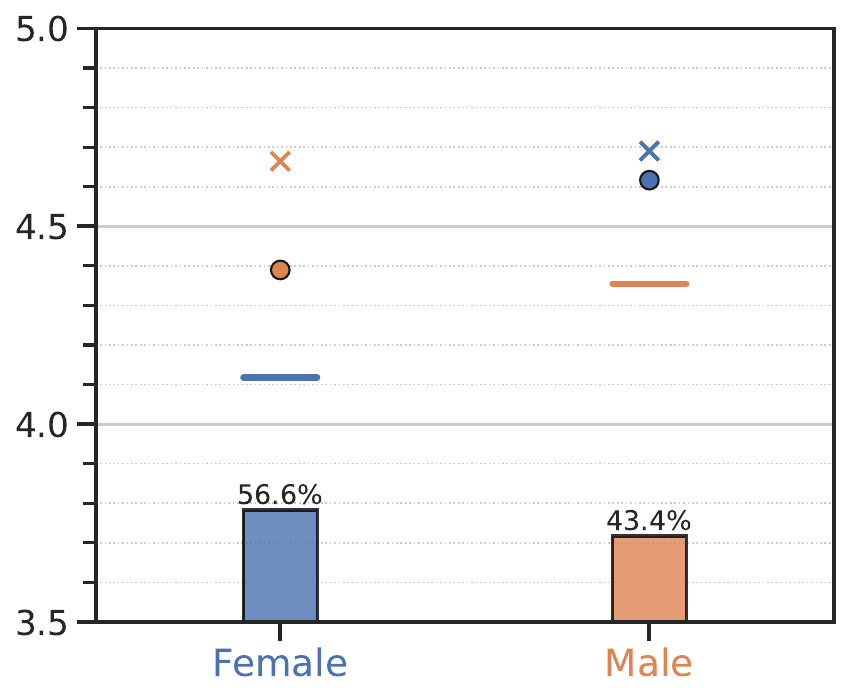}
\includegraphics[width=0.49\linewidth]{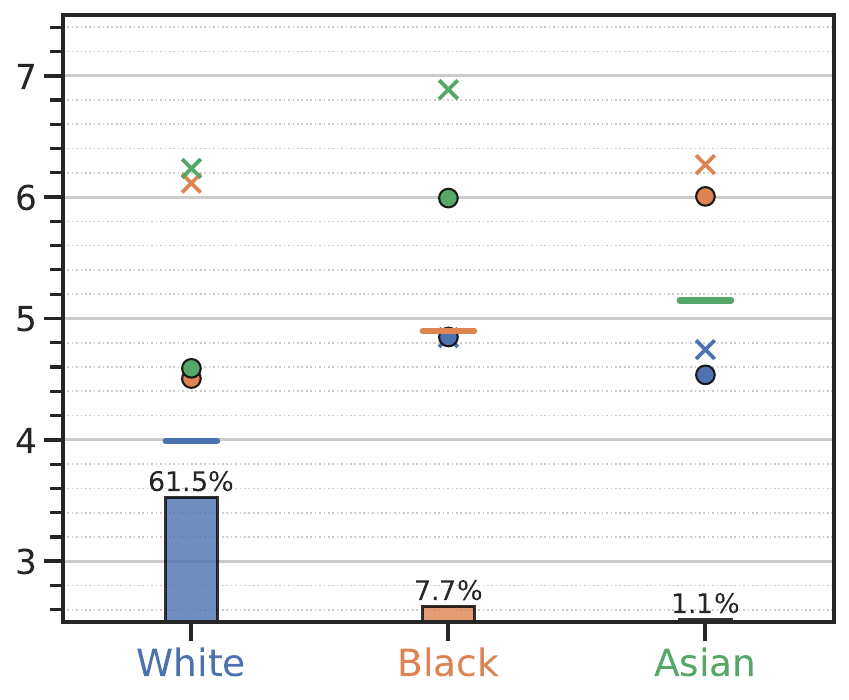}\\
\includegraphics[width=1.0\linewidth]{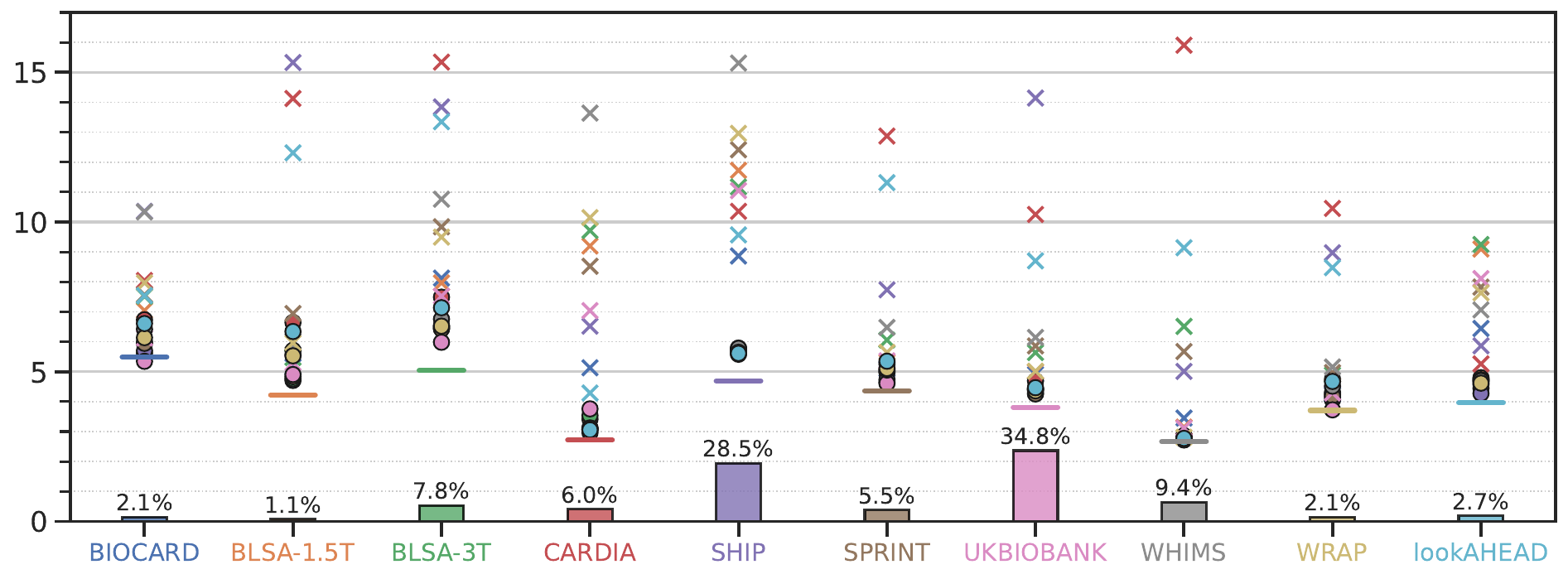}\\
\includegraphics[width=0.55\linewidth]{legend}
\caption{
\textbf{Brain age prediction (see~\cref{tab:age_table} for numerical data).}
Markers denote the mean absolute error (MAE) in years of an ensemble that predicts the brain age on different target groups in the population computed using five-fold nested cross-validation, for models trained only on data from the target group (e.g., Female subjects, denoted by the blue horizontal line), only on data from the source group (crosses), and trained on all data from the source group and 10\% data from the target group (circles). In general, the MAE of brain age prediction is remarkably small, it is below 7 years for age and race and below 15 years in most settings when models were trained on different clinical studies. Ensembles trained using 10\% data from the target group in addition to all data from the source group improve the MAE in all cases (circles are below crosses) except one (when source is White and target is Black). The third panel has 10 different clinical studies, with very different amounts of data. Even in this case the MAE of brain age prediction is smaller than 8 years in all cases when the ensemble has access to some data from the target group, in some cases there are significant improvements as compared to the corresponding crosses. Magnetic field strength of the scanners affects the models performance significantly. For example, only BLSA-1.5T and SHIP are acquired from 1.5T devices and others are from 3T ones. We can see big MAE gaps between the horizontal lines and crosses in BLSA-1.5T and SHIP studies. We also observe that larger  data size gves rise to better the performance. For example, UKBB has the largest sample size among all studies and models trained on UKBB usually have lower MAE when adapting to other studies.
}
\label{fig:age}
\end{figure}

For all three problems, for all groups, $\alpha$-weighted ERM with 10\% data improves the performance of ensemble models significantly\footnote{We discuss the performance of the ensemble models here in the main text; neural network-based experiments shown in the Appendix exhibit similar trends but with lower AUCs and higher MAEs.}. For AD classification in~\cref{fig:ad}, the AUC for target groups such as sex (Male), age group (< 65 and 65-70 years old), race (White), and clinical studies (ADNI-1, ADNI-2/3, PENN and AIBL) can be improved no matter which source group the model was adapted from. For SCZ classification in~\cref{fig:scz}, all target groups show increased AUC except two: 25-30 and 30-35 years old. For brain age prediction in~\cref{fig:age}, we also observe improved performance on nearly all target groups except one (Black). The difference in AUC from using 10\% vs.\@ 20\% data from the target group is not statistically significant ($p$-value > 0.01) or AD and SCZ classification as also MAE for brain age prediction. 
%However, for brain age prediction, the extra data reduces the MAE for target groups pertaining to race (White) and clinical study (BLSA-3T, SHIP and UKBB). 
More details are provided in the Appendix.

Ensemble models adapted using only 10\% target data achieve comparable performance to that of training on all data from the target group. They are sometimes even better. For example, in AD classification, models adapted on target groups (< 65, 70-75 years old and AIBL) have a similar AUC to that achieved when training on all data from the target group ($p$-value > 0.01) irrespective of the source group they were trained with. A model trained on the 70-75 years old group achieves 0.971 ± 0.002 AUC when adapted to the 65-70 years old target group---this is much higher than training directly on all data from the target group (0.924 ± 0.019, $p$-value 4.3$\times10^{-3}$). For SCZ classification, there is no statistically significant difference between models trained on all data from target groups (Male, > 35 years old, Munich and Utrecht) and models adapted from any of the source groups. For brain age prediction, we found that models trained on any source group when adapted to groups in the WHIMS study perform on par with the ones directly trained on all WHIMS data ($p$-value > 0.01).

%\newtext{
As shown in the lower panel in Fig.~\ref{fig:ad} and Fig.~\ref{fig:scz}, we compare the proposed model with 8 representative domain adaptation/generalization techniques, including IRM~\autocite{arjovsky2019invariant}{}, DANN~\autocite{ganin2016domain}{}, JAN~\autocite{long2017deep}{}, JDOT~\autocite{courty2017joint}{}, TENT~\autocite{wang2020tent}{}, SHOT~\autocite{liang2020we}{}, DALN~\autocite{chen2022reusing}{}, and TAST~\autocite{jang2022test}{}, on Alzheimer’s disease and schizophrenia classification under the exact same experimental setup as that of our method. See~\cref{s:baselines} for details of these existing methods.
We find that these approaches are unable to outperform even empirical risk minimization (denoted by a cross) significantly. This is very surprising because these are different domain adaptation techniques, some that do not use data from the target task, some that do; ERM does not use any data from the target task.
This may seem unusual and counter-intuitive to many readers, but it is actually consistent with many rigorous empirical comparison studies in the literature in computer vision~\autocite{gulrajani2020search, galstyan2022failure} as also medical imaging~\autocite{zhang2021empirical, korevaar2023failure, guo2022evaluation, pfisterer2022evaluating} communities.
On the other hand, our weighted ERM approach (denoted by circles) consistently outperforms existing methods in all scenarios.
%}

%\newtext{
We also assessed the fairness of these models with respect to sensitive attributes including sex, age group, race, and clinical study. In~\cref{fig:app:fairness} we show the demographic parity differences (DPD)~\autocite{hardt2016equality, agarwal2019fair} and equalized odds difference (EOD)~\autocite{feldman2015certifying} on five different held-out subsets of data for diagnosis of Alzheimer’s disease and schizophrenia, and brain age prediction. We find that our $\alpha$-weighted ERM models have significantly lower disparities (zero indicates a fair model under these metrics) for all sensitive attributes in both metrics compared to a deep network  and ensembles models.
%}

\subsection{Diagnostic models for Alzheimer’s disease can also be effective for early diagnosis of stable and progressive mild cognitive impairment. Adapting models to the target group further improves this ability.}

\begin{figure}
\begin{subfigure}[b]{\linewidth}
\centering
\includegraphics[width=0.325\linewidth]{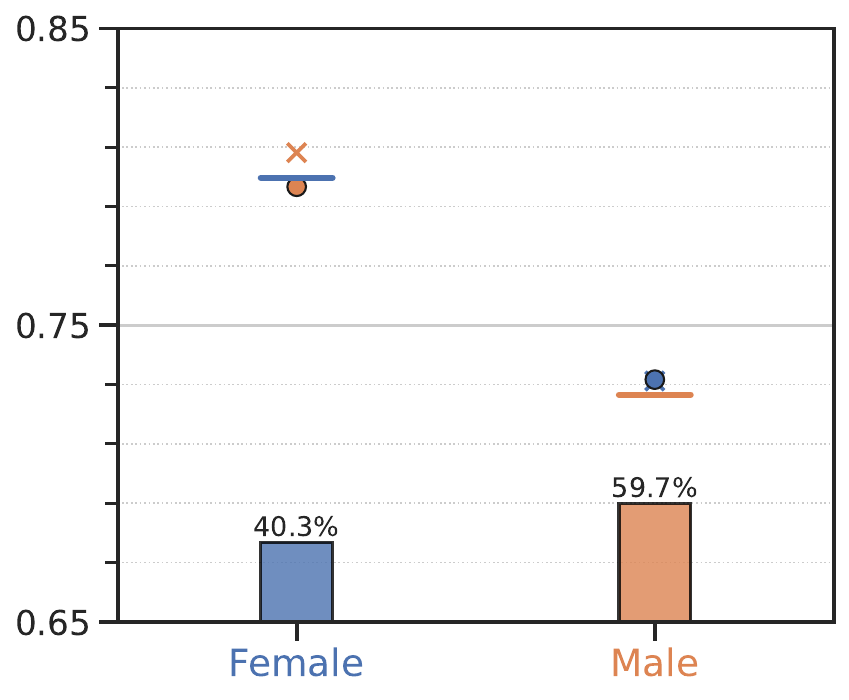}
\includegraphics[width=0.325\linewidth]{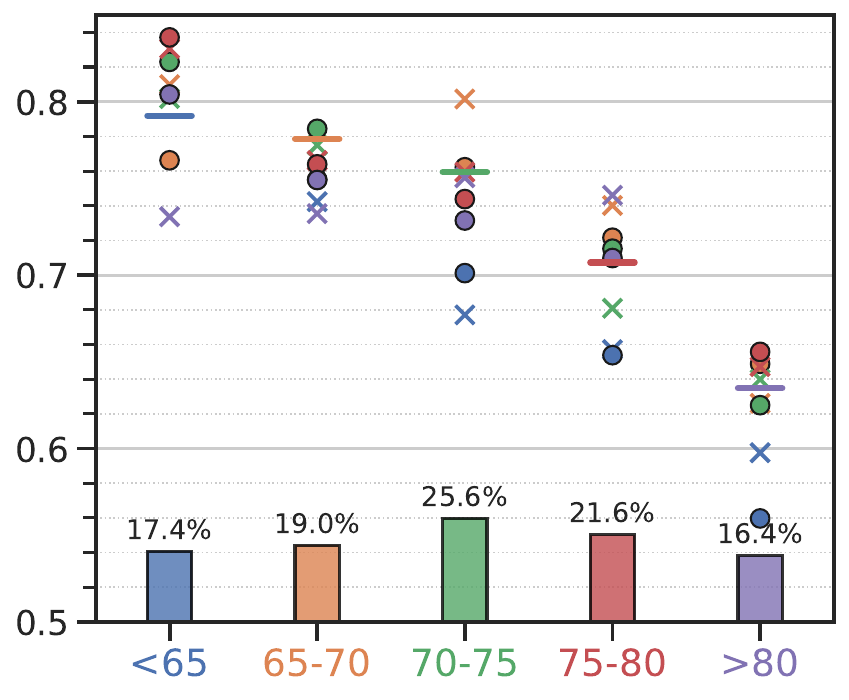}
\includegraphics[width=0.325\linewidth]{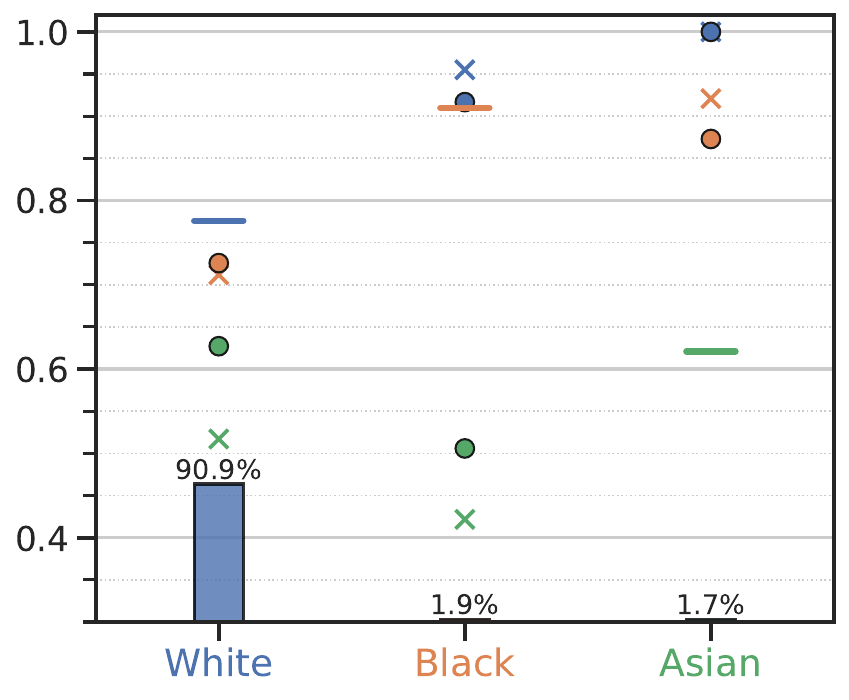}
\includegraphics[width=0.35\linewidth]{legend}
\caption{Progressive MCI v.s. stable MCI}
\label{fig:s_mci}
\end{subfigure}

\begin{subfigure}[b]{\linewidth}
\centering
\includegraphics[width=0.49\linewidth]{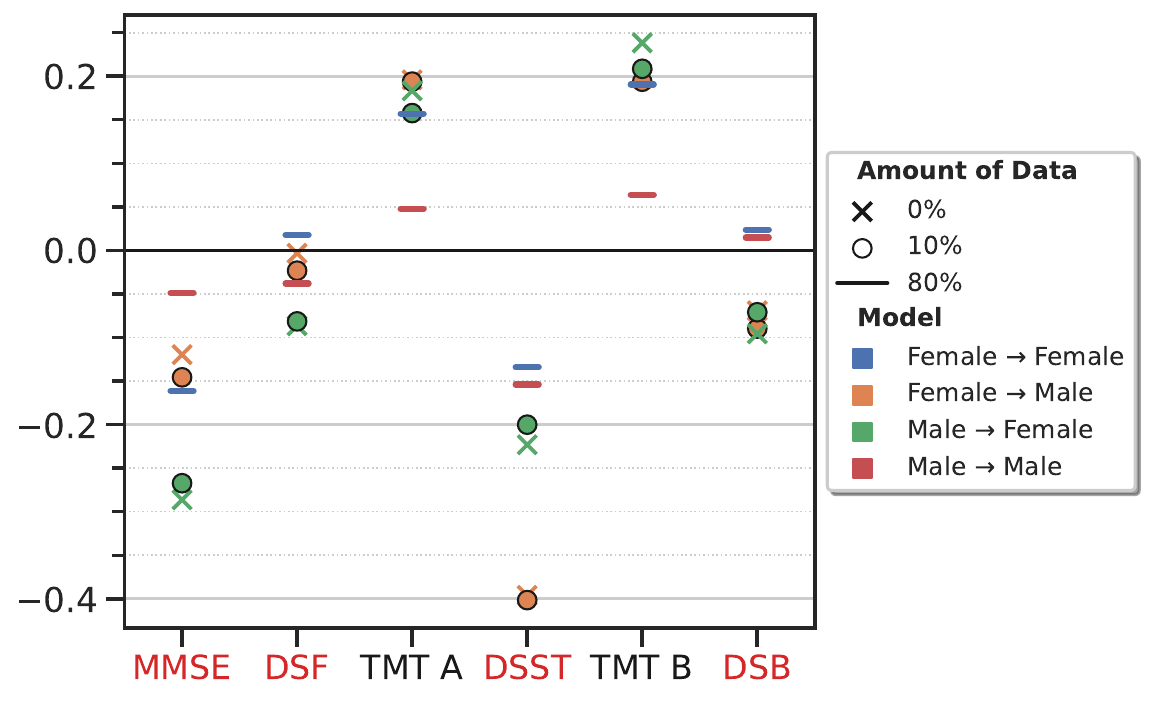}
\includegraphics[width=0.49\linewidth]{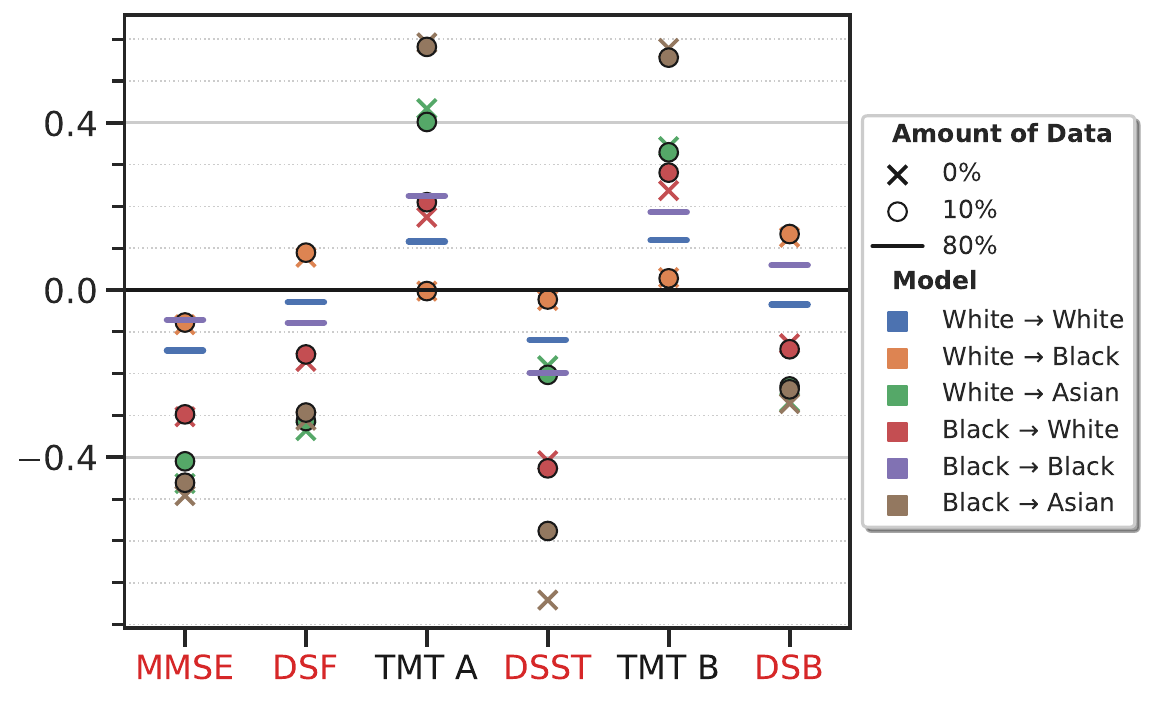}
\caption{Pearson's correlation of brain age regression with neuropsychological tests}
\label{fig:cognitive}
\end{subfigure}
\caption{
\textbf{Adapting diagnostic models to target groups using a small amount of data also improves their ability to make predictions on secondary tasks; see~\cref{tab:mci_table,tab:cog_table_1,tab:cog_table_2} for numerical data.}
\textbf{(a)} Linear discriminant analysis on the output probabilities (that determines AD vs. cognitively normal CN) of the ensemble models trained for Alzheimer’s disease diagnosis is used to study whether subjects with mild cognitive impairment (MCI) progress to AD (known as pMCI) or remain stable MCI (known as sMCI) using only the baseline scans. The AUC of pMCI vs. sMCI on the target group is shown for three different attributes (sex, age group and race) when models are trained only on data from the source group (crosses), using $\a$-weighted ERM using all data from the source and 10\% data from the target group (circles) and with access to only all data from the target group (horizontal lines). Improvements in the AD vs. CN AUC of these models with 10\% data translate to improvements in the ability to distinguish between pMCI and sMCI subjects, using only baseline scans (circles above cross) except when target groups are Black or Asian (due to very little data in these groups). For all type of models, performance decreases as the age of the participants increases; this is because predicting progressive MCI using baseline scans is more and more challenging when the time difference to the target age group and the normal aging effect increases.\\[0.25em]
\textbf{(b)} Pearson's correlation between the brain age residual (predicted brain age minus chronological age) and neuropsychological tests for two different attributes (sex and race) for models trained only on source data (crosses), using $\a$-weighted ERM on all source data and 10\% target data (circles) and only on all target data (horizontal lines). Unlike other plots, colors denote different pairs of source and target groups. Tests (X-axis) marked in red are expected to be negatively correlated with brain aging whereas those marked in black are expected to be positively correlated with brain aging according to the existing literature. Mini-mental state examination (MMSE) is a questionnaire test that measures global cognitive impairment. Digit span forward/backward (DSF/B) test is a way of measuring the storage capacity of a person's working memory. Trail making test part A/B (TMT A/B) measures a person's executive functioning. Digit symbol substitution test (DSST) is another global measure of cognitive ability, requiring multiple cognitive domains to complete effectively. In almost all cases, we observe stronger correlations than those reported in the literature. Models trained using 10\% target data improve the correlation with these neuropsychological tests. Brain age models trained from other groups usually have larger correlations to cognitive scores than the ones directly trained on the target group.
}
\label{fig:mci_cognitive}
\end{figure}

Mild cognitive impairment (MCI) refers to the transition stage between the expected cognitive changes associated with normal aging and those due to dementia, most frequently of AD dementia. MCI progresses to AD at a rate of approximately 15\% annually  (referred to as progressive MCI; pMCI), but sometimes it remains stable (referred to as stable MCI; sMCI). Herein we adopt the definition of progression to reflect changes occurring within three years since the baseline visit~\autocite{wen2020convolutional}{}. Sometimes, MCI can also revert back to cognitively normal state, in which case it is oftentimes called normal MCI (nMCI). It is important to predict the future clinical progression in MCI, in order to guide patient management and clinical trial stratification~\autocite{jack2010hypothetical, sperling2011toward, zhu2021dual, bron2021cross, lou2021leveraging}{}.

We investigated whether our machine learning models from the previous section, which were built for diagnosing AD, can be used to predict which individuals with MCI will progress to AD. Individuals in the MCI group typically exhibit heterogeneous patterns with features that are intermediate between AD and CN; this is why using models trained to diagnose AD for this task is challenging and unusual. We are also interested in understanding whether the improvements to these models, when adapted to data from the target group, also translate to improvements in their predictive ability on MCI subjects. If this is the case, then we can fruitfully improve the predictive ability on MCI subjects by adapting models on AD data. In this study, we apply AD diagnostic models directly on the baseline MCI subjects to check if pMCI, which has a similar pathology as that of AD, can be distinguished from sMCI and nMCI at the earliest clinical stage. All of our models are probabilistic classifiers and return the (uncalibrated) probability of an AD diagnosis. Using this predicted probability as the covariate, we perform linear discriminant analysis (LDA)~\autocite{balakrishnama1998linear} using 20\% of the pMCI, sMCI and nMCI retrospectively-created labels of the baseline scans; the remainder of the MCI samples as the test set. This experiment is repeated in a five-fold cross-validation fashion to estimate the test error.  \cref{tab:app:mci} summarizes the data used for this analysis.

As~\cref{fig:s_mci} shows, we find that models trained for diagnosing AD from CN can also distinguish pMCI from sMCI  accurately using only baseline scans;~\cref{fig:app:n_mci} shows a similar result for pMCI vs. nMCI. For example, within the Female population, we obtained 0.891 ± 0.008 AUC for pMCI v.s. nMCI classification and 0.800 ± 0.010 AUC for pMCI v.s. sMCI classification. When an ensemble model trained using data from group > 80 years old is evaluated on data from age group < 65 years old, it obtained an AUC of 0.769 ± 0.010 and 0.734 ± 0.007 for pMCI vs. nMCI and pMCI vs. sMCI tasks, respectively. These numerical results are remarkable in the model was trained using AD vs. CN data and refashioned to make these predictions using 20\% of the data and LDA. When the ensemble model is adapted to the target AD group using 10\% of the data, we find that the performance improves further in some cases, in particular when the discrepancy between the data distributions the groups is large (see~\cref{fig:dendrogram_ad}); there are also situations when the AUC remains the same. For example, in the example above (source group is > 80 years old and target group is < 65 years old) $\a$-weighted ERM improves the AUC to 0.804 ± 0.010 and 0.804 ± 0.011 for pMCI vs. nMCI and pMCI vs. sMCI tasks, respectively. Just like the previous section, even in this task, we observed that sometimes the performance of the adapted AD models surpasses that of models trained directly on the target domain. For example, within Black population, AUC for pMCI vs. nMCI is 0.825 ± 0.170 using a model trained only on data from this group, while a model trained using $\a$-weighted ERM on AD data  from the White group along with 10\% data from the Black group achieves an AUC of 0.945 ± 0.048.

Our results indicate that models which can accurately diagnose AD from CN are also able to make predictions on whether MCI subjects will evolve into progressive, stable or normal MCI by only using baseline scans. Access to a small subset of the target group data for AD subjects further improves the performance of these predictions.

\subsection{Improved models for predicting brain chronological age lead to new clinical conclusions about the correlation between brain age residual and neuropsychological tests}

The brain age residual (BAR) or brain age gap is the difference between the age predicted from a brain scan and actual chronological age of the subject. BAR is indicative of the deviation of a subject's cognitive performance from that of the normal population over time, a positive BAR suggests an accelerated aging process~\autocite{cole2017predicting}{}. Many recent studies~\autocite{smith2019estimation, jonsson2019brain, leonardsen2022deep, zhou2023multiscale} have argued that BAR can be used to quantify the brain aging process, e.g., via its correlation with cognitive impairment~\autocite{jonsson2019brain, boyle2021brain, zhou2023multiscale}{}. In light of this, we evaluate the association between BAR as predicted by our models and several neuropsychological tests, including visual memory, visual attention, and visual executive tests. We compute Pearson's correlation between BAR and the performance on mini-mental state examination (MMSE), digital symbol substitution test (DSST), digit span forward/backward (DSF and DSB), and trial making tests (TMT A and B). We also use cardiovascular features and lifestyle factors such as systolic and diastolic blood pressure, and body mass index (BMI) for association evaluation.

As~\cref{fig:cognitive} shows, we find that, in general, the BAR predicted by our ensemble models is negatively correlated with cognitive measures that decrease with age (e.g., MMSE and DSST) and positively correlated with those that increase with age (e.g., TMT A and B) with $p$-values < 0.01. Interestingly, we observe stronger correlations than those reported in the literature~\autocite{jonsson2019brain, zhou2023multiscale}{}. This is perhaps due to the improved predictive ability of our BAR models. In contrast, the correlations between the predicted BAR and cardiovascular and lifestyle factors are not statistically significant ($p$-value > 0.01)---this contradicts several previous studies~\autocite{smith2019estimation, zhou2023multiscale}{} (see~\cref{fig:app:cardio}). Our BAR prediction models are quite accurate (see~\cref{fig:app:dendrogram_nn_ensemble_age}), so this could be potentially consistent with the fact that our study did not use brain measurements which more directly reflect effects of cardiovascular risk factors, such as infarcts and microbleeds.

Similar to the previous section, we found that improving BAR predictions on target groups using 10\% data translates to improved associations with the clinical factors, as compared to the associations computed using BAR prediction models that did not use any target data. For example, Pearson's correlations between BAR and DSST/TMT A are -0.453 and 0.128 respectively when a model trained on Asian subjects is evaluated on subjects from the White group. If we instead improve this model using $\a$-weighted ERM and 10\% data from the White group, these correlations increase to -0.556 and 0.191 respectively. Just like the previous sections, models adapted to target groups often lead to greater associations between BAR and cognitive measures compared to those trained only on the target group. For example, within the Female population, the correlation between BAR and MMSE is -0.161, whereas a BAR prediction model trained using $\a$-weighted ERM trained on the source Male group achieves -0.267 on the target Female group. %\newtext{
Note that in this case, the correlation between BAR and MMSE in females is more negative when no data (i.e., 0\%) is used to adapt the male-trained model; this is perhaps due to the inadequacy of our hyper-parameter search for the weighing parameter $\a$, because having extra should never reduce the performance.
%}
% !TEX root = ../main.tex

\section{Discussion}

\paragraph{Carefully-designed machine learning models with effective domain adaptation techniques are necessary to ensure accurate predictions on different groups in heterogeneous populations.}
The one-fits-all paradigm, namely learning salient features from data, which is so central to machine learning, may not work well for problems with heterogeneous data, such as some encountered in clinical neuroscience ~\autocite{davatzikos2019machine, greene2022brain, kopal2023end}{}. In order to develop effective tools that can provide safe, robust and unbiased decisions, we need to address the heterogeneity at its different levels of granularity. For example, a large body of existing work builds representations that are invariant to features of the subjects in different groups in the population using domain generalization techniques~\autocite{tzeng2014deep,long2015learning,sun2016return,ganin2016domain,tzeng2017adversarial,liu2018detach,meng2020learning,dinsdale2020unlearning}{}. If the variability in data is not just nuisance variability, but is instead correlated with the outcome, building such invariance comes at the cost of predictive ability. For example, it has been shown using an information-theoretic argument that the performance of such a model is limited with the least predictive feature in the data~\autocite{moyer2021harmonization}{}. Building models that do not explicitly address this heterogeneity can lead to deterioration in performance when these models are used to make predictions, especially on groups that are under-represented in the population~\autocite{howard2021impact, ricci2022addressing, greene2022brain}{}. This debate has also led to concerns over the hype of precision medicine~\autocite{wilkinson2020time, finlayson2021clinician, roberts2021common}{}.

We provide a positive data point in the debate. We show that appropriately constructed machine learning models can predict both robustly and accurately on many different target populations. We also show theoretically that one model cannot predict accurately on all groups of a heterogeneous population. This suggests a natural approach to mitigating the deterioration of accuracy: it is based on an $\a$-weighted ERM objective using both the source and target data. Depending upon the value of $\a$ which is chosen using nested cross-validation, the source data can \emph{optimally} inform the representation that is eventually used to make predictions on the target data in different ways. Theoretically, we argued that the weighted ERM approach is at least as good as any other domain adaptation technique. The AUCs and MAEs demonstrated in this paper are at least as good, often much better, than those reported in the literature on these data. In contrast to existing techniques, the $\a$-weighted ERM objective is also extremely easy to implement, e.g., in this paper we have implemented it effectively for over 850 configurations of different source-target group pairs, amounts of target data and machine learning models with the exact same code. It may seem surprising that weighted ERM outperforms all existing domain adaptation methods. But there is a lot of corroborative evidence in different parts of the computer vision~\autocite{gulrajani2020search, galstyan2022failure,dhillon2019a} and medical imaging literature~\autocite{zhang2021empirical, korevaar2023failure, guo2022evaluation, pfisterer2022evaluating}{}.

%\newtext{
Generalizing to different groups in a problem with heterogeneous data is a multi-faceted issue and therefore one must be careful in drawing broad generalizations.
In general, we find that for attributes like sex and age where most groups have sufficient amounts of data, it is possible to predict accurately (AUC > 90 for Alzheimer’s disease classification, AUC > 70 for schizophrenia classification and MAE < 5 years for brain age regression) with 10\% of the target data. In many cases, this can be done with no data from the target group whatsoever. Groups formed using race as the attribute often lead to highly imbalanced data, where there is extremely small amounts of data available for Black and Asian populations; our approach is extremely useful here but it will need to be evaluated more rigorously with large sample sizes.
Variations in the data that arise from scanner/acquisition protocol changes in the different cohorts lead to the biggest gaps between using our approach (using 10\% target data) v.s. using all the target data (80\%, denoted by horizontal lines in our figures). This is a large room for improvement in this aspect. This observation is consistent with the findings in the machine learning for health literature~\autocite{howard2021impact, perkonigg2021dynamic, roschewitz2023automatic, chekroud2024illusory}{}, which is a huge barrier for the clinical deployment of machine learning-based medical systems.
%}

%But the trends are consistent with those in several recent studies in healthcare~\autocite{gao2020deep, qiu2020meta, kiyasseh2021clinical, perkonigg2021dynamic, wang2022embracing, he2022meta}{}.

%\newtext{
\textbf{Incorporating domain knowledge is important when building machine learning models for precision medicine.}
The medical imaging community has rapidly adopted convolutional neural networks and other variants of deep networks on neuroimaging data to replace expert-designed imaging features~\autocite{qiu2020development, qiu2022multimodal}{}.
However, this strategy may not be ideal. Medical datasets have comparatively fewer samples than datasets in mainstream computer vision and machine learning problems, this can lead to overfitting while training large networks which needs to be countered with thorough cross-validation Medical images typically exhibit similar anatomical patterns with subtle variations; the information in the data that pertains to the task is relatively sparse compared to, say, visual recognition problems on natural images.
For example, one study shows that a model pre-trained on an ImageNet has little benefits on medical tasks whereas simple and lightweight models perform on par with this sophisticated architecture~\autocite{raghu2019transfusion}{}.
To inject medical domain knowledge into deep networks, a series of investigations has been done in the literature~\autocite{xie2021survey}{}.
For example, one study explicitly extracts hand-crafted radiomic features from chest X-ray images to guide a vision transformer model for learning local fine-grained features~\autocite{han2022radiomics}{}.
Similarly, another work leverages brain morphological change prediction (i.e., anatomical region-of-interest regression) as the auxiliary task to improve Alzheimer’s disease diagnosis~\autocite{ong2023evidence}{}.
%}

%\paragraph{Training using derived imaging features performs on par with raw brain images for classification and regression tasks.}
%With the recent great success of deep learning, the medical imaging community has gradually adopted convolutional neural networks on neuroimaging data to replace expert-designed imaging features~\autocite{qiu2020development, qiu2022multimodal}{}. However, these expert-informed hand-crafted imaging features can be more robust to changes in the image intensity than raw images, e.g., those coming from different image acquisition protocols and scanners, and hence more amenable to domain adaptation.

%\newtext{
Diagnostic models developed in this paper are built using such derived features; they can predict effectively on target groups, in many cases even with no access to the target data. Another benefit of using these derived features is that they can be combined much more easily with information from multiple sources e.g., demographic and clinical variables, genetic factors, and cognitive scores. In principle, this extra information can also be used when deep networks, e.g., convolutional networks, are trained on volumetric magnetic resonance (MR) images (as is indeed done in a number of places in the literature)~\autocite{qiu2022multimodal, xu2022multi, zhou2023transformer}{}. But it is difficult to build architectures where the information in the multi-source features is not overwhelmed by the enormous degree of nuisance variability in the raw MR images.
%}

%\newtext{
There is a growing body of recent work that has re-recognized the value of these hand-crafted features for neuroimaging~\autocite{schulz2020different, dufumier2022representation, wang2023bias, dufumier2024exploring}{}. For example in~\autocite{wang2023bias}{}, machine learning models trained with derived imaging features perform similar to, sometimes even better than, those built using raw brain scans for Alzheimer’s disease, schizophrenia, and autism spectrum disorder diagnosis tasks. It has also been found that, with large-scale neuroimaging data, deep learning algorithms using raw brain images do not improve the accuracy for sex and age prediction compared to models that use such derived features~\autocite{schulz2020different}{}. These existing results, as also our findings in this paper coming from the angle of domain shift and adaptation, indicate that it is difficult to glean representations from neuroimaging data using neural networks that can outperform hand-crafted imaging features in a robust and reproducible way, and there there is a lot of room for improvement in how deep networks are used for such data~\autocite{dufumier2024exploring, dufumier2022representation,geirhos2020shortcut, degrave2021ai}{}.
%}

%\newtext{
\textbf{Effective domain adaptation can improve clinical decision-making and fairness.}
In the broader context of being able to make predictions for parts of the population for which we have few samples, it is an interesting question as to whether we can glean auxiliary information from trained models. For example, we trained diagnostic models on normative tasks such as binary classification of Alzheimer’s disease, and used this model to prospectively diagnose Alzheimer’s disease for subjects with mild cognitive impairment (MCI). This is similar to approaches such as SPARE-AD~\autocite{davatzikos2009longitudinal, davatzikos2011prediction} which build upon RAVENS maps~\autocite{davatzikos2001voxel} to detect progression of the pathology in MCI subjects. We showed that our models are much more effective at predicting such progression. Improvements to domain adaptation also translate to improvements in this auxiliary task.
Similarly, our improved domain adaptation techniques enable us to corroborate existing results in the literature on the association between brain age and various neuropsychological tests. In some cases we find differences with respect to existing literature. For example, our results suggest that there is weak correlation between brain age and cardiovascular features and lifestyle factors. These conclusions again improve using our domain adaptation techniques.
%}

%\newtext{
Machine learning-based diagnostic models can be biased when evaluated on different parts of the population~\autocite{kopal2023end}{}.
There are a number of techniques to mitigate bias, e.g., by careful training~\autocite{wang2023bias}{}, but it has also been found that bias can recur when these models evaluated on newer groups in the population~\autocite{an2022transferring, schrouff2022diagnosing, jiang2023chasing, wang2023robust, mukherjee2022domain, chen2022fairness}{}.
For example, a case study found that the maximum performance gap between age groups increases starkly when deploying a pre-trained skin lesion detection model to a new hospital~\autocite{schrouff2022diagnosing}{}.
We showed that for all three problems (see~\cref{fig:app:fairness}), Alzheimer’s disease diagnosis, schizophrenia diagnosis and brain age prediction, weighted ERM using a small fraction of the target data reduces bias under demographic parity differences (DPD)~\autocite{hardt2016equality, agarwal2019fair} and equalized odds difference (EOD)~\autocite{feldman2015certifying}{}. This is corroborated by other results in the literature~\autocite{creager2021environment, mukherjee2022domain}{} on domain adaptation for mitigating bias.
%}

% In summary, we used a weighted ERM approach to mitigate the detrimental effects of domain shifts in machine learning models derived from brain MRI, and to develop robust and generalizable ML models for classification and prediction of brain diseases. We have derived extensive experimental results that demonstrate that this approach leads to ML models that work better in new patient cohorts, in cohorts having different demographics, and under different image acquisition conditions. We therefore provide a framework not only for higher reproducibility of ML models, but also for increased fairness when ML models are used on under-represented patient populations.
% !TEX root = ../main.tex

\section{Methods}
\label{s:methods}

\subsection{Datasets}
We use 15,363 3D magnetic resonance (MR) images together with their corresponding demographics, clinical variables, genetic factors, and cognitive test scores from two large-scale consortia -- iSTAGING~\autocite{habes2021brain} and  PHENOM~\autocite{satterthwaite2010association, wolf2014amotivation, zhang2015heterogeneity, zhu2016neural, zhuo2016schizophrenia, chand2020two} for Alzheimer’s disease (AD) diagnosis, schizophrenia (SZ) diagnosis and brain age prediction tasks. This data covers a broad population in terms of biological genders, age span, race or ethnicity diversity, scan acquisition protocols and devices, and phenotypes of diseases. \cref{tab:data} summarizes this data.

\paragraph{iSTAGING.}
For AD classification, we use four studies from iSTAGING~\autocite{habes2021brain} dataset including Alzheimer’s Disease Neuroimaging Initiative (ADNI)~\autocite{jack2008alzheimer}{}, Penn Memory Center and Aging Brain Cohort (PENN),  Australian Imaging, Biomarkers and Lifestyle (AIBL)~\autocite{ellis2010addressing}{}, and Open Access Series of Imaging Studies (OASIS)~\autocite{lamontagne2019oasis}{}. Healthy controls and patient groups are the two labels for the AD classification task (see \cref{tab:data}). In the brain chronological age regression experiment, we leverage nine studies from iSTAGING~\autocite{habes2021brain} dataset including Biomarkers of Cognitive Decline Among Normal Individuals (BIOCARD)~\autocite{albert2014cognitive}{}, Baltimore Longitudinal Study of Aging (BLSA)~\autocite{resnick2003longitudinal, armstrong2019predictors}{}, Coronary Artery Risk Development in Young Adults (CARDIA)~\autocite{friedman1988cardia, davidson2000depression}{}, Study of Health in Pomerania (SHIP)~\autocite{volzke2011cohort}{}, Systolic Blood Pressure Intervention Trial (SPRINT)~\autocite{nasrallah2019association}{}, UK Biobank (UKBB)~\autocite{sudlow2015uk}{}, Women’s Health Initiative Memory Study (WHIMS)~\autocite{coker2009postmenopausal}{}, Wisconsin Registry for Alzheimer’s Prevention (WRAP)~\autocite{johnson2018wisconsin}{}, and Action for Health in Diabetes (lookAHEAD)~\autocite{espeland2016brain}{}. Only healthy controls are used for brain age prediction task (see \cref{tab:data}). In addition to imaging features, we use demographic information such as gender, age, race, education level, and smoking status; clinical variables such as diabetes, hypertension, hyperlipidemia, systolic / diastolic blood pressure and body mass index (BMI); genetic factors such as apolipoprotein E (APOE) alleles 2, 3 and 4; and cognitive scores such as mini-mental state exam (MMSE) as features besides the imaging features from the T1-weighted structural MR images in this work. Non-imaging features are sparsely distributed among the participants in the studies.

\paragraph{PHENOM.}
In the SZ diagnosis task, we use scans which are acquired from five different sites around the world in PHENOM~\autocite{satterthwaite2010association, wolf2014amotivation, zhang2015heterogeneity, zhu2016neural, zhuo2016schizophrenia, chand2020two} dataset, namely Penn (United States), China, Munich, Utrecht, and Melbourne. Healthy controls and patient groups are the two diagnosis labels for SCZ classification task (see~\cref{tab:data}). To train, we use demographic information such as gender, age, race, marital status, employment status, and handedness; and cognitive scores such as full-scale intelligence quotient (FIQ), verbal intelligence quotient (VIQ), and performance intelligence quotient (PIQ) besides features computed from T1-weighted structural MR images. Again, non-imaging features are sparsely populated in these studies.

\paragraph{Inclusion criteria.}
All included subjects have passed quality control (QC) where they were visually examined by a radiologist to screen the pre-processed scans; distorted images were dropped. We only use the baseline (initial time point) scans from each study. All follow-up sessions are excluded; this ensures that there is no data leakage and the same participant is not included in the train and test sets. For AD cohorts, we select stable cognitively normal (CN) and AD patients based on each participant's longitudinal diagnosis status. We only include participants who were diagnosed as CN or AD at the baseline and stayed stable during the follow-up sessions. For mild cognitive impairment (MCI) cohorts, we select progressive MCI (pMCI), stable MCI (sMCI) and normal MCI (nMCI) patients based on each participant's longitudinal diagnosis status. We define MCIs as follows: pMCIs are subjects who were diagnosed as MCI at baseline and progressed to AD during the three years following the first visit; sMCIs are subjects who were diagnosed as MCI at baseline and remained MCI during the three years following the first visit; nMCIs are subjects who were diagnosed as MCI at baseline and reverted back to CN during the three years following the first visit.

\subsection{Data pre-processing pipeline}
\label{s:pipeline}

\paragraph{Pre-processing images to obtain imaging features.}
We pre-process T1-weighted MR images using a standard pipeline as follows. The scans are first bias-field corrected~\autocite{tustison2010n4itk} and skull-stripped with a multi-atlas algorithm~\autocite{doshi2013multi}{}. Then we use a multi-atlas label fusion segmentation method~\autocite{doshi2016muse} to obtain the anatomical region-of-interest (ROI) masks including 119 grey matter ROIs, 20 white matter ROIs, and 6 ventricle ROIs of the brain. 145 ROI volumes are calculated based on the corresponding extracted brain masks. We further segment white matter hyperintensities (WMH) by applying a deep learning-based algorithm~\autocite{doshi2019deepmrseg} on the fluid-attenuated inversion recovery (FLAIR) and T1-weighted images. The white matter lesion (WML) volumes are obtained by summing up the WMH mask voxels.

\paragraph{Pre-processing methodology.}
We pre-process and normalize imaging measures such as ROI volumes and WML volumes, demographic, clinical and genetic features and cognitive scores. Since many features, e.g. clinical variables, genetic factors, and cognitive scores, are sparse, we need to impute these missing values. For continuous features, we first impute missing values with the median of each feature and then normalize the features to have zero mean and unit variance. We apply quantile normalization to skewed distributions. For discrete features, we introduce an ``unknown'' category that indicates missing values. For each feature with missing values, we introduce an additional feature to indicate whether the particular entry was missing; this helps preserve the information of the absence. We did not remove batch or site effects from the pooled data, i.e., we did not use any harmonization~\autocite{pomponio2020harmonization, wang2021harmonization}{}.

\subsection{Training and evaluation methodology}
\label{s:models}

\paragraph{Creating the train and test sets.}
We use nested five-fold cross-validation for all our experiments. Consider the situation when we are training and evaluating on the same group, i.e., we split the data into five equal-sized folds (stratified by labels), use four for training and validation (80\% data) and the fifth for testing (20\% data). All hyper-parameter tuning is performed using a further five-fold cross-validation within the 80\% data. This way, the remaining 20\% data forms a completely independent test set (which is not used for choosing any hyper-parameters) and can be used for reporting the area under the receiver operating curve (AUC) or the mean absolute error (MAE). When this procedure is repeated 5 times, each time for a different test fold, we can calculate the mean and standard deviation of the AUC or the MAE on the test set.

We constructed a similar nested cross-validation procedure for situations where we train and evaluate on different groups. We again split target data into five equal-sized folds (stratified by labels). Suppose we have access to 20\% data from the target group. The train and validation sets together therefore consist of all data from the source group and 20\% data from the target group; the remainder 80\% data from the target group is the test set. We further sub-divide the train set (i.e., the 100\% source data and the 20\% target data) into 5 equal parts (stratified by labels, and with equal amount of target data in each fold) to perform cross-validation and model selection. We perform the search for the hyper-parameter $\a = k/(k+1)$ over $k \in \{1,2,\dots,10\}$; the search grid therefore becomes finer as the value of $\a$ goes closer to 1. Again, we can report the mean and standard deviation of AUC or MAE across the entire data from the target group after repeating this procedure for each of the five test folds.  When the amount of data used from the target group is 0\%, we report the AUC and MAE on all the data from the target group without any test folds.

Even when we have access to only 10\% of the data from the target group, we still create five equal-sized folds from the target data. In other words, our train and validation set consists of 100\% data from the source group, 10\% data from the target group and the test set consists of 80\% data from the target group. This was done for convenience: there are hundreds of experiments in this paper across complicated multi-source datasets, each of which uses five-fold nested cross-validation, and is conducted using multiple models such as ensembles and neural networks---book-keeping all this information gets unwieldy quickly.

This evaluation methodology is very expensive computationally, but it is rigorous and allows us to glean statistically precise conclusions. The experiments and data analysis in this study required an estimated 10,000 GPU hours.

\paragraph{Training the neural network and the ensemble.}
We use three-layer feed-forward neural network (multi-layer perceptron) with rectified linear unit (ReLU) activations, dropout~\autocite{srivastava2014dropout} and batch normalization~\autocite{ioffe2015batch} after each layer; there is a skip-connection~\autocite{he2016deep} that connects the first and the penultimate layer. Pre-processed categorical features are first passed to their corresponding embedding layers individually and all these embeddings are concatenated with the pre-processed numerical features.

Ensembling techniques such as bagging~\autocite{breiman1996bagging}{}, boosting~\autocite{bartlett1998boosting} and stacking~\autocite{wolpert1992stacked} are well-established techniques in machine learning. We used AutoGluon~\autocite{erickson2020autogluon} which is a software framework that implements these techniques efficiently, along with parallel hyper-parameter search and model selection across CPUs and multiple GPUs. AutoGluon also implements a novel ensembling strategy called multi-layer stacking and repeated $k$-fold bagging~\autocite{parmanto1996reducing}{}. This is quite useful for the kinds of data studied in this paper and we discuss this strategy briefly next. In multi-layer stacking, we first train models such as neural networks, $k$-nearest neighbors, random forests, CatBoost boosted trees~\autocite{prokhorenkova2018catboost}{}, and LightGBM trees~\autocite{ke2017lightgbm} separately. Predictions from these base models, concatenated with features in the original data serve as input features to a second ``layer'' of stacked models; this is inspired from skip-connections in deep learning~\autocite{he2016deep}{}. A multi-layer stacking approach can grow an ensemble in this fashion by treating the stacked models as the base models for the next layer; in practice to mitigate over-fitting, stacking is done only for two levels. To make predictions, the highest-level stacked models are combined using ensemble selection~\autocite{caruana2004ensemble}{}; this has been found to be resilient to over-fitting when combining high-capacity models. To curb over-fitting and to avoid covariate shift at inference time, while fitting each layer of the stacked model, the training set is randomly split into $k$ folds (stratified by labels) and $k$ different models fitted on $(k-1)$ of these folds are used to obtain out-of-fold (OOF) predictions for each sample. This $k$-fold bagging can also be repeated several times to average the OOF predictions and further mitigate potential over-fitting within out-of-fold predictions.

\subsection{Baseline methods comparison}
\label{s:baselines}
we compare the proposed method with 8 representative state-of-the-art domain adaptation/generalization approaches covering different streams of techniques including invariant learning, adversarial learning, distribution alignment, domain translation, and source-free/test-time adaptation, as follows
\begin{itemize}[noitemsep]
\item IRM: Invariant risk minimization (IRM)~\autocite{arjovsky2019invariant} learns an invariant
representation that the optimal predictor using this representation is simultaneously optimal in all environments.
\item DANN: Domain-adversarial neural network (DANN)~\autocite{ganin2016domain} leverages a discriminator to extract domain-irrelevant features by jointly training on source and target datasets.
\item JAN: Joint adaptation network (JAN)~\autocite{long2017deep} aligns feature distribution from different domains based on the joint maximum mean discrepancy metric.
\item JDOT: Joint distribution optimal transport (JDOT)~\autocite{courty2017joint} performs domain translation/mapping by using distributional optimal transportation.
\item TENT: Test entropy minimization (TENT)~\autocite{wang2020tent} adapts a pre-trained model by minimizing the entropy of the model predictions on test data.
\item SHOT: Source hypothesis transfer (SHOT)~\autocite{liang2020we} maximizes the mutual information between intermediate feature representations and model predictions while augments the model with self-supervised pseudo-labeling.
\item DALN: Discriminator-free adversarial learning network (DALN)~\autocite{chen2022reusing} performs domain alignment without using a discriminator by leveraging nuclear-norm Wasserstein discrepancy regularization.
\item TAST: Test-time adaptation via self-training (TAST)~\autocite{jang2022test} improves self-supervised pseudo-labeling by using prototype-based classification with nearest neighbor information.
\end{itemize}
In the experiments, we evaluate above 8 domain adaptation/generalization baselines on Alzheimer’s disease and schizophrenia classification separately under the exact same experimental setups as our method.
%}

\subsection{Learning the maximum mean discrepancy statistic}
\label{s:two_sample_test}

Consider the case when we have two sets of samples $D_p = \{x_i: x_i \sim p\}_{i=1}^n$ and $D_q = \{x'_i: x'_i \sim q\}_{i=1}^n$ from two probability distributions $p$ and $q$ respectively. A two-sample test relies on finding a function (called the ``statistic'') which is large on samples drawn from $p$ and small on samples draw from $q$. The maximum mean discrepancy (MMD)~\autocite{gretton2012kernel} statistic takes the difference between the mean function value on the two sets of samples, and determines the two sample sets as coming from different distributions if this maximum is large. The MMD is defined as
\beq{
    \text{MMD}[\FF, p, q] = \sup_{f \in \FF} \rbr{\E_x \sbr{f(x)} - \E_{x'} \sbr{f(x')}},
    \label{eq:mmd}
}
where $\FF$ is a class of functions with appropriate regularity properties, e.g., that it has uniform convergence~\autocite{scholkopf2002learning}{}. A classical example of such a class of functions is the reproducing kernel Hilbert space which has the property (called the reproducing kernel property) that for any input $x$, there exists a function $\phi_x \in \FF$ (called the canonical feature map) such that the evaluation of $f \in \FF$ at the input $x$ can be written as an inner product in the reproducing kernel Hilbert space (RKHS):
\[
    \reals \ni f(x) = \inner{f}{\phi_x}_\FF.
\]
Note that $\phi_x$ is a function. Observe that using the same property above, we can write $\phi_x(x') = \inner{\phi_x}{\phi_{x'}}_\FF \equiv k(x, x') \geq 0$; here the quantity $k(x, x')$ is called the kernel of the RKHS and measures the similarity between the two inputs $x$ and $x'$. An unbiased estimate of~\cref{eq:mmd} using samples is given by
\beq{
    \text{MMD}_u^2[\FF, p, q] = \f{1}{(n^2 -n)} \sum_{i,j=1, i\neq j}^n \rbr{k(x_i, x_j) + k(x_i',x_j') - k(x_i, x_j') - k(x_j, x_i')}.
    \label{eq:mmd_u}
}

Neural networks are universal function approximators and possess uniform convergence properties~\autocite{cybenko1989approximation}{}. Therefore, we can fruitfully set the class $\FF$ to be the class of functions parameterized by a multi-layer perceptron (MLP)~\autocite{sriperumbudur2010hilbert,ben2006analysis}{}. We will next describe how we use an MLP to estimate the differences between the probability distribution of the covariates of subjects belonging to different groups in the Alzheimer’s disease data. We build an MLP that takes as input the covariates $x$ and has four different types of outputs corresponding to each of the four attributes (sex: male vs.\@ female, age: < 65 vs.\@ 65--70 vs.\@ 70-75 vs.\@ etc., race: White vs.\@ Black vs.\@ Asian, and clinical study: ADNI-1 vs.\@ ADNI-2/3 vs.\@ PENN vs. AIBL). We fit one MLP to predict each of these groups using four different output layers (for sex we have a binary classification problem, for age we have a 5-way problem, etc.) and treat its learned features as the canonical feature map $\phi_x$ of the input $x$. The problem of training the MLP to minimize of errors incurred while predicting the correct group for each attribute for each subject, is equivalent to the maximization over the functions $f \in \FF$ in~\cref{eq:mmd}. The pairwise MMD statistic between groups for each attribute is calculated from these learned features using~\cref{eq:mmd_u} and reported in a summarized form in~\cref{fig:dendrogram_ad,fig:app:dendrogram_scz,fig:app:dendrogram_age} with details provided in~\cref{fig:app:distance_ad,fig:app:distance_scz,fig:app:distance_age}. We can also examine the relative difference between the MMD statistic corresponding to different pairs of groups and attributes, e.g., the MMD statistic between males and females in~\cref{fig:app:distance_ad,fig:app:dendrogram_ad} is 0.17 while the statistic between ADNI-1 and ADNI-2/3 is 0.26; there is a larger discrepancy between the probability distributions in the latter case than the former. Note that if we had fitted different MLPs independently for each of the attributes, then we would not have been able to examine the relative differences because the learned feature maps would be different for different attributes.

\subsection{Building models that also have access to data from the target group improves performance}
\label{s:theory}

Consider a dataset $D = \cbr{(x_i, y_i)_{i=1}^n}$ with $n$ samples where $x_i$ are the input features. For the purposes of this argument we will think of the ground-truth $y_i \in \cbr{0,1}$ as being binary but all the claims in the sequel can be formalized analogously for multi-class classification and regression problems. We will assume that each datum is drawn from a probability distribution $(x_i, y_i) \sim P$. A machine learning model with weights $\th \in \reals^p$ is a model $p_\th(y \mid x)$ of the probability that an input $x$ belongs to a category $y \in \cbr{0,1}$. We would like this model to generalize to new data outside the training set, i.e., we would like to find a model
\beq{
    \th^* = \argmin_\th R(\th) := -\E_{(x,y) \sim P} \sbr{\log p_\th(y \mid x)},
    \label{eq:population_risk}
}
which minimizes the ``population risk'' $R(\th)$ which is the average negative log-likelihood over all samples $(x,y) \sim P$. The population risk is also called the test risk, or the test loss. In practice, we do not know the entire distribution of data $P$ but only have access to a training set $D$ with samples from $P$. Our goal is to find a best approximation for the weights $\th^*$, which is typically done by minimizing the so-called empirical risk (also called training loss)
\beq{
    \hat{\th} = \argmin_\th \hat R(\th) + \Om(\th) := -\f 1 n \sum_{i=1}^n \log p_\th(y_i \mid x_i) + \Om(\th),
    \label{eq:erm}
}
where $\Om(\th)$ is a regularization term, e.g., $\Om(\th) = \norm{\th}_2^2/2$, that avoids over-fitting to the training dataset. The objective is usually minimized by the stochastic gradient descent~\autocite{bottou2010large} algorithm. The discrepancy between the two, the objective~\cref{eq:population_risk} that we wish to minimize and the objective in~\cref{eq:erm} that we actually minimize, is measured using classical inequalities in machine learning theory:
\beq{
        R(\th) \leq \hat R(\th) + c \sqrt{\f{V - \log \delta}{n}},
        \label{eq:vc}
}
which hold for all $\th$ (in particular $\hat \th$ in~\cref{eq:erm}) and any $\delta \in (0,1)$ with probability at least $1-\delta$ over independent draws of the samples in the training dataset; here $V$ denotes the Vapnik-Chervonenkis (VC) dimension~\autocite{scholkopf2002learning} and $c$ is a constant. Roughly speaking, the VC-dimension measures the complexity of the function class (e.g., the neural network) parameterized by the weights. The above inequality says that larger the complexity of the function class larger the VC-dimension and larger the number of samples $n$ that we need in order to ensure that the population/test risk $R(\th)$ is small, provided the empirical risk, or the training loss, $\hat R(\th)$ is also small. One can think of the objective in~\cref{eq:erm} as corresponding to the right-hand side of~\cref{eq:vc}: the training cross-entropy loss is equal to the empirical risk $\hat R(\th)$ and the regularization term $\Om(\th)$ is a proxy for the second complexity term.

\paragraph{Domain generalization can be poor if the source and target groups have very different distributions.}
If the training data and the test data are not from the same probability distribution $P$, then we should not expect the weights $\hat \th$ obtained using the training data to predict accurately on the test data, i.e., have a low test loss. We can quantify this deterioration using inequalities similar to~\cref{eq:vc} as follows. Suppose the training dataset $D$ consists of samples from a source group $P_s$ and we seek to minimize the population risk on data drawn from a different target group with a probability distribution $P_t \neq P_s$. In the context of this study, these groups represent different sex, age, race or clinical studies. For any $\delta \in (0,1)$ with probability at least $1-\delta$ over independent draws of the training dataset, we have
\beq{
    R_t(\th) \leq R_s(\th) + c \sqrt{\f{V - \log \delta}{n}} + \f{d(P_s, P_t)}{2} + \l
    \label{eq:domain_adaptation_bound}
}
for all weights $\th$; here $c$ is a constant, $R_s(\th)$ is the population risk on the source group, $R_t(\th)$ is the population risk on the target group~\autocite{ben2010theory}{} and
\[
    \l = \min_\th R_s(\th) + R_t(\th)
\]
is the combined risk of a model that minimizes the average population risk of the two groups. The term $d(P_s, P_t)$ measures the difference between the two probability distributions $P_s$ and $P_t$. The key point is that if $P_t \neq P_s$, both the terms $d(P_s, P_t)$ and $\l$ are non-zero. And therefore, even if the weights $\th$ were to obtain a good population risk $R_s(\th)$ (which, as we discussed above, is not a given), there would still be a gap between the population risk $R_t(\th)$ due to the discrepancy between $P_s$ and $P_t$. If the distributions of the source and target groups $P_s$ and $P_t$ are very different, then a model trained only on source data will perform poorly on the target group~\autocite{rameshModelZooGrowing2022,ben2003exploiting}{}. This is a fundamental hurdle in clinical applications where data is heterogeneous~\autocite{moyer2021harmonization}{}.

\paragraph{Weighted-ERM on a combined dataset from the source and target groups.}
A natural technique to mitigate the deterioration of accuracy due to the change in data distribution between the source and target groups is to use some data from the target group in addition to that from the source group in the training set while fitting the weights $\th$. Assume that the distribution $P$ is a mixture of two groups $P_s$ and $P_t$ with $P_s \neq P_t$. Given $m$ samples from the source sub-population $P_s$ and $n$ samples from the target sub-population $P_t$, we can construct a combined training set $\{(x_i, y_i)_{i=1}^{m+n}\}$. We are interested in settings where we do not have access to a large number of samples from the target group, i.e., $n \ll m$. Consider a weighted empirical risk minimization (ERM) objective
\beq{
    \aed{
    \hat \th = \argmin_\th \hat R(\th) &:= -\f{1-\alpha}{m} \sum_{i=1}^{m} \log p_\th(y_i \mid x_i) - \f{\alpha}{n} \sum_{i=m+1}^{m+n} \log p_\th(y_i \mid x_i) + \Om(\th)\\
    &= (1-\a) \hat R_s(\th) + \a \hat R_t(\th) + \Om(\th),
    }
    \label{eq:weighted_erm}
}
where we have denoted the empirical source and target risks by $\hat R_s(\th)$ and $\hat R_t(\th)$ respectively. The hyper-parameter $\a \in [0,1]$ controls the relative weight of these risks in finding the optimal weights $\hat \th$. To understand its effect, observe that if $\a = 0$, i.e., we do not use any data from the target group, we should not expect the weights fitted using data from the source to predict well on test data from the target group; the consequent deterioration in this setting was discussed in~\cref{eq:domain_adaptation_bound}. If we set $\a = 1$, i.e., we do not use any data from source group, then the small number of samples $n$ that we have from the target group will lead to larger right-hand side in~\cref{eq:vc} and thereby a poor predictor on the test data from the target. A naive strategy would set $\a=1/2$, i.e., weigh the two sub-populations equally in the objective, but it turns out that one often do better. Formally, for any $\delta \in (0,1)$ with probability at least $1-\delta$ over independent draws of training samples from $P_s$ and $P_t$, it can be shown that~\autocite{ben2010theory}{}
\beq{
    R_t(\hat \th) \leq R_t(\th^*_t) + 4 \sqrt{ \rbr{\f{\a^2}{n} + \f{(1-\a)^2}{m}} \rbr{V - \log \delta}} + 2 (1-\alpha) d(P_s, P_t),
    \label{eq:weighted_erm_bound}
}
where $\th^*_t$ is the optimal, i.e., it minimizes~\cref{eq:population_risk} for the target population. The third term again is a measure of the difference between the two populations $P_s$ and $P_t$. This theorem suggests that if we choose an appropriate value of $\a$ that minimizes the entire right-hand side, then we can expect to draw maximal utility from the source data to predict accurately on the target population. For example, observe that if $d(P_s, P_t)$ is very large, then the right-hand side is small when $\a \approx 1$, i.e., when the weighted-ERM objective neglects the source data (because it is quite dissimilar to the target data). Guidelines for picking this hyper-parameter can be computed~\autocite{de2022value}{}: if the number of samples from the target population is larger than a threshold $4 (V - \log \delta)/(d(P_s, P_t))^2$, then $\a = 1$ minimizes the right-hand side. In practice, it is difficult to estimate, both the VC dimension $V$ and the difference between the two probability distributions $P_s$ and $P_t$ given by $d(P_s, P_t)$. But this hyper-parameter $\a$ can be chosen in practice using cross-validation.

\paragraph{Relationship of weighted-ERM to other transfer learning techniques.}
The weighted-ERM procedure that we have discussed above can be thought of as an optimal mechanism to adapt a model trained on a source data to the target data as follows. Formally, in supervised learning using a dataset $D = \{(x_i,y_i)\}_{i=1}^n$, the prediction on a test datum $x$ is made by computing:
\[
    \hat y = \argmax_y p(y \mid x, D).
\]
If the class of functions that we search over to find one candidate $p(\cdot \mid x, D)$ (say, all possible linear predictors) is large, then this search is a computationally expensive endeavor at inference time. The ``weights'' $\hat \th = \argmin \hat R(\th)$ of a parametric model provide a classical way to work around this issue; we can think of the weights as a statistic of the dataset $D$, i.e., $\hat \th \equiv \hat \th(D)$. This statistic is sufficient~\autocite{keener2010theoretical} to predict on samples within the training set $D$. Using this for predicting on the test set by setting
\[
    p(y \mid x, D) \equiv p_{\hat \th}(y \mid x)
\]
entails a loss of generality and causes the gap between population risk and empirical risk in~\cref{eq:vc} but these weights conveniently separate the training and inference phases. This same argument holds for transfer learning where the weights $\hat \th$ are learned using a source dataset and adapted, e.g., using mini-batch updates, to new data from the target dataset. Explicitly searching for the weights using both the source and target data, like we do in weighted-ERM in~\cref{eq:weighted_erm} is at least as good as adapting the weights trained on the source data. This is an instance of the data processing inequality~\autocite{keener2010theoretical}{}. In other words, we can prove that the population risk of the solution obtained from~\cref{eq:weighted_erm} is at least as good as the one obtained using transfer learning techniques. There are a number of theoretical and empirical works in the machine learning literature which have argued this point for transfer learning~\autocite{gao2022deep,gao2020information}{}, meta-learning and few-shot learning~\autocite{dhillon2019a,fakoor2019meta} and continual learning~\autocite{rameshModelZooGrowing2022}{}.

\section*{Acknowledgments}
This work was supported by the National Institute on Aging (RF1AG054409 and U01AG068057), the National Institute of Neurological Disorders and Stroke (U24NS130411), the National Institute of Mental Health (R01MH112070), the National Institute on Drug Abuse (75N95019C00022), the National Science Foundation (IIS-2145164) and cloud computing credits from Amazon Web Services.
The data used in this study are part of the iSTAGING consortium~\autocite{pomponio2020harmonization, habes2021brain}{} for brain aging and the PHENOM consortium~\autocite{chand2020two}{} for psychosis.

\section*{Author contributions}
%All authors designed the research, analyzed the results, wrote and edited the manuscript. R.W. conducted the experiments.
R.W., P.C. and C.D. conceptualized the research. R.W. conducted the experiments. R.W., P.C. and C.D. wrote the manuscript. All authors edited and agreed to the published version of the manuscript.

\section*{Competing interests}
The authors declare no competing interests.

\section*{Data availability}
Data that support the findings of this study are available from their respective institutions. Restrictions apply to the availability of these data, which were used under license for the current study. Data may be made available by the authors upon request and with permission.

\section*{Code availability}
The code used to train and evaluate Weighted-ERM is available on GitHub at \href{https://github.com/rongguangw/weightedERM}{https://github.com/rongguangw/weightedERM}.
%Source code for all experiments is available \href{https://drive.google.com/file/d/1fMiqO-4xW8GtKvmvYMqEVomd20y6sxCl}{here}.

\printbibliography[title=Bibliography]

\clearpage
% !TEX root = ../main.tex

\renewcommand\thesection{S.\arabic{section}}
\renewcommand\thefigure{S.\arabic{figure}}
\renewcommand\thetable{S.\arabic{table}}
\setcounter{figure}{0}
\setcounter{section}{0}
\setcounter{table}{0}
\renewcommand{\figurename}{Figure}
\renewcommand{\tablename}{Table}

\begin{appendix}

\section{Additional Tables and Figures}
\label{app:additional_and_figures}

\begin{table}[htpb]
\caption{\textbf{Summary of the data from the iSTAGING consortium used for early diagnosis of stable and progressive mild cognitive impairment.}}
\label{tab:app:mci}
\begin{center}
\begin{footnotesize}

\resizebox{0.85\linewidth}{!}{
\begin{tabular}{p{0.15\linewidth}rrrrrr}
\toprule
\textbf{Mild Cognitive} && ADNI-1 & ADNI-2/3 & PENN & AIBL & Total \\
\textbf{Impairment} && (44.60\%) & (47.69\%) & (2.31\%) & (5.40\%)  \\
\cmidrule{3-7}
% & Subjects & 289 & 309 & 15 & 35 & 648  \\
Subjects\\
& Progressive MCI & 197 & 113 & 5 & 12 & 327  \\
& Stable MCI & 79 & 151 & 10 & 11 & 251  \\
& Normal MCI & 13 & 45 & - & 12 & 70  \\
Sex (\%)\\
& Female & 15.90 & 21.30 & 1.08 & 2.01 & 40.28 \\
& Male & 28.70 & 26.39 & 1.23 & 3.40 & 59.72 \\
Age (\%, years)\\
& 0--65 & 5.86 & 10.80 & 0.62 & 0.15 & 17.44 \\
& 65--70 & 5.86 & 11.27 & 0.46 & 1.39 & 18.98 \\
& 70--75 & 11.57 & 11.73 & 0.46 & 1.85 & 25.62 \\
& 75--80 & 10.19 & 11.73 & 0.62 & 1.39 & 21.60 \\
& > 80 & 11.11 & 4.48 & 0.15 & 0.62 & 16.36 \\
Race (\%)\\
& White & 42.75 & 43.83 & 2.16 & 2.16 & 90.90 \\
& Black & 0.77 & 0.93 & 0.15 & - & 1.85 \\
& Asian & 1.08 & 0.62 & - & - & 1.70 \\
\bottomrule\\
\end{tabular}
}
\end{footnotesize}
\end{center}
\end{table}

% \subsection{Even if machine learning models can predict very accurately on the groups that they were trained on, they do not generalize sufficiently well to data from other groups}
% \label{s:app:generalization_to_groups}

% AD: 0.949 ± 0.016 (Female), 0.928 ± 0.010 (Male); 0.923 ± 0.044 (<65), 0.927 ± 0.021 (65-70), 0.952 ± 0.017 (70-75), 0.939 ± 0.024 (75-80), 0.915 ± 0.024 (>80); 0.936 ± 0.010 (White), 0.946 ± 0.033 (Black), 0.933 ± 0.133 (Asian); 0.949 ± 0.024 (ADNI-1), 0.956 ± 0.016 (ADNI-2/3), 0.969 ± 0.012 (PENN), 0.956 ± 0.035 (AIBL).
% %
% SCZ: 0.790 ± 0.044 (Female), 0.787 ± 0.014 (Male); 0.753 ± 0.039 (<25), 0.816 ± 0.037 (25-30), 0.822 ± 0.042 (30-35), 0.800 ± 0.029 (>35); 0.706 ± 0.075 (Native), 0.734 ± 0.127 (Asian); 0.748 ± 0.062 (Penn), 0.877 ± 0.073 (China), 0.741 ± 0.032 (Munich), 0.771 ± 0.016 (Utrecht), 0.888 ± 0.042 (Melbourne).
% %
% Age: 4.35 ± 0.14 (Female), 4.58 ± 0.12 (Male); 4.20 ± 0.09 (White), 4.89 ± 0.39 (Black), 4.53 ± 0.62 (Asian); 5.14 ± 0.75 (BIOCARD), 5.14 ± 0.71 (BLSA-1.5T), 5.58 ± 0.42 (BLSA-3T), 4.09 ± 0.20 (CARDIA), 4.88 ± 0.23 (SHIP), 4.65 ± 0.34 (SPRINT), 4.17 ± 0.20 (UKBIOBANK), 3.05 ± 0.12 (WHIMS), 4.17 ± 0.32 (WRAP), 4.98 ± 0.50 (LookAHEAD).

\begin{figure}
\centering
\begin{subfigure}[b]{0.35\linewidth}
\centering
\includegraphics[width=0.2\linewidth]{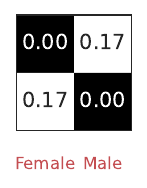}
\includegraphics[width=0.5\linewidth]{ad_dist_age}\\
\includegraphics[width=0.3\linewidth]{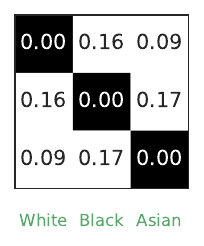}
\includegraphics[width=0.5\linewidth]{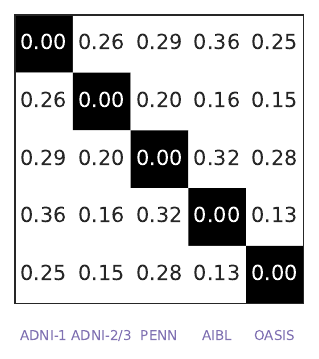}
\caption{}
\label{fig:app:distance_ad}
\end{subfigure}
\begin{subfigure}[b]{0.35\linewidth}
\centering
\includegraphics[width=\linewidth]{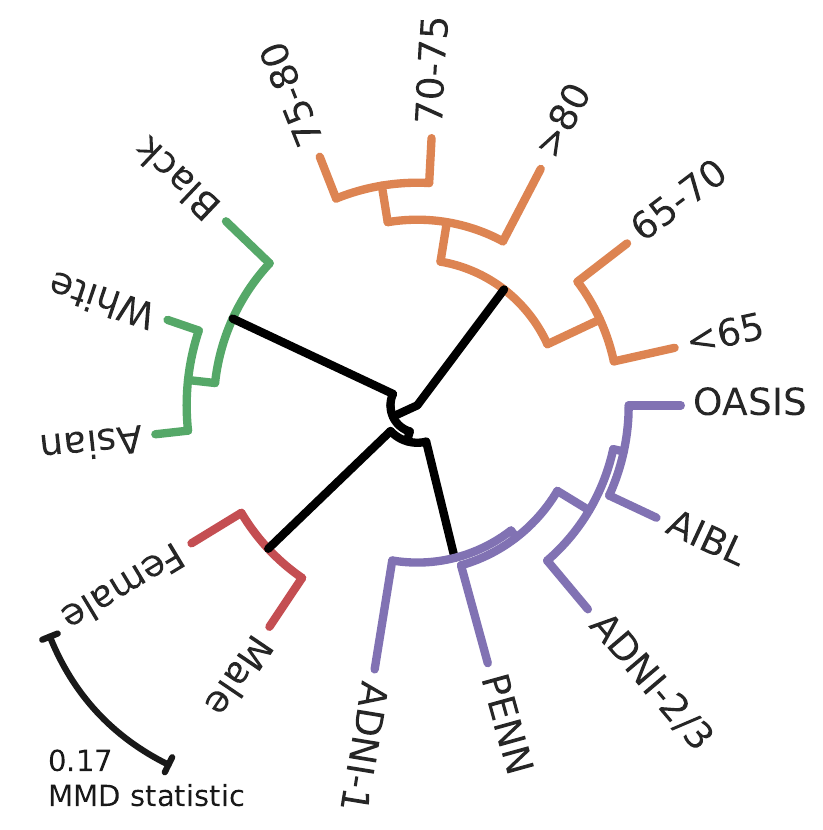}
\caption{}
\label{fig:app:dendrogram_ad}
\end{subfigure}

\begin{subfigure}[b]{\linewidth}
\centering
\includegraphics[width=0.35\linewidth]{ad_nn_sex}
\includegraphics[width=0.35\linewidth]{ad_nn_age}\\
\includegraphics[width=0.35\linewidth]{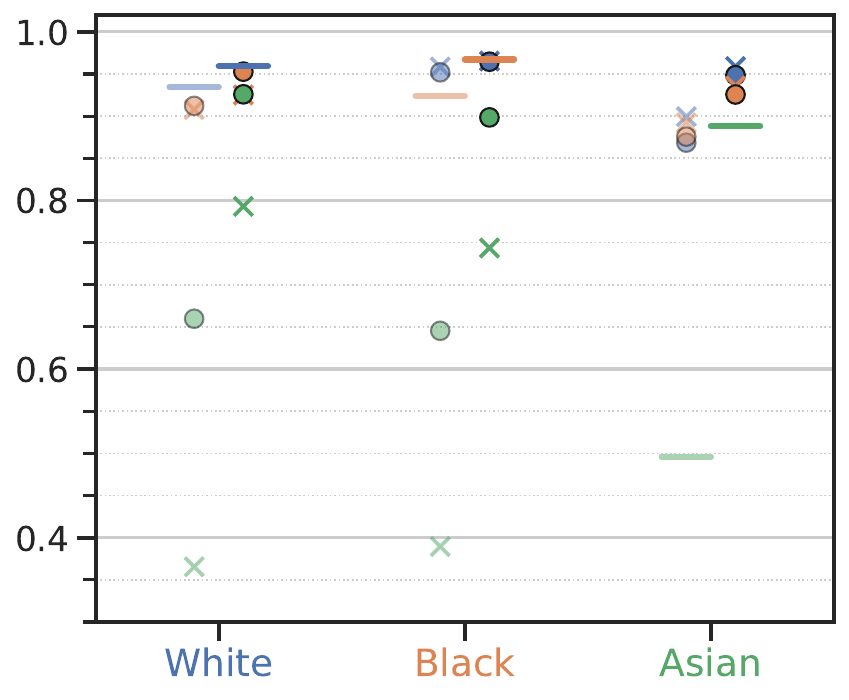}
\includegraphics[width=0.35\linewidth]{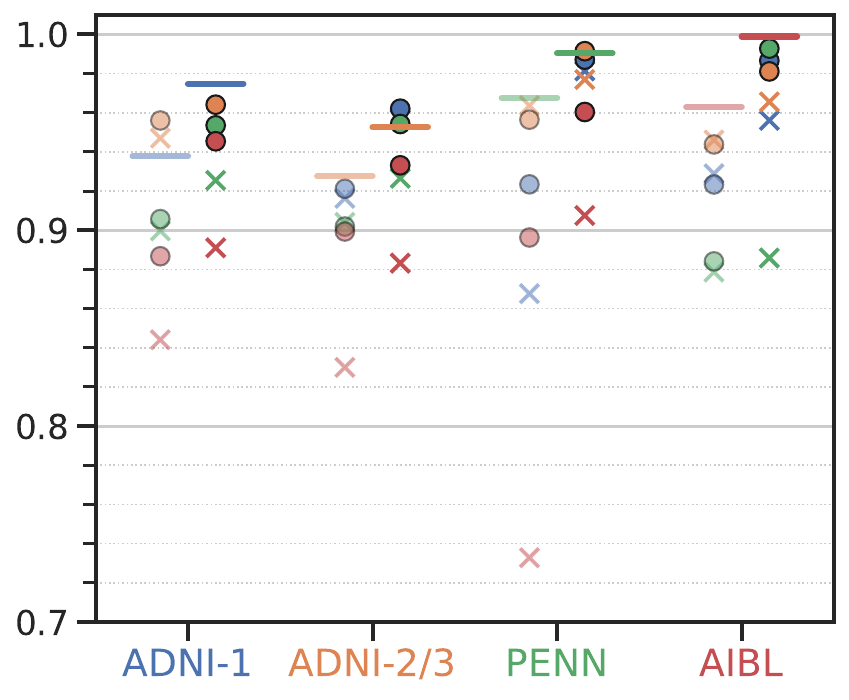}\\
\includegraphics[width=0.8\linewidth]{nn_legend}
\caption{}
\label{fig:app:ad_nn}
\end{subfigure}
\caption{
\textbf{Discrepancy in the data and AUC of diagnostic models of Alzheimer’s disease on different groups in the population.}
\textbf{(a-b)} Distance between leaves of this dendrogram in \textbf{(b)} indicates the pairwise MMD statistic in \textbf{(a)} between learned features of pairs of groups, e.g., distributional discrepancy between Male-Female groups is 0.17, while the distributional discrepancy between < 65 years and > 80 years, or between ADNI-1 and ADNI-2/3, is larger (0.42 and 0.26 respectively).
\textbf{(c)} Average AUC of Alzheimer’s disease classification computed using five-fold nested cross-validation. We trained a machine learning model, either a deep neural network (translucent markers) or an ensemble using boosting, bagging and stacking (bold markers), using data from different source groups (different colors) and evaluated this model (cross marks) on data from different target groups (X-axis); circles denote model fitted using our $\a$-weighted empirical risk minimization (ERM) procedure with access to 10\% data from the target group; horizontal lines denote models that are directly trained on the target group using 80\% of data (the rest for testing). All models use data from multiple sources, namely structural measures, demographic, clinical variables, genetic factors, and cognitive scores. %In general, (i) the AUC of ensemble models is higher than that of the neural network in all cases ($p$ < 0.01), (ii) AUC of a model trained on a source group remains remarkably high when evaluated on the target group (crosses), (iii) in most cases, it further improves when one has access to a small fraction of data from the target group (circles are higher than crosses), and (iv) often times even beyond the AUC of a model trained only on the target group (circles above the horizontal lines).
}
\label{fig:app:dendrogram_nn_ensemble_ad}
\end{figure}

%\rw{fix~\cref{fig:app:dendrogram_nn_ensemble_ad}}

\begin{figure}
\centering
\begin{subfigure}[b]{0.4\linewidth}
\centering
\includegraphics[width=0.2\linewidth]{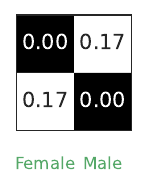}
\includegraphics[width=0.4\linewidth]{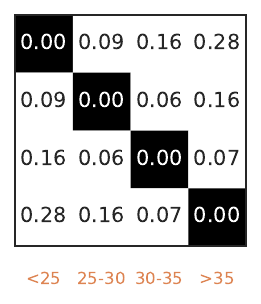}\\
\includegraphics[width=0.2\linewidth]{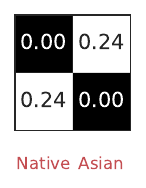}
\includegraphics[width=0.5\linewidth]{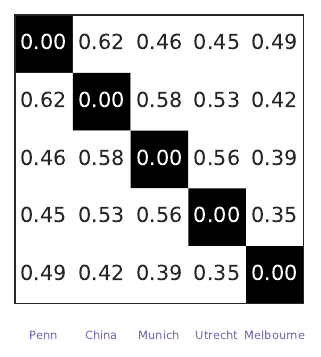}
\caption{}
\label{fig:app:distance_scz}
\end{subfigure}
\begin{subfigure}[b]{0.4\linewidth}
\centering
\includegraphics[width=\linewidth]{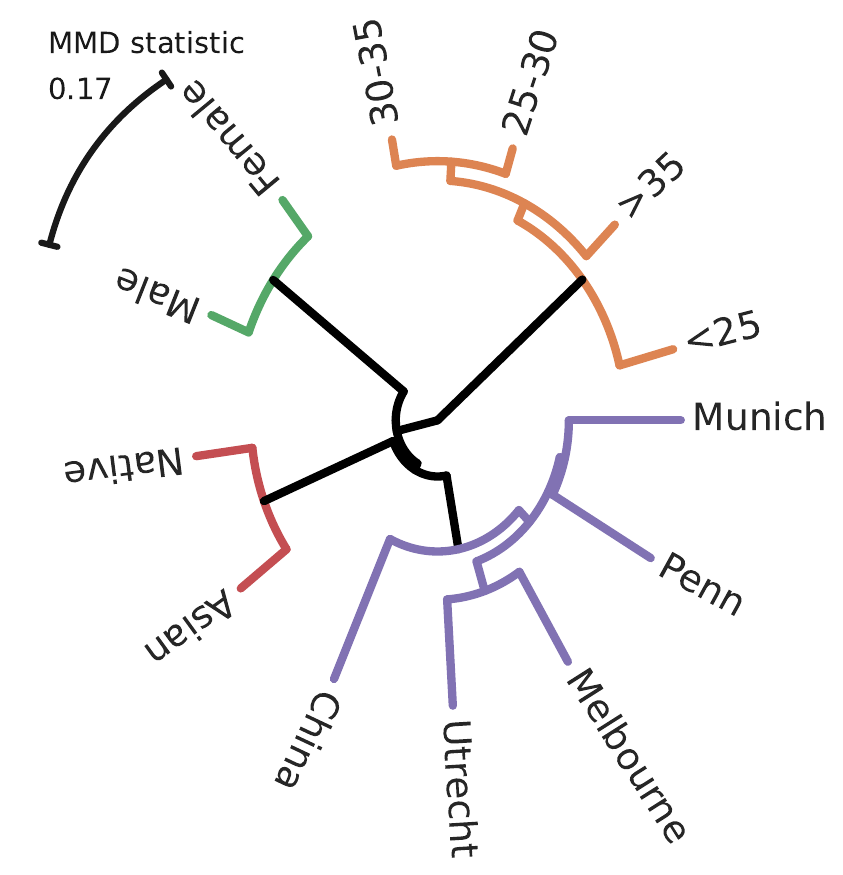}
\caption{}
\label{fig:app:dendrogram_scz}
\end{subfigure}

\begin{subfigure}[b]{\linewidth}
\centering
\includegraphics[width=0.35\linewidth]{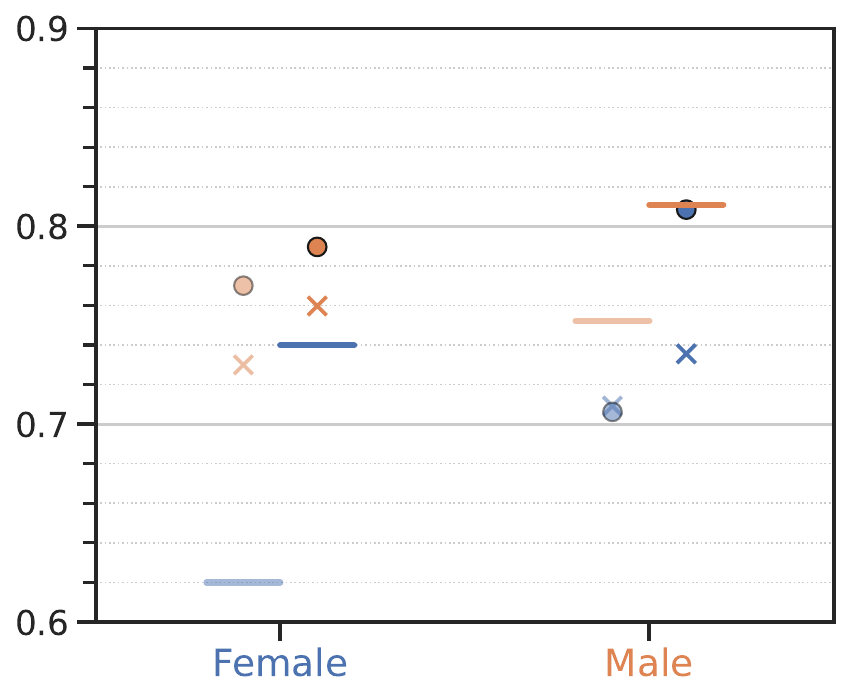}
\includegraphics[width=0.35\linewidth]{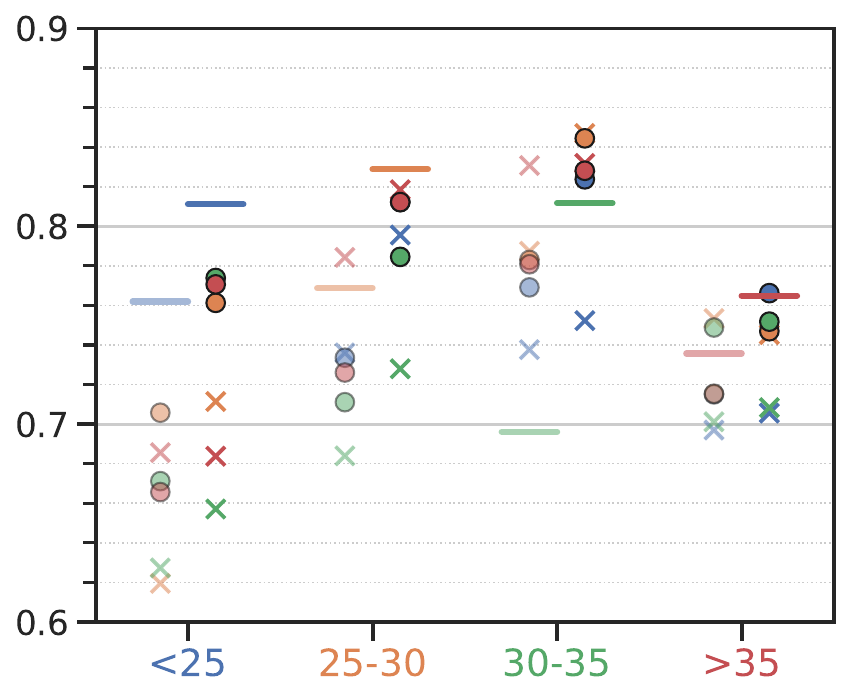}\\
\includegraphics[width=0.35\linewidth]{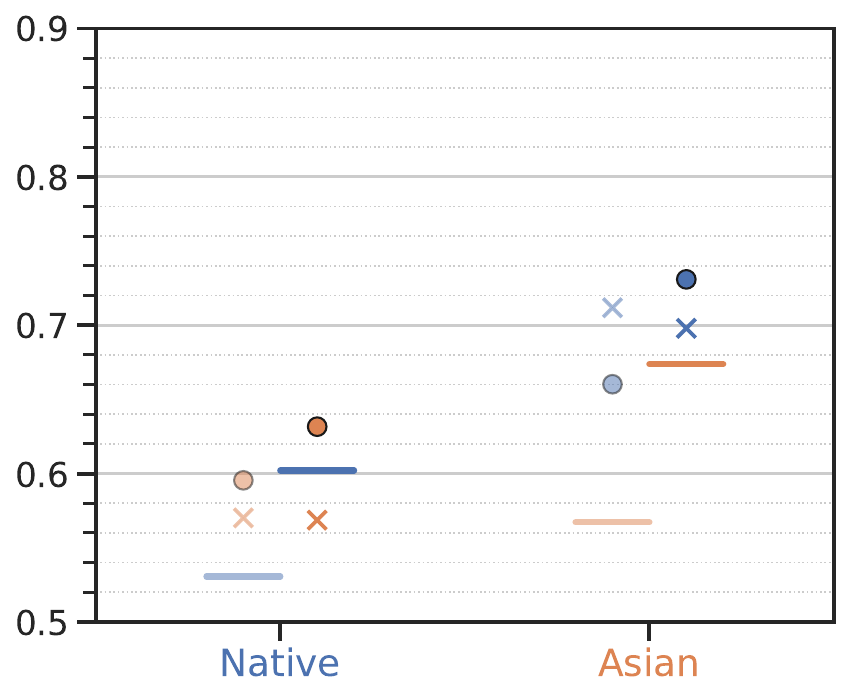}
\includegraphics[width=0.35\linewidth]{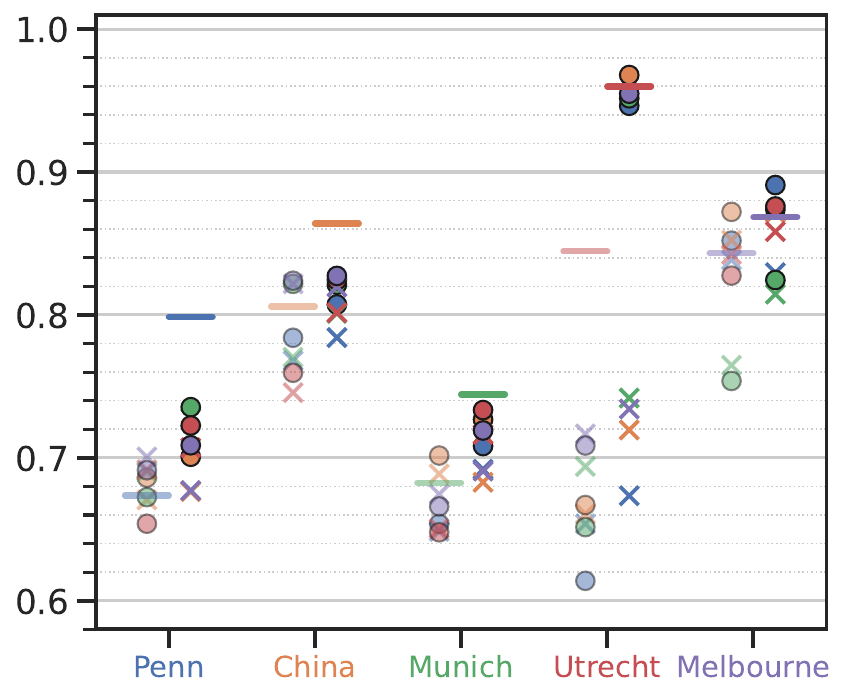}\\
\includegraphics[width=0.9\linewidth]{nn_legend}
\caption{}
\label{fig:app:scz_nn}
\end{subfigure}
\caption{
\textbf{Discrepancy in the data and AUC of diagnostic models of schizophrenia on different groups in the population.}
\textbf{(a-b)} Distance between leaves of this dendrogram in \textbf{(b)} indicates the pairwise MMD statistic in \textbf{(a)} between learned features of pairs of groups. \textbf{(c)} Average AUC of schizophrenia classification computed using five-fold nested cross-validation. We trained a deep neural network (translucent markers) and an ensemble using boosting, bagging and stacking (bold markers), using data from different source groups (different colors) and evaluated this model (cross marks) on data from different target groups (X-axis); circles denote model fitted using our $\a$-weighted empirical risk minimization (ERM) procedure with access to 10\% data from the target group; horizontal lines denote models that are directly trained on the target group using 80\% of data (the rest for testing). Similar to diagnostic models of Alzheimer’s disease, in general, (i) the AUC of ensemble models is higher than that of the neural network in all cases ($p$ < 0.01), (ii) AUC of a model trained on a source group remains remarkably high when evaluated on the target group (crosses), (iii) in almost all cases for the ensemble, it further improves when one has access to a small fraction of data from the target group (circles are higher than crosses), and (iv) some times even beyond the AUC of a model trained only on the target group (circles above horizontal lines).
}
\label{fig:app:dendrogram_nn_ensemble_scz}
\end{figure}

\begin{figure}
\centering
\begin{subfigure}[b]{0.375\linewidth}
\centering
\includegraphics[width=0.2\linewidth]{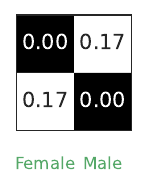}
\includegraphics[width=0.3\linewidth]{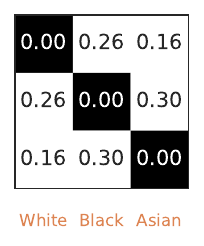}\\
\includegraphics[width=0.75\linewidth]{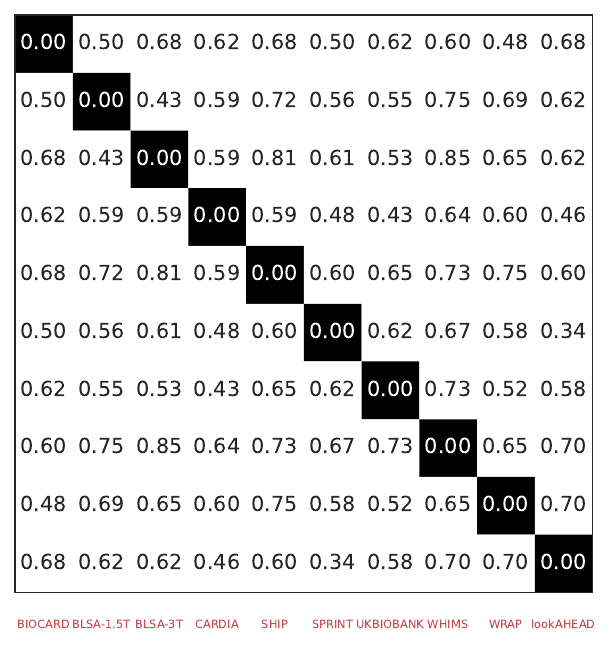}
\caption{}
\label{fig:app:distance_age}
\end{subfigure}
\begin{subfigure}[b]{0.35\linewidth}
\centering
\includegraphics[width=\linewidth]{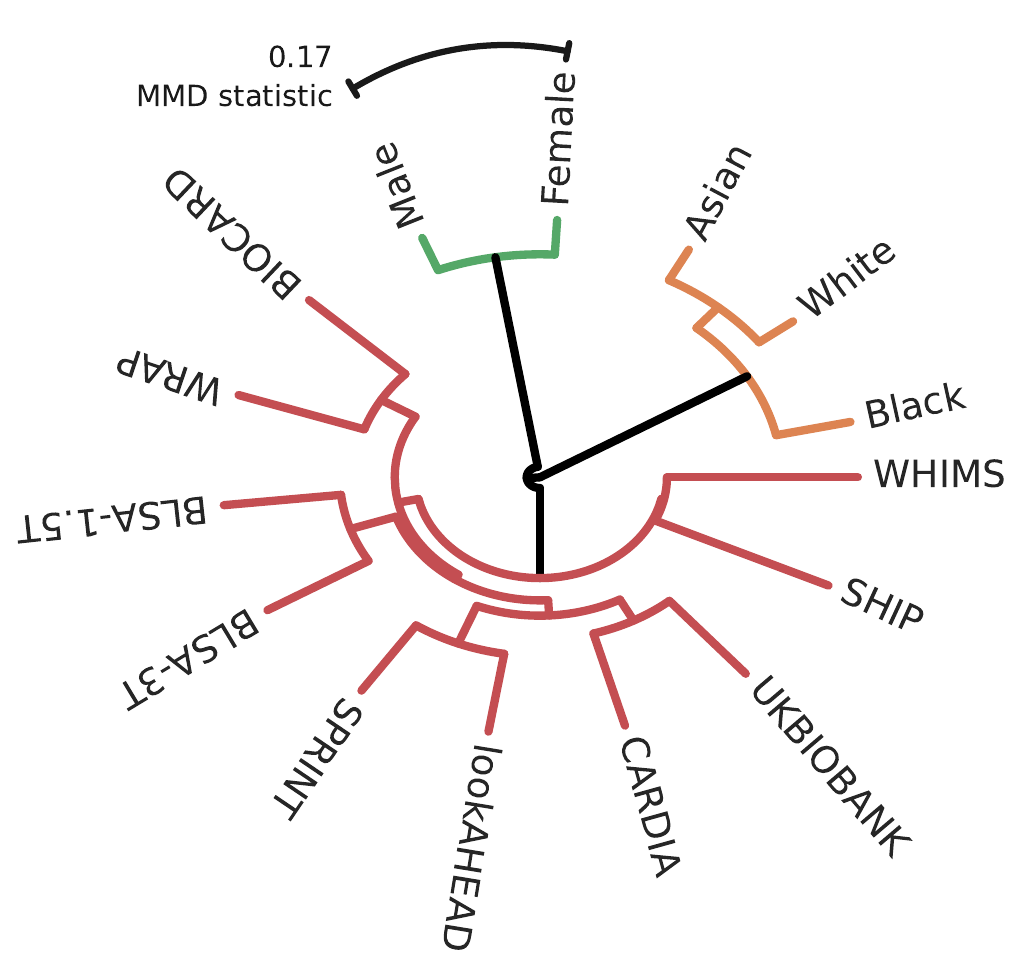}
\caption{}
\label{fig:app:dendrogram_age}
\end{subfigure}

\begin{subfigure}[b]{\linewidth}
\centering
\includegraphics[width=0.4\linewidth]{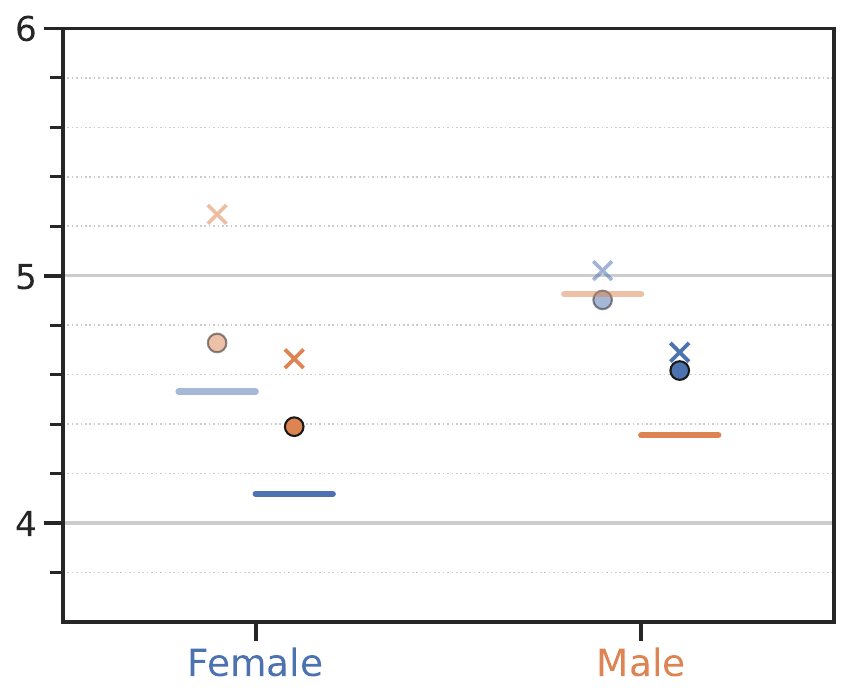}
\includegraphics[width=0.4\linewidth]{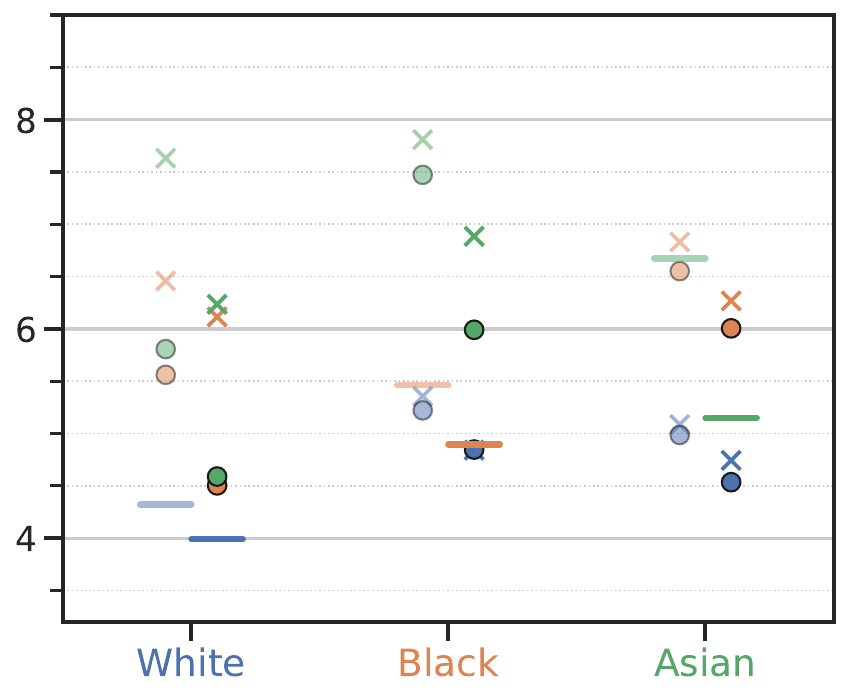}\\
\includegraphics[width=0.8\linewidth]{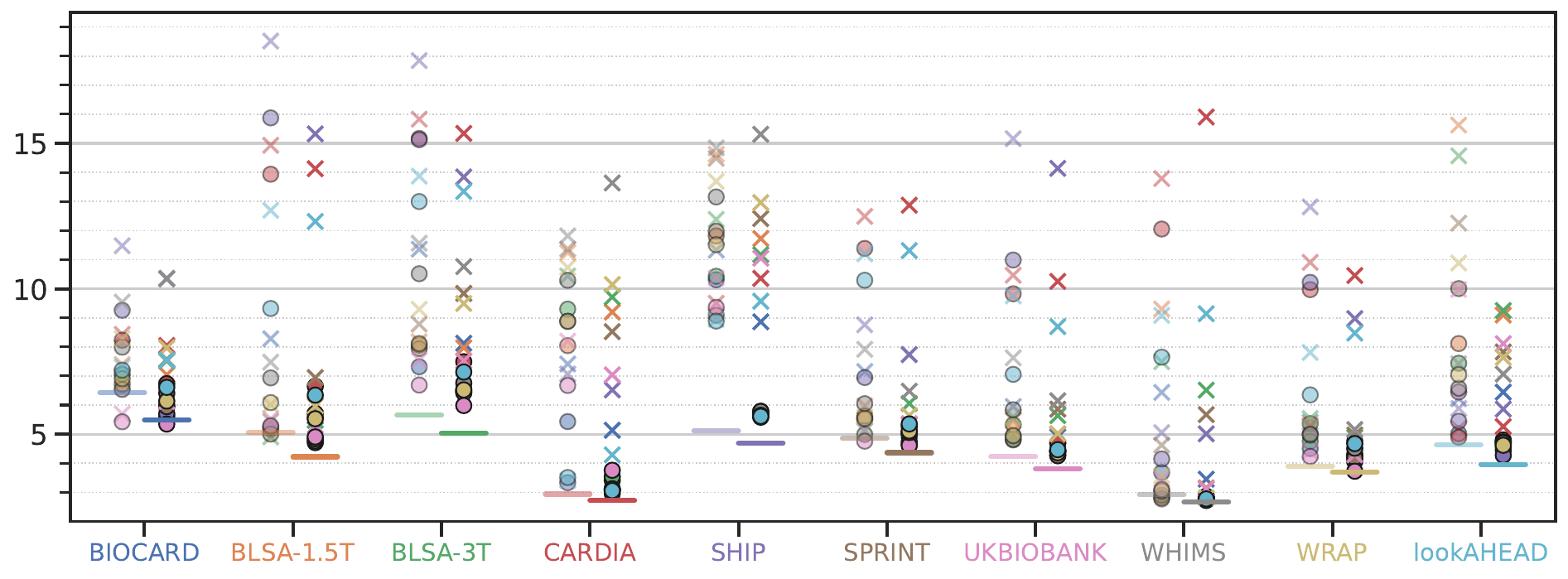}\\
\includegraphics[width=0.8\linewidth]{nn_legend}
\caption{}
\label{fig:app:age_nn}
\end{subfigure}
\caption{
\textbf{Discrepancy in the data and MAE (in years) models that predict the brain age.}
\textbf{(a-b)} Distance between leaves of this dendrogram in \textbf{(b)} indicates the pairwise MMD statistic in \textbf{(a)} between learned features of pairs of groups. \textbf{(c)} Average AUC of schizophrenia classification computed using five-fold nested cross-validation. We trained a deep neural network (translucent markers) and an ensemble using boosting, bagging and stacking (bold markers), using data from different source groups (different colors) and evaluated this model (cross marks) on data from different target groups (X-axis); circles denote model fitted using our $\a$-weighted empirical risk minimization (ERM) procedure with access to 10\% data from the target group; horizontal lines denote models that are directly trained on the target group using 80\% of data (the rest for testing). Similar to diagnostic models of Alzheimer’s disease and schizophrenia, in general, (i) the MAE of ensemble models is lower than that of the neural network in all cases ($p$ < 0.01), and (ii) in almost all cases for the ensemble, the MAE improves when one has access to a small fraction of data from the target group (circles are lower than crosses).
}
\label{fig:app:dendrogram_nn_ensemble_age}
\end{figure}

% \subsection{Generalization to groups outside the training set can be improved with access to a small amount of labeled samples}
% \label{s:app:adaptation_to_groups}

\begin{figure}
\centering
\includegraphics[width=0.4\linewidth]{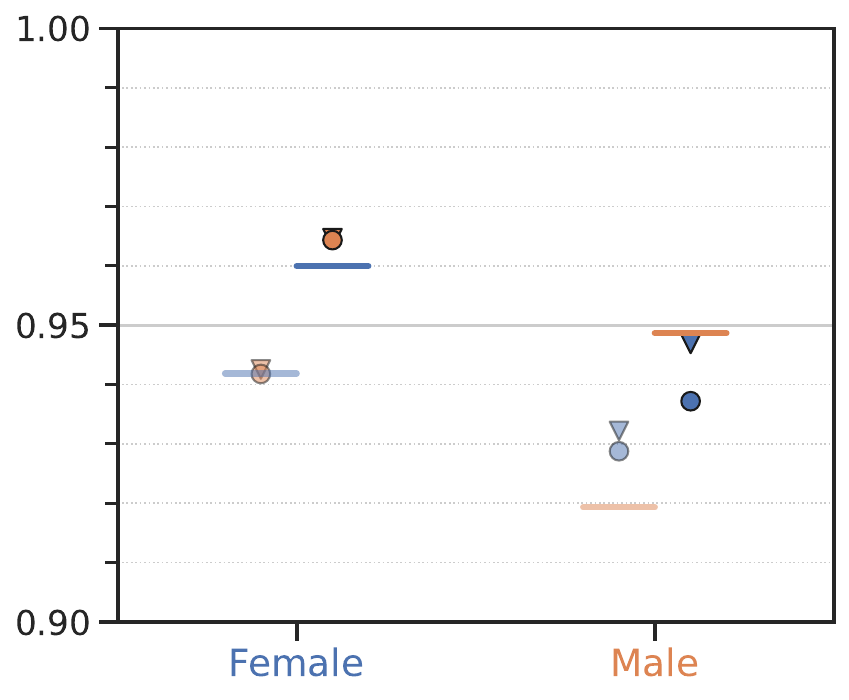}
\includegraphics[width=0.4\linewidth]{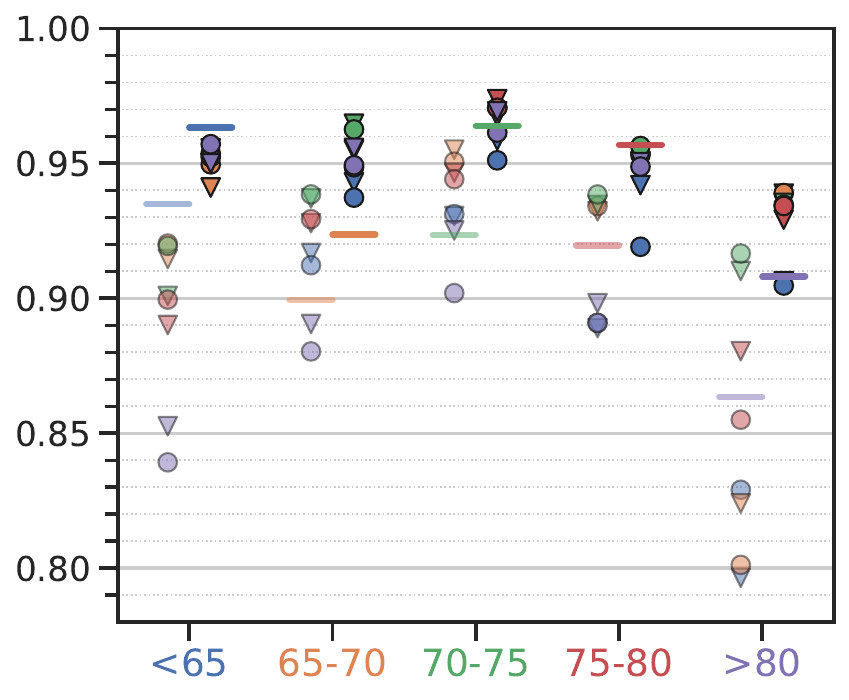}\\
\includegraphics[width=0.4\linewidth]{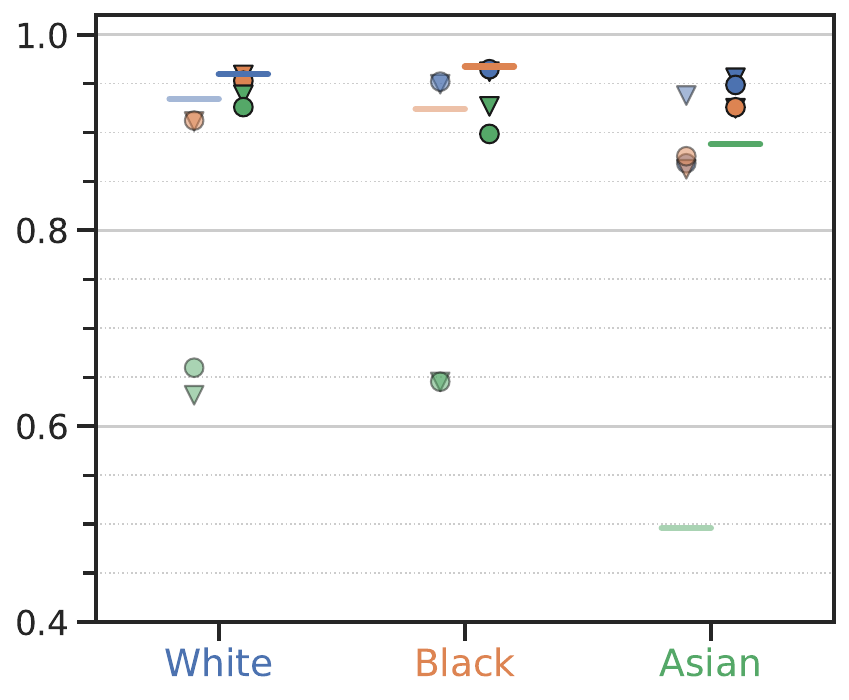}
\includegraphics[width=0.4\linewidth]{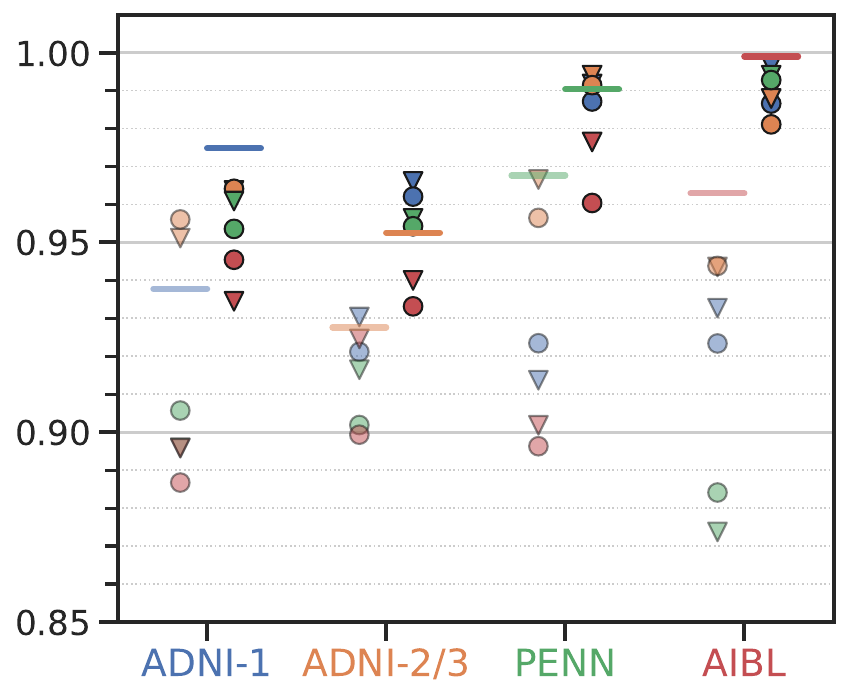}\\
\includegraphics[width=0.95\linewidth]{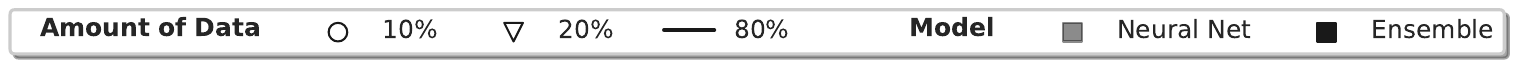}
\caption{
\textbf{Average AUC of Alzheimer’s disease classification computed using five-fold nested cross-validation}. We trained a machine learning model, either a deep neural network (translucent markers) or an ensemble using boosting, bagging and stacking (bold markers), using data from different source groups (different colors); circles and triangles denote model fitted using our $\a$-weighted empirical risk minimization (ERM) procedure with access to 10\% and 20\% data respectively from the target group; horizontal lines denote models that are directly trained on the target group using 80\% of data (the rest for testing). All models use data from multiple sources, namely structural measures, demographic, clinical variables, genetic factors, and cognitive scores. In some cases, the AUC using 20\% data is better (triangles above circles) but overall the AUC of models trained with 20\% target data is statistically the same as that of using 10\% data. This indicates that we can fruitfully predict the Alzheimer’s disease diagnosis for subjects from different groups in the population with access to as little as 10\% data from those groups.
}
\label{fig:app:nn_ensemble_20_ad}
\end{figure}

\begin{figure}
\centering
\includegraphics[width=0.4\linewidth]{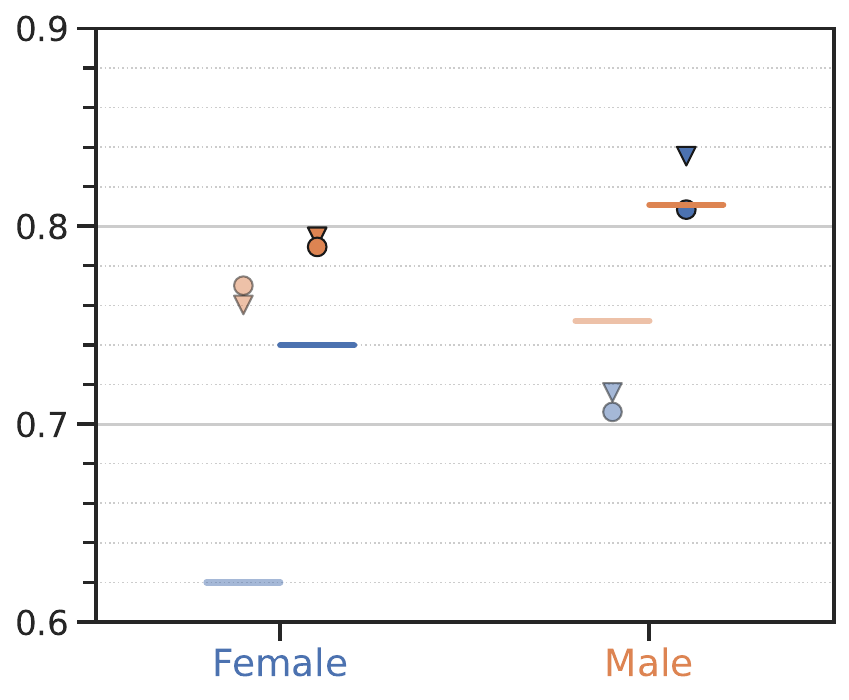}
\includegraphics[width=0.4\linewidth]{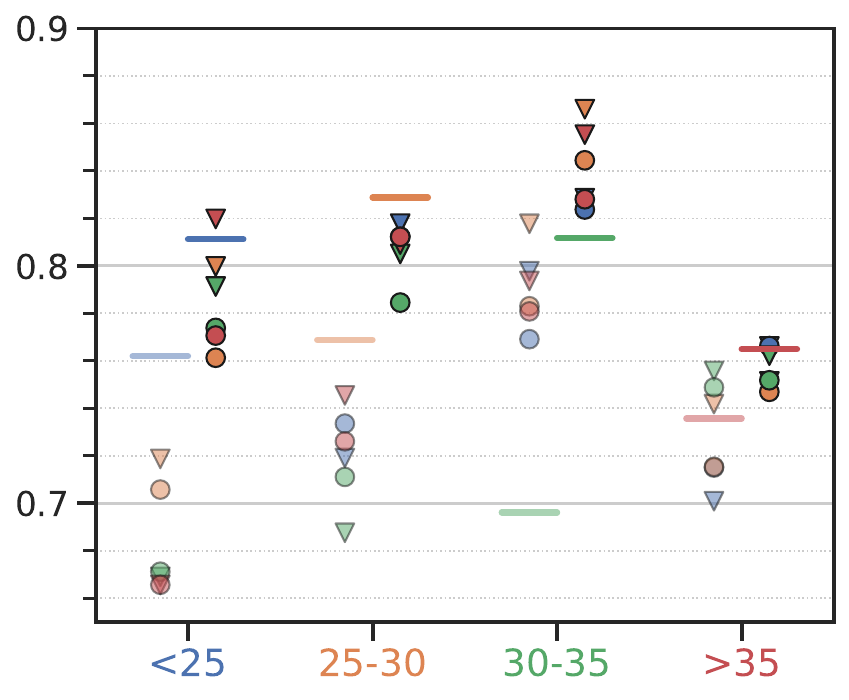}\\
\includegraphics[width=0.4\linewidth]{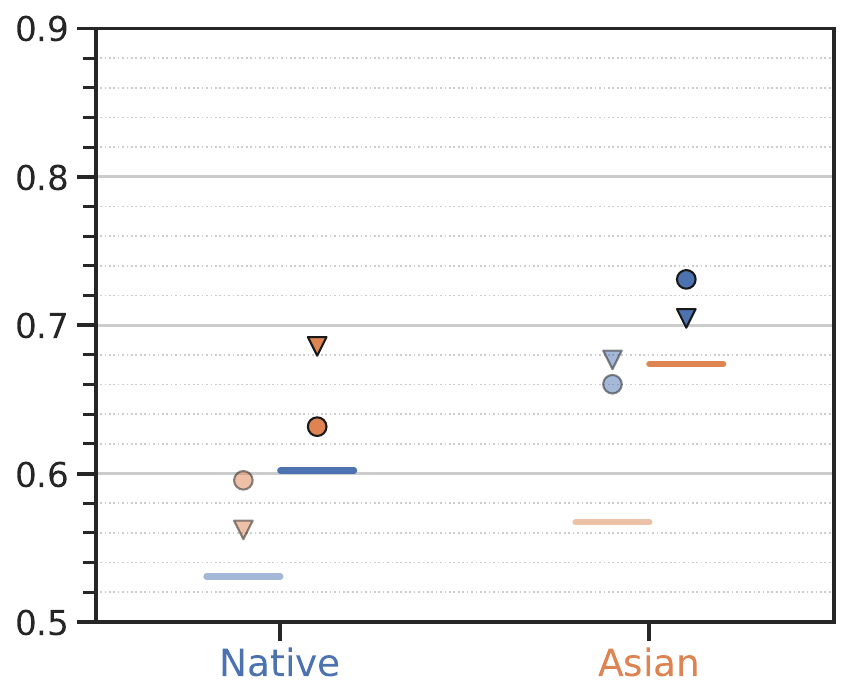}
\includegraphics[width=0.4\linewidth]{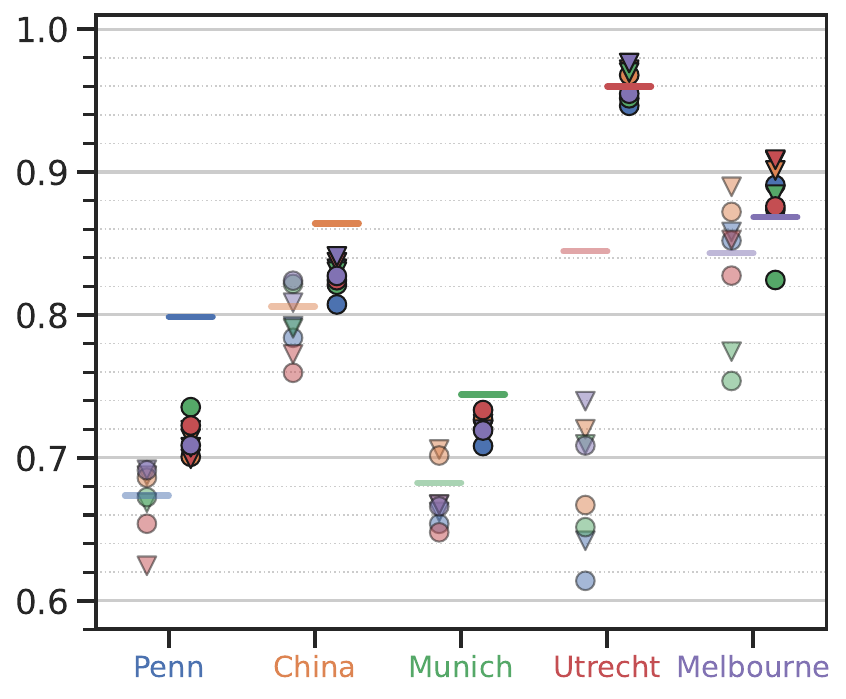}\\
\includegraphics[width=0.95\linewidth]{nn_20_legend}
\caption{
\textbf{Average AUC of schizophrenia classification computed using five-fold nested cross-validation}. We trained a machine learning model, either a deep neural network (translucent markers) or an ensemble using boosting, bagging and stacking (bold markers), using data from different source groups (different colors); circles and triangles denote model fitted using our $\a$-weighted empirical risk minimization (ERM) procedure with access to 10\% and 20\% data respectively from the target group; horizontal lines denote models that are directly trained on the target group using 80\% of data (the rest for testing). All models use data from multiple sources, namely structural measures, demographic, clinical variables, genetic factors, and cognitive scores. In some cases, the AUC using 20\% data is better (triangles above circles) but overall the AUC of models trained with 20\% target data is statistically the same as those trained with 10\% data. This indicates that we can fruitfully predict the schizophrenia diagnosis for subjects from different groups in the population with access to as little as 10\% data from those groups.
}
\label{fig:app:nn_ensemble_20_scz}
\end{figure}

\begin{figure}
\centering
\includegraphics[width=0.4\linewidth]{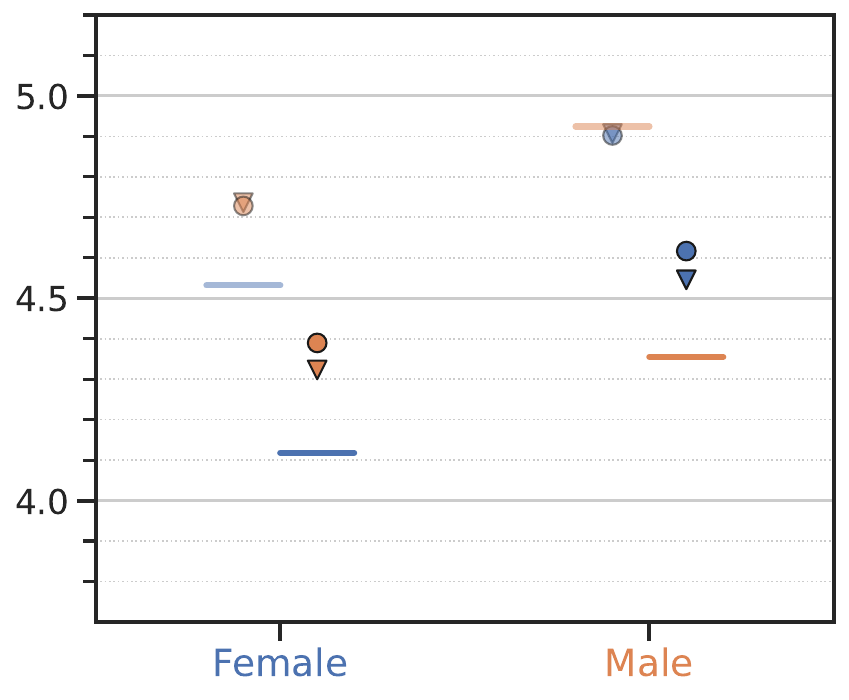}
\includegraphics[width=0.4\linewidth]{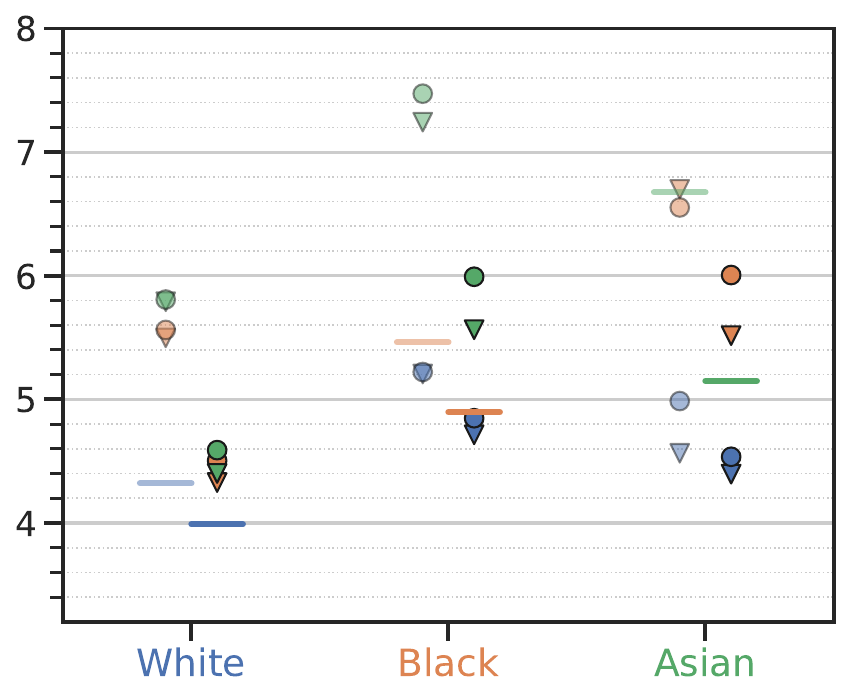}\\
\includegraphics[width=0.8\linewidth]{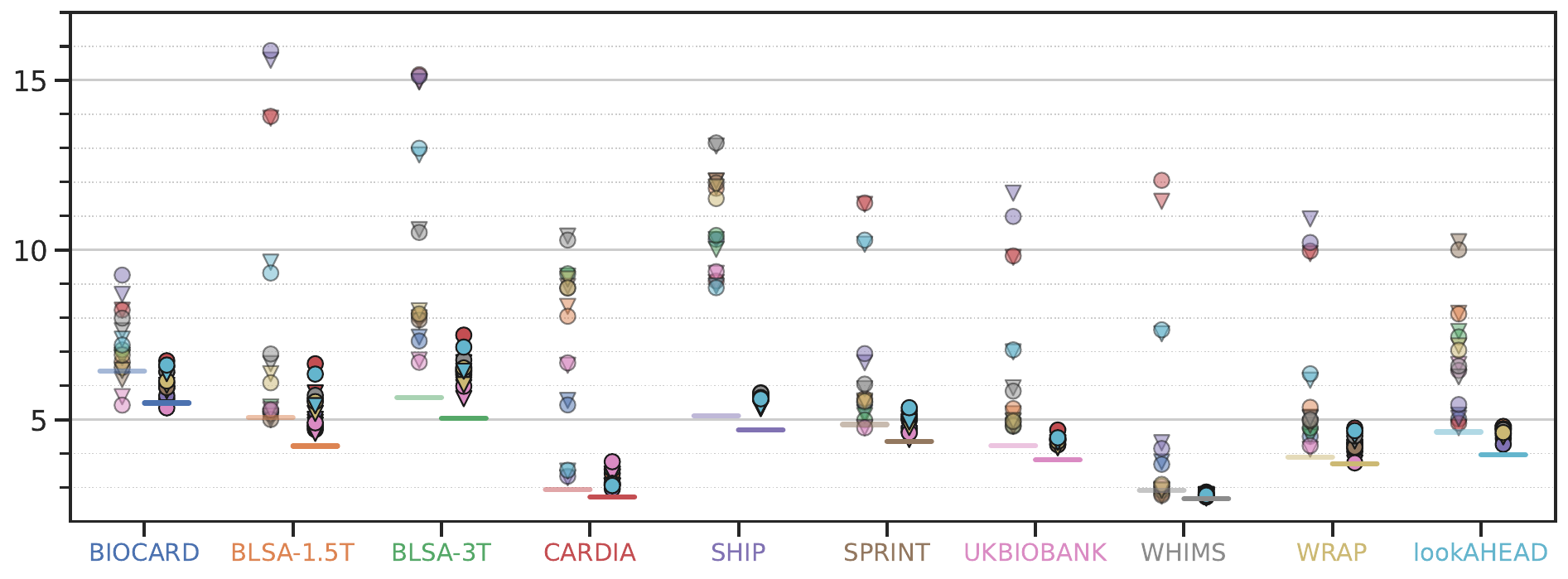}\\
\includegraphics[width=0.95\linewidth]{nn_20_legend}
\caption{
\textbf{Average mean average error (MAE) in years of predicting the brain age, computed using five-fold nested cross-validation}. The setup is the same as that of~\cref{fig:app:nn_ensemble_20_ad,fig:app:nn_ensemble_20_scz}. For brain age prediction, again, we see that the MAE obtained using 20\% data is some times better than of 10\% data (triangles below circles) but overall the two settings achieve the same MAE statistically. This indicates that we can fruitfully predict the brain age of subjects from different groups in the population with access to as little as 10\% data from those groups.
}
\label{fig:app:nn_ensemble_20_age}
\end{figure}

\begin{figure}
\begin{subfigure}[b]{\linewidth}
\centering
\includegraphics[width=0.32\linewidth]{s_mci_sex}
\includegraphics[width=0.32\linewidth]{s_mci_age}
\includegraphics[width=0.32\linewidth]{s_mci_race}
\caption{}
\label{fig:app:s_mci}
\end{subfigure}

\begin{subfigure}[b]{\linewidth}
\centering
\includegraphics[width=0.32\linewidth]{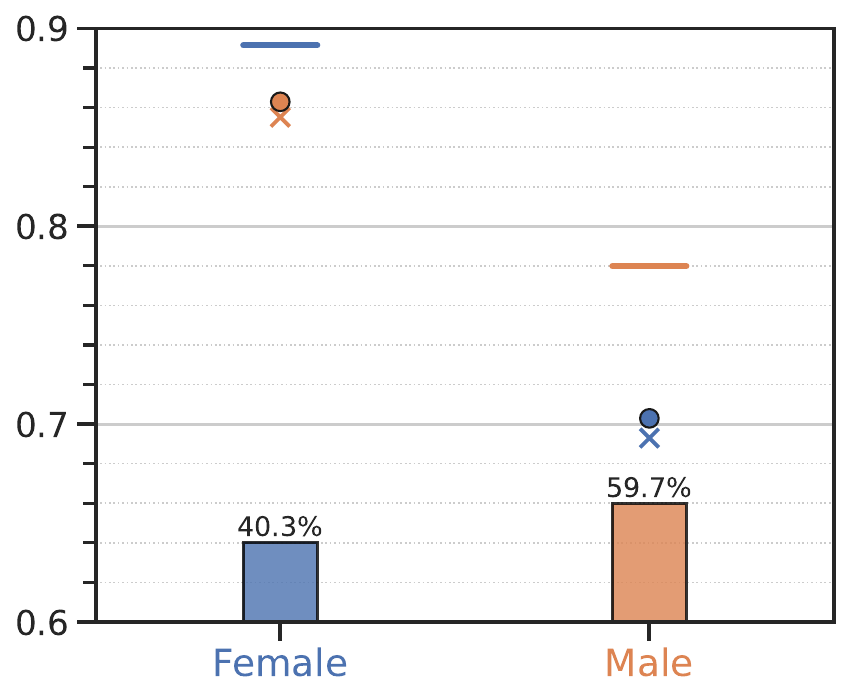}
\includegraphics[width=0.32\linewidth]{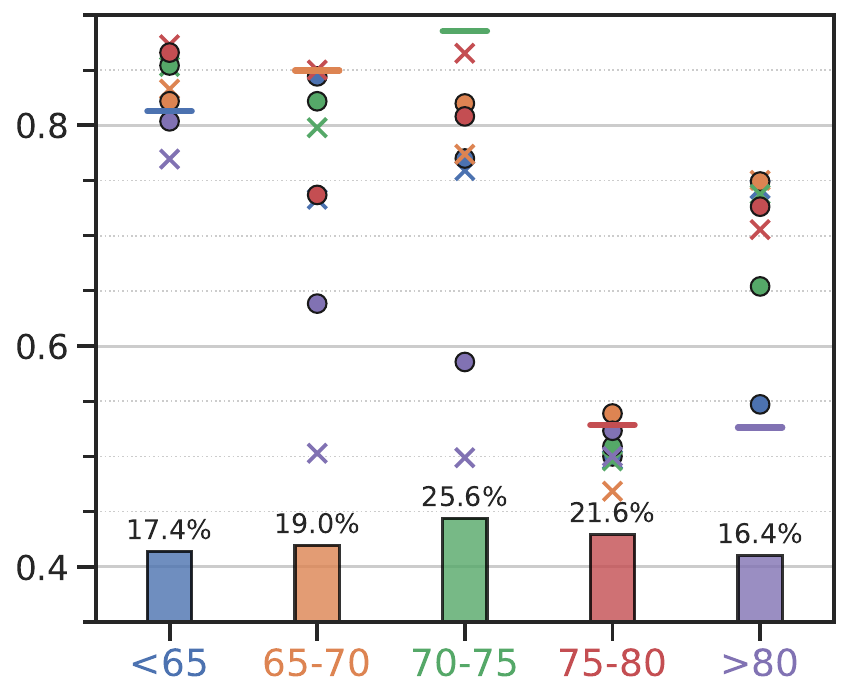}
\includegraphics[width=0.32\linewidth]{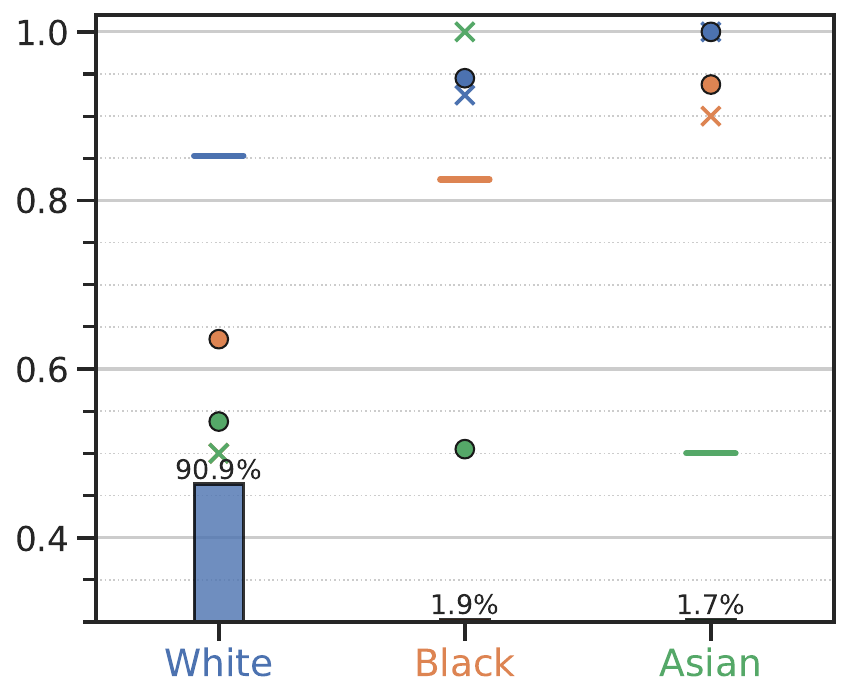}\\
\includegraphics[width=0.55\linewidth]{legend}
\caption{}
\label{fig:app:n_mci}
\end{subfigure}
\caption{
Linear discriminant analysis on the output probabilities (that determines AD vs. control) of the ensemble models trained for Alzheimer’s disease diagnosis is used to study whether subjects with mild cognitive impairment (MCI) progress to AD (known as pMCI) or remain stable MCI (known as sMCI) in \textbf{(a)}, and pMCI vs. nMCI in \textbf{(b)} where the latter refers to MCI subjects diagnosed as normal within three years of the first clinical visit. The AUC of pMCI vs. sMCI on the target group is shown for three different attributes (sex, age group and race) when models are trained only on data from the source group (crosses), using $\a$-weighted ERM using all data from the source and 10\% data from the target group (circles) and with access to only all data from the target group (horizontal lines). Improvements in the AD vs. control AUC of these models with 10\% data translate to improvements in the ability to distinguish between pMCI and sMCI subjects, using only baseline scans (circles above cross) except when target groups are Black or Asian (due to very little data in these groups).
}
\label{fig:app:mci}
\end{figure}

\begin{figure}[p]
\centering
\includegraphics[width=0.49\linewidth]{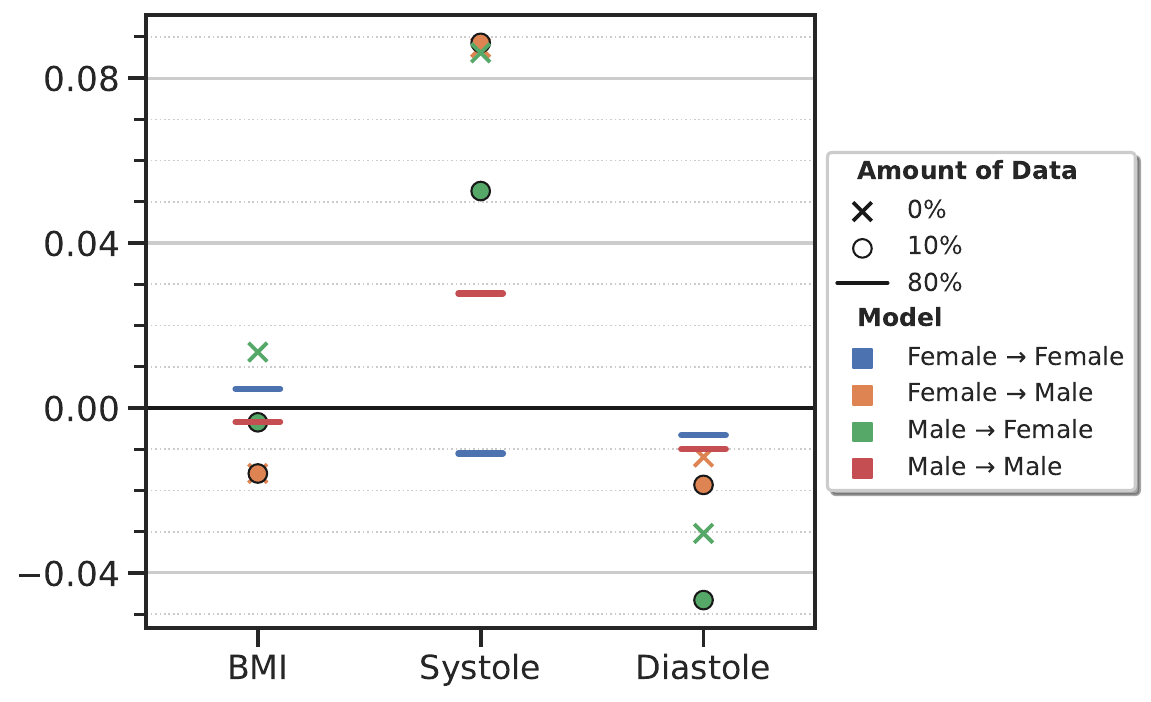}
\includegraphics[width=0.49\linewidth]{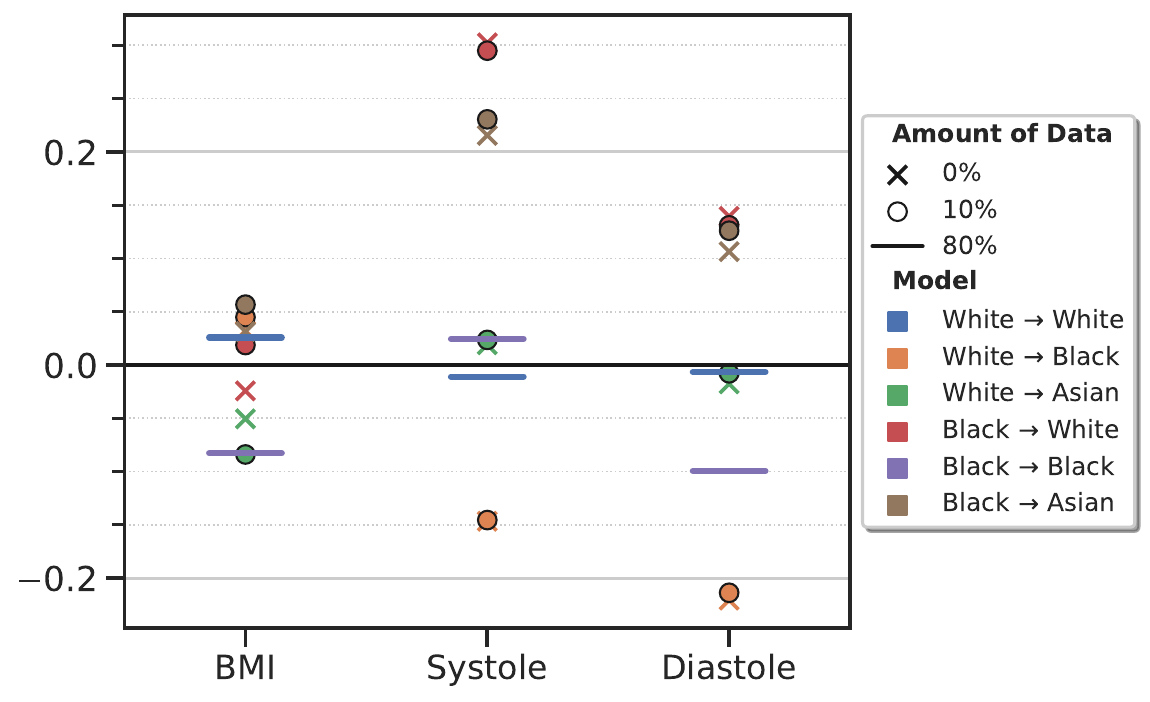}
\caption{
\textbf{Correlation of brain age regression with cardiovascular and lifestyle factors.}
Pearson's correlation between the brain age residual (predicted brain age minus chronological age) and clinical factors for two different attributes (sex and race) for models trained only on source data (crosses), using $\a$-weighted ERM on all source data and 10\% target data (circles) and only on all target data (horizontal lines). Unlike other plots, colors denote different pairs of source and target groups. Tests (X-axis) marked in red are expected to be negatively correlated with brain aging whereas those marked in black are expected to be positively correlated with brain aging according to the existing literature. Body mass index (BMI) is an individual's health indicator based on tissue mass (muscle, fat, and bone) and height. Systolic and diastolic blood pressures are the measures of heart functioning. In almost all cases, we observe weak correlations which are statistically insignificant ($p$-value > 0.01).
}
\label{fig:app:cardio}
\end{figure}

% \subsection{Demographic and acquisition biases are mitigated by adapting models to new domains}

\begin{figure}
\begin{subfigure}[b]{\linewidth}
\centering
\includegraphics[width=0.35\linewidth]{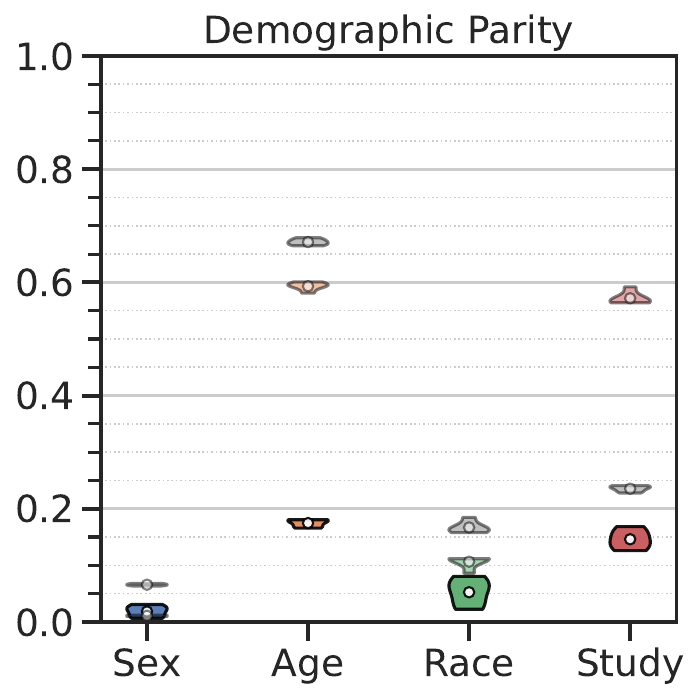}
\hspace*{4ex}
\includegraphics[width=0.35\linewidth]{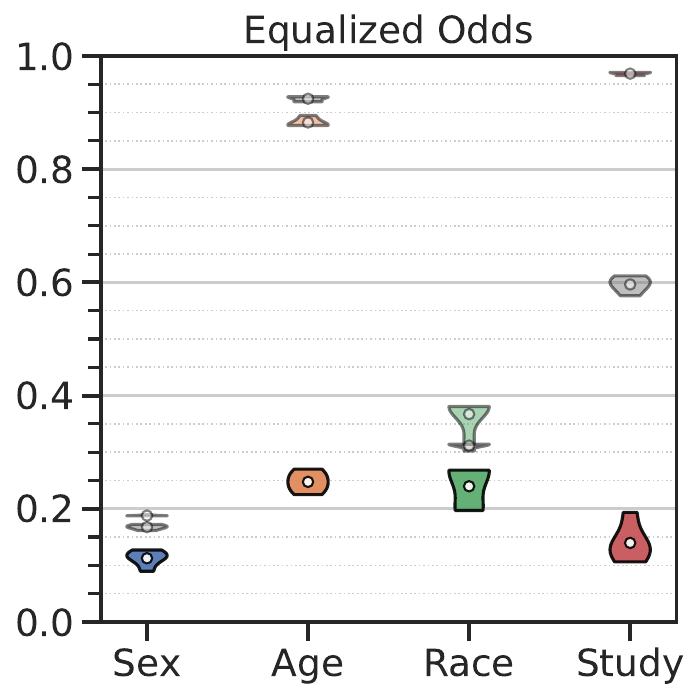}
\caption{Alzheimer's disease}
\label{fig:app:ad_fair}
\end{subfigure}

\begin{subfigure}[b]{\linewidth}
\centering
\includegraphics[width=0.35\linewidth]{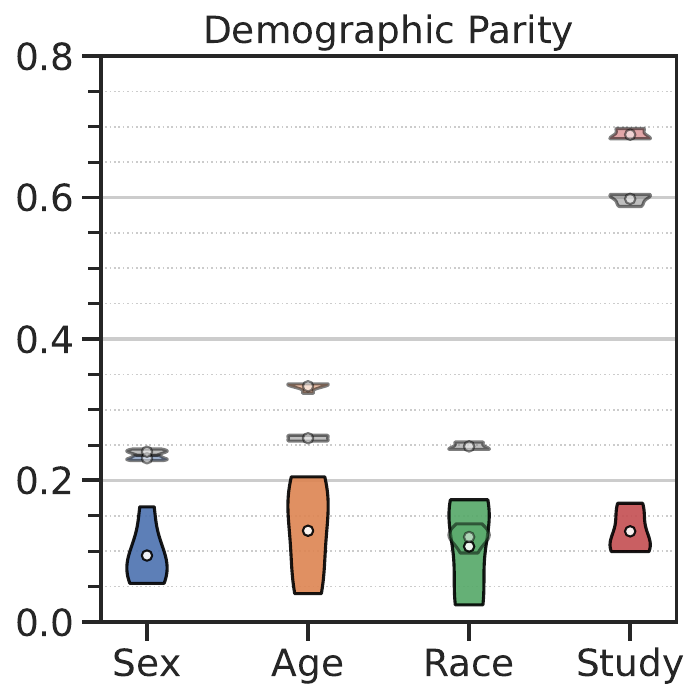}
\hspace*{4ex}
\includegraphics[width=0.35\linewidth]{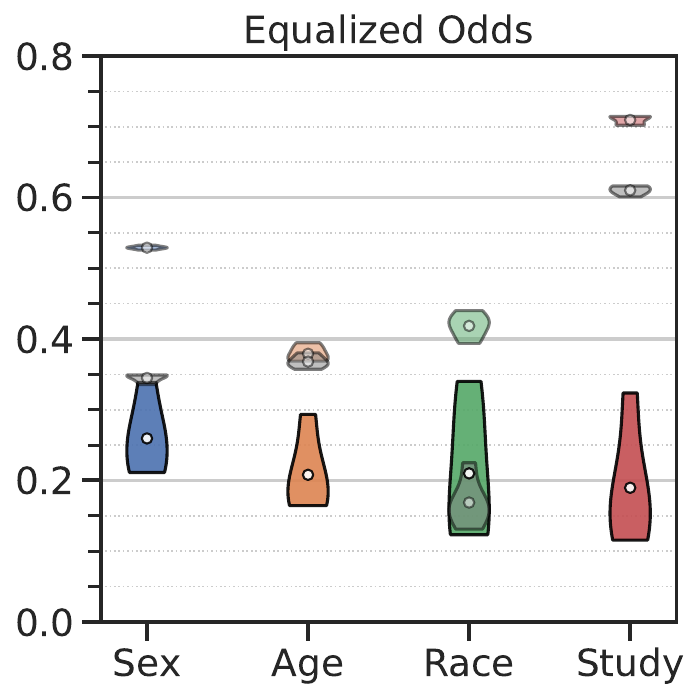}
\caption{Schizophrenia}
\label{fig:app:scz_fair}
\end{subfigure}

\begin{subfigure}[b]{\linewidth}
\centering
\includegraphics[width=0.35\linewidth]{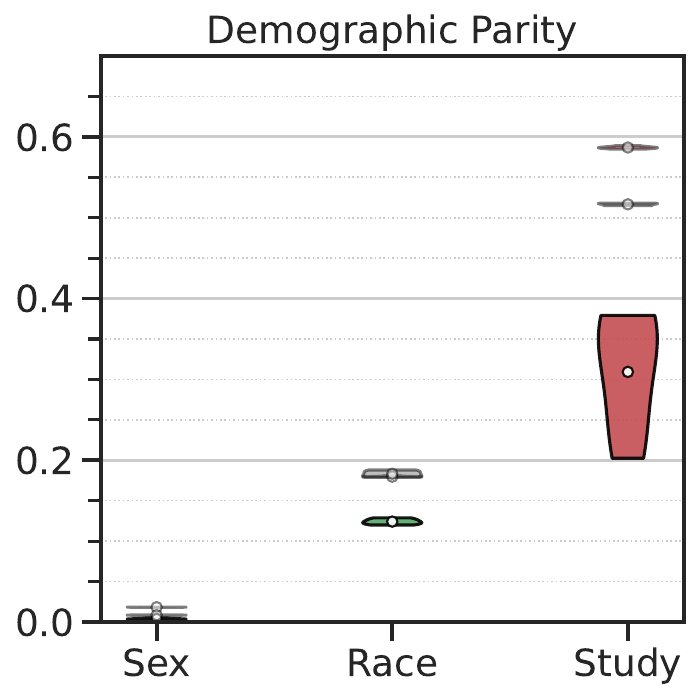}
\caption{Brain age prediction}
\label{fig:app:age_fair}
\end{subfigure}
\caption{
\textbf{Fairness assessment of machine learning models with respect to sensitive attributes including sex, age group, race, and clinical study.} Violin plots denote the test demographic parity differences (DPD) and equalized odds difference (EOD) on five different held-out subsets of data. Translucent gray denotes the performance of a baseline deep network while translucent colors indicate the ensemble models. Solid colors indicate using $\alpha$-weighted ERM on all source data and 10\% target data. White dots denote the average performance of each group. A model is perfectly fair if the fairness disparity (DPD or EOD) is zero. In general, we observe that our $\alpha$-weighted ERM models have lower disparities for all sensitive attributes in both metrics compared to the neural net and ensemble models.
}
\label{fig:app:fairness}
\end{figure}

\begin{table}[p]
\begin{spacing}{1}
\caption{
%\newtext{
\textbf{Summary of variables in the data from the iSTAGING consortium (Alzheimer’s disease and brain age) and the PHENOM consortium (schizophrenia) used in this study.} The numbers next to the variable names are the dimensionality.}
%}
\label{tab:variables}
\begin{center}
%\begin{footnotesize}
\begin{small}
\begin{tabular}{p{0.15\linewidth}rrrr}
\toprule
\textbf{Variables} &&&  iSTAGING & PHENOM  \\
\cmidrule{4-5}
MR imaging\\
& Region-of-interest volumes & 145 & $\checkmark$ & $\checkmark$ \\
& White matter lesion volume & 1 & $\checkmark$ &  \\
Demographics\\
& Gender & 1 & $\checkmark$ & $\checkmark$ \\
& Age & 1 & $\checkmark$ & $\checkmark$ \\
& Race & 1 & $\checkmark$ & $\checkmark$ \\
& Education level & 1 & & $\checkmark$ \\
& Marital status & 1 & & $\checkmark$\\
& Employment status & 1 & & $\checkmark$\\
& Handedness & 1 & & $\checkmark$  \\
& Smoking status & 1 & $\checkmark$ \\
Clinical \\
& Diabetes & 1 & $\checkmark$ \\
& Hypertension & 1 & $\checkmark$ \\
& Hyperlipidemia & 1 & $\checkmark$ \\
& Blood pressure (systolic/diastolic) & 2 & $\checkmark$ \\
& Body mass index & 1 & $\checkmark$ \\
Genetic factor \\
& Apolipoprotein E allele 2 & 1 & $\checkmark$ \\
& Apolipoprotein E allele 3 & 1 & $\checkmark$ \\
& Apolipoprotein E allele 4 & 1 & $\checkmark$ \\
Cognitive score \\
& Mini-mental state exam & 1 & $\checkmark$ \\
\bottomrule
\end{tabular}
%\end{footnotesize}
\end{small}
\end{center}
\end{spacing}
\end{table}

\begin{figure}
\centering
\includegraphics[width=0.49\linewidth]{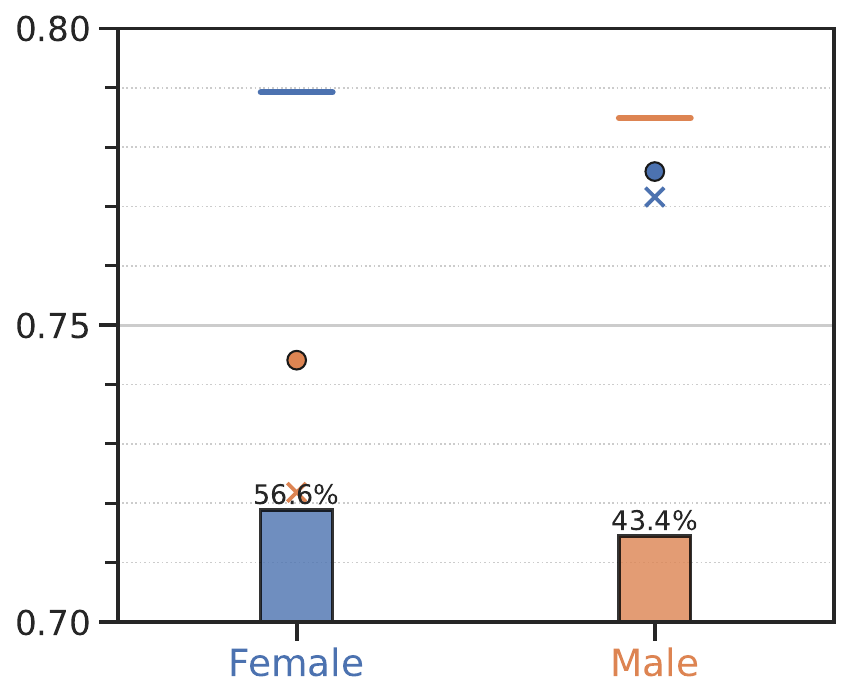}
\includegraphics[width=0.49\linewidth]{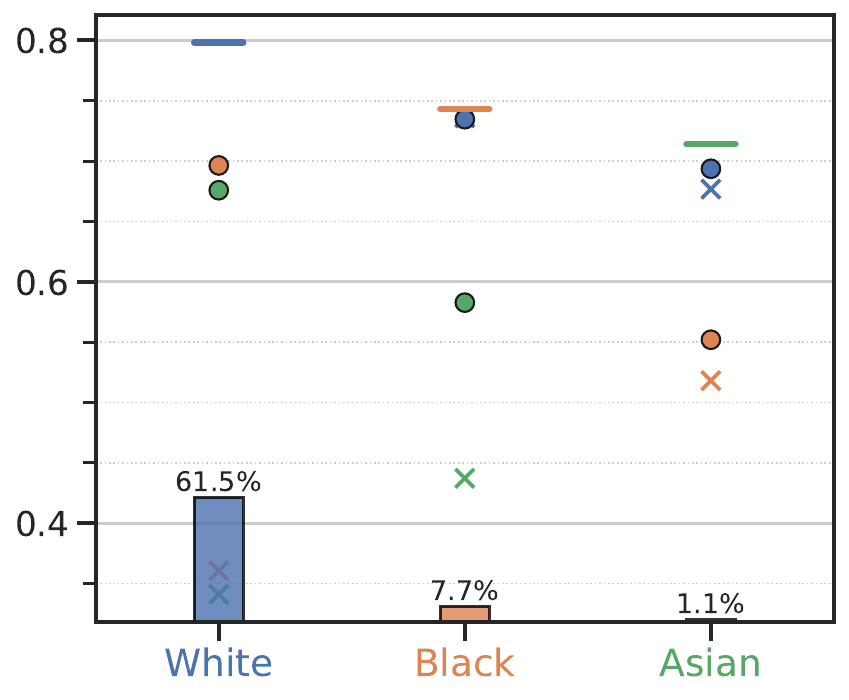}\\
\includegraphics[width=1.0\linewidth]{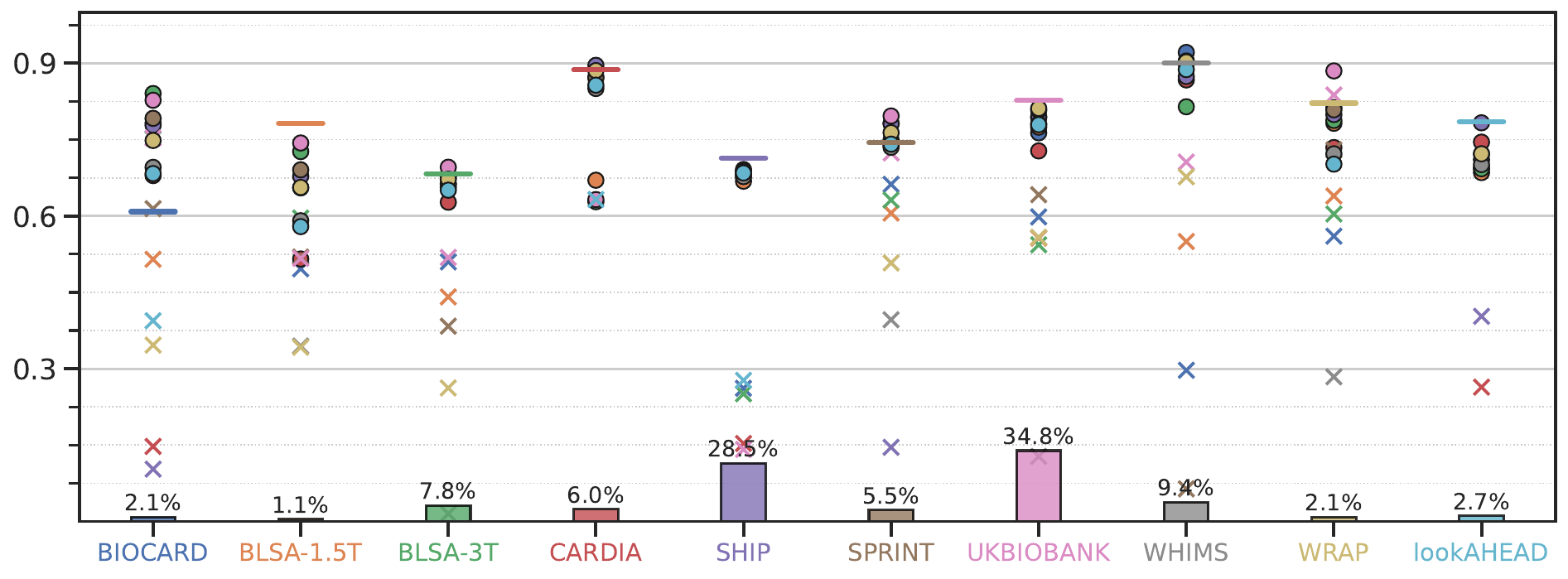}\\
\includegraphics[width=0.55\linewidth]{legend}
\caption{
%\newtext{
\textbf{Brain age prediction with coefficient of determination metric.}
Markers denote the coefficient of determination ($R^2$) of an ensemble that predicts the brain age on different target groups in the population computed using five-fold nested cross-validation, for models trained only on data from the target group (e.g., Female subjects, denoted by the blue horizontal line), only on data from the source group (crosses), and trained on all data from the source group and 10\% data from the target group (circles). 
we find that using $R^2$ agrees with our original conclusions drawn from MAE for brain age prediction.
Note that $R^2$ can be negative on the test data; this is important when a model trained on one group, say age group, is being evaluated on another where subjects are from a different age group. We have therefore truncated our plot at zero. A baseline model (crosses) cannot generalize well to a new out-of-distribution group in general, whereas our weighted-ERM (circles in the figures) improves the performance significantly.
It should also be noted that many baseline models (crosses) have negative $R^2$ scores in the cross-study scenarios (lower panel).
Our proposed approach that uses a small amount of target data for adaptation (circles in the figures) has consistently high $R^2$ scores.}
%}
\label{fig:age_r2}
\end{figure}

\begin{table}
\caption{
%\newtext{
\textbf{Summary of the data from the iSTAGING consortium (Alzheimer's disease and brain age) and the PHENOM consortium (schizophrenia) used in this study.}}
%}
\label{tab:data_absolute}
\begin{center}
\begin{footnotesize}

\resizebox{0.7\linewidth}{!}{
\begin{tabular}{p{0.15\linewidth}rrrrrrr}
\toprule
\textbf{Alzheimer’s} && ADNI-1 & ADNI-2/3 & PENN & AIBL & OASIS & Total \\
\textbf{Disease} && (17.37\%) & (23.29\%) & (25.44\%) & (10.07\%) & (23.82\%) \\
\cmidrule{3-8}
% & Subjects & 364 & 488 & 533 & 211 & 499 & 2095  \\
Subjects\\
& Control & 173 & 261 & 228 & 119 & 276 & 1057  \\
& Patient & 191 & 227 & 305 & 92 & 223 & 1038  \\
Sex \\
& Female & 174 & 250 & 340 & 129 & 266 & 1160 \\
& Male & 190 & 238 & 192 & 82 & 233 & 935 \\
Age (years) \\
& 0--65 & 24 & 56 & 101 & 29 & 103 & 312 \\
& 65--70 & 31 & 130 & 112 & 41 & 107 & 421 \\
& 70--75 & 116 & 111 & 109 & 60 & 95 & 491 \\
& 75--80 & 104 & 110 & 105 & 41 & 101 & 461 \\
& > 80 & 89 & 81 & 106 & 40 & 93 & 409 \\
Race \\
& White & 337 & 271 & 401 & 117 & 409 & 1537 \\
& Black & 19 & 14 & 107 & - & 85 & 225 \\
& Asian & 5 & 9 & 10 & - & 4 & 28 \\
\bottomrule\\
\end{tabular}
}

\vspace*{2em}
\resizebox{0.75\linewidth}{!}{
\begin{tabular}{p{0.15\linewidth} rrrrrrr}
\toprule
\textbf{Schizophrenia} &  & Penn & China & Munich & Utrecht & Melbourne & Total \\
& & (22.28\%) & (13.94\%) & (29.64\%) & (20.12\%) & (14.03\%)\\
\cmidrule{3-8}
Subjects\\
%& Subjects & 227 & 142 & 302 & 205 & 143 & 1019 \\
& Control & 131 & 76 & 157 & 115 & 84 & 563 \\
& Patient & 96 & 66 & 145 & 90 & 59 & 456 \\
Sex\\
& Female & 121 & 69 & 79 & 71 & 41 & 381 \\
& Male & 106 & 73 & 223 & 134 & 102 & 638 \\
Age ( years)\\
& 0--25 & 59 & 50 & 96 & 103 & 60 & 368 \\
& 25--30 & 64 & 24 & 74 & 42 & 22 & 226 \\
& 30--35 & 36 & 22 & 61 & 29 & 14 & 162 \\
& > 35 & 68 & 46 & 71 & 31 & 47 & 263 \\
Race\\
& Native & 107 & - & - & - & - & 107 \\
& Asian & 75 & - & - & - & - & 75 \\
\bottomrule\\
\end{tabular}
}

\vspace*{2em}
\resizebox{\linewidth}{!}{
\begin{tabular}{p{0.13\linewidth}rrrrrrrrrrrr}
\toprule
\textbf{Brain Age} && BIOCARD & BLSA-1.5T & BLSA-3T & CARDIA & SHIP & SPRINT & UKBB & WHIMS & WRAP & lookAHEAD & Total \\
&& (2.12\%) & (1.05\%) & (7.82\%) & (5.97\%) & (28.54\%) & (5.54\%) & (34.79\%) & (9.40\%) & (2.09\%) & (2.68\%) \\
\cmidrule{3-13}
Subjects\\
& Control & 246 & 122 & 907 & 693 & 3311 &  643 & 4036 & 1090 & 242 & 311 & 11601 \\
Sex\\
& Female & 152 & 53 & 506 & 344 & 1695 & 227 & 2101 & 1081 & 169 & 222 & 6562 \\
& Male & 94 & 68 & 400 & 348 & 1609 & 416 & 1931 & - & 73 & 90 & 5039 \\
Age (years)\\
& Mean & 57.61 & 67.87 & 64.06 & 51.00 & 52.81 & 68.58 & 62.81 & 69.59 & 63.57 & 58.05 & 60.14 \\
& Min & 21 & 48 & 22 & 42 & 21 & 50 & 45 & 64 & 50 & 44 & 21 \\
& Max & 86 & 85 & 92 & 61 & 90 & 91 & 79 & 79 & 78 & 74 & 92 \\
Race\\
& White & 243 & 111 & 597 & 388 & - & 420 & 3915 & 1003 & 230 & 227 & 7133 \\
& Black & 2 & 11 & 229 & 304 & - & 202 & 24 & 46 & 5 & 70 & 899 \\
& Asian & - & - & 57 & - & - & 6 & 48 & 15 & 1 & - & 126 \\
\bottomrule\\
\end{tabular}
}
\end{footnotesize}
\end{center}
\end{table}

\begin{table}[htpb]
\caption{
%\newtext{
\textbf{Summary of the data from the iSTAGING consortium used for early diagnosis of stable and progressive mild cognitive impairment.}}
%}
\label{tab:mci_absolute}
\begin{center}
\begin{footnotesize}

\resizebox{0.85\linewidth}{!}{
\begin{tabular}{p{0.15\linewidth}rrrrrr}
\toprule
\textbf{Mild Cognitive} && ADNI-1 & ADNI-2/3 & PENN & AIBL & Total \\
\textbf{Impairment} && (44.60\%) & (47.69\%) & (2.31\%) & (5.40\%)  \\
\cmidrule{3-7}
% & Subjects & 289 & 309 & 15 & 35 & 648  \\
Subjects\\
& Progressive MCI & 197 & 113 & 5 & 12 & 327  \\
& Stable MCI & 79 & 151 & 10 & 11 & 251  \\
& Normal MCI & 13 & 45 & - & 12 & 70  \\
Sex\\
& Female & 103 & 138 & 7 & 13 & 261 \\
& Male & 186 & 171 & 8 & 22 & 387 \\
Age (years)\\
& 0--65 & 38 & 70 & 4 & 1 & 113 \\
& 65--70 & 38 & 73 & 3 & 9 & 123 \\
& 70--75 & 75 & 76 & 3 & 12 & 166 \\
& 75--80 & 66 & 76 & 4 & 9 & 140 \\
& > 80 & 72 & 29 & 1 & 4 & 106 \\
Race\\
& White & 277 & 283 & 14 & 14 & 589 \\
& Black & 5 & 6 & 1 & - & 12 \\
& Asian & 7 & 4 & - & - & 11 \\
\bottomrule\\
\end{tabular}
}
\end{footnotesize}
\end{center}
\end{table}

\begin{table}[htpb]
\caption{
%\newtext{
\textbf{Summary of the data from the iSTAGING consortium used for association study between the brain age residual and neuropsychological tests.}
Mini-mental state examination (MMSE) is a questionnaire test that measures global cognitive impairment. Digit span forward/backward (DSF/B) test is a way of measuring the storage capacity of a person’s working memory. Trail making test part A/B (TMT A/B) measures a person’s executive functioning. Digit symbol substitution test (DSST) is another global measure of cognitive ability, requiring multiple cognitive domains to complete effectively.}
%}
\label{tab:cog_data}
\begin{center}
\begin{footnotesize}

\resizebox{0.85\linewidth}{!}{
\begin{tabular}{p{0.15\linewidth}rrrrrrr}
\toprule
\textbf{Neuropsychological} && MMSE & DSF & TMT A & DSST & TMT B & DSB \\
\textbf{Test} &  \\
\cmidrule{3-8}
Subjects\\
& Control & 491 & 1822 & 1741 & 1547 & 1724 & 595  \\
Sex (\%)\\
& Female & 60.08 & 56.70 & 57.04 & 55.07 & 56.90 & 56.13 \\
& Male & 39.92 & 43.30 & 42.96 & 44.93 & 43.10 & 43.87 \\
Age (years)\\
& Mean & 62.05 & 61.46 & 61.48 & 58.98 & 61.43 & 60.95 \\
& Min & 22 & 22 & 22 & 43 & 22 & 22 \\
& Max & 91 & 91 & 91 & 87 & 91 & 91 \\
Race (\%)\\
& White & 70.47 & 88.42 & 88.00 & 88.30 & 87.88 & 69.92 \\
& Black & 18.94 & 6.92 & 7.29 & 10.28 & 7.37 & 19.83 \\
& Asian & 6.72 & 2.47 & 2.53 & 0.52 & 2.55 & 6.05 \\
\bottomrule\\
\end{tabular}
}
\end{footnotesize}
\end{center}
\end{table}

\begin{table}%[!t]
\begin{spacing}{1}
\caption{
%\newtext{
\textbf{Summary of missing value sample size for each variable in the data from the iSTAGING consortium (Alzheimer’s disease and brain age) and the PHENOM consortium (schizophrenia) used in this study.}}
%}
\label{tab:missing_vales}
\begin{center}
\begin{footnotesize}
%\begin{small}
\begin{tabular}{p{0.12\linewidth}rrrr}
\toprule
\textbf{Variables} &&  Alzheimer’s & Schizophrenia & Brain Age \\
\cmidrule{3-5}
MR imaging\\
& Region-of-interest volumes & 0 & 0 & 0 \\
& White matter lesion volume & 706 & - & 374 \\
Demographics\\
& Gender & 0 & 0 & 0 \\
& Age & 0 & 0 & 0 \\
& Race & 287 & 820 & 3323 \\
& Education level & - & 566 & - \\
& Marital status & - & 789 & - \\
& Employment status & - & 783 & - \\
& Handedness & - & 565 & -  \\
& Smoking status & - & - & 3694 \\
Clinical \\
& Diabetes & - & - & 2389 \\
& Hypertension & - & - & 3826 \\
& Hyperlipidemia & - & - & 5616 \\
& Blood pressure (systolic/diastolic) & - & - & 3457 \\
& Body mass index & - & - & 3465 \\
Genetic factor \\
& Apolipoprotein E allele 2 & 550 & - & 5301 \\
& Apolipoprotein E allele 3 & 550 & - & 5301 \\
& Apolipoprotein E allele 4 & 550 & - & 5301 \\
Cognitive score \\
& Mini-mental state exam & 1027 & - & - \\
\bottomrule
\end{tabular}
\end{footnotesize}
%\end{small}
\end{center}
\end{spacing}
\end{table}

\begin{figure}
\centering
\includegraphics[width=0.325\linewidth]{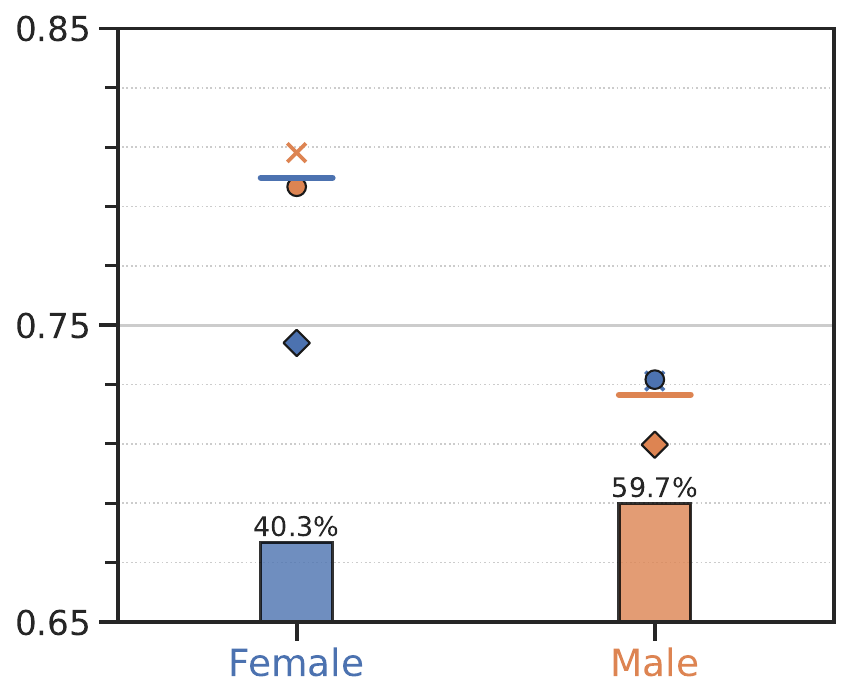}
\includegraphics[width=0.325\linewidth]{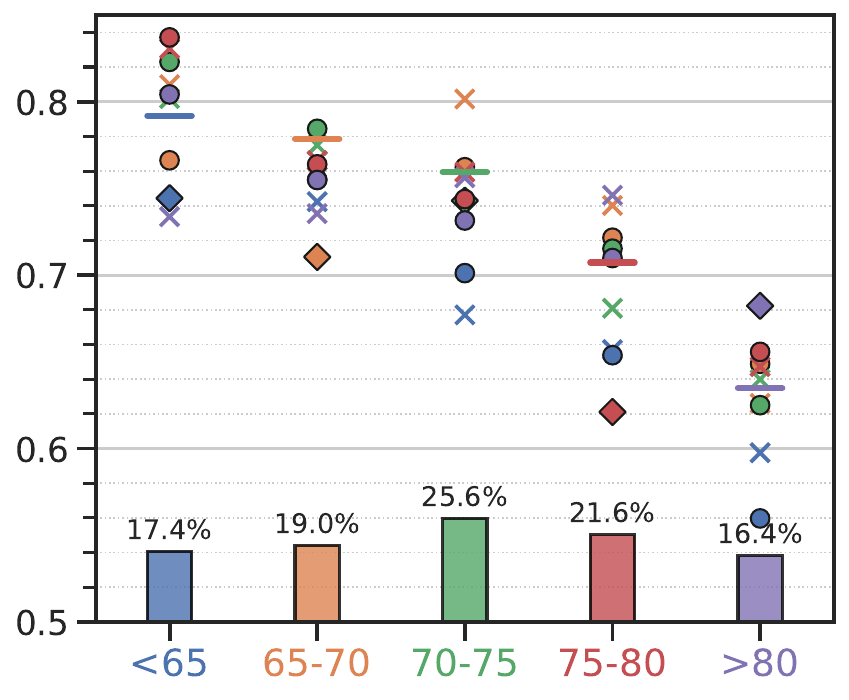}
\includegraphics[width=0.325\linewidth]{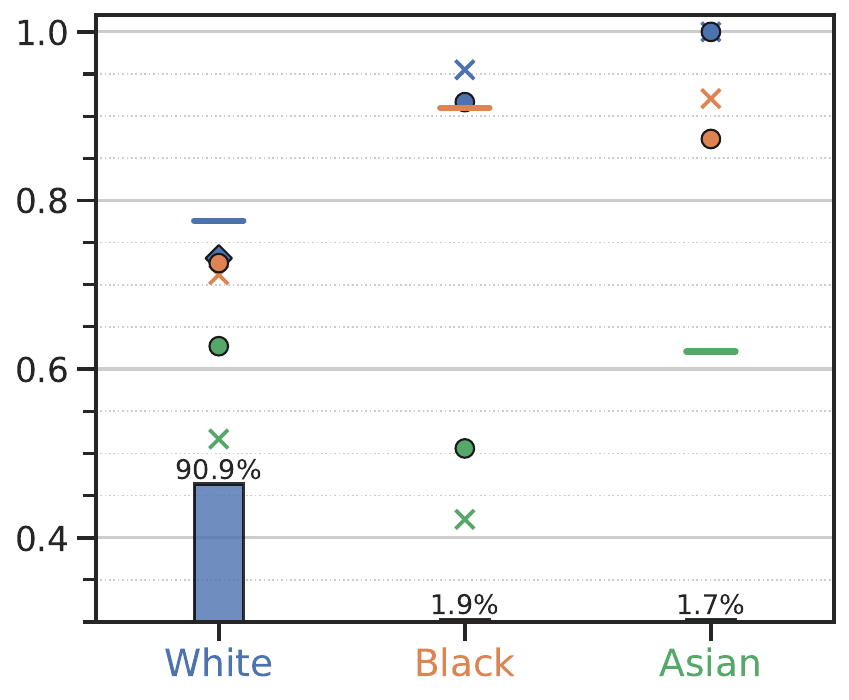}
\includegraphics[width=0.50\linewidth]{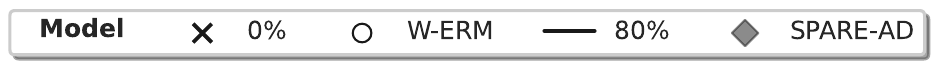}
\caption{
%\newtext{
\textbf{Early diagnosis of stable
and progressive mild cognitive impairment.}
Linear discriminant analysis on the output probabilities (that determines AD vs. cognitively normal CN) of the ensemble models trained for Alzheimer’s disease diagnosis is used to study whether subjects with mild cognitive impairment (MCI) progress to AD (known as pMCI) or remain stable MCI (known as sMCI) using only the baseline scans. The AUC of pMCI vs. sMCI on the target group is shown for three different attributes (sex, age group and race) when models are trained only on data from the source group (crosses), using $\alpha$-weighted ERM using all data from the source and 10\% data from the target group (circles) and with access to only all data from the target group (horizontal lines). We also train models using SPARE-AD index instead of AD-probabilities from the target group for comparison (diamonds).}
%}
\label{fig:mci_spare}
\end{figure}

\begin{table}
\caption{
%\newtext{
\textbf{Alzheimer’s disease classification results.}
We report the average AUC on the target group (columns in the tables) computed using five-fold nested cross-validation for models trained only on data from the target group (denoted by 80\% in the tables), only on data from the source group (denoted by 0\% in the tables), and trained on all data from the source group and 10\% data from the target group (denoted by 10\% in the tables). Panels denote groups stratified by one of the four attributes, namely sex, age group, race and clinical study. Bar plots denote the proportion of subjects in these groups in our study. All models are ensembles trained using features derived from structural measures, demographic and clinical variables, genetic factors, and cognitive scores.}
%}
\label{tab:ad_table}
\begin{center}
\begin{footnotesize}

\resizebox{0.25\linewidth}{!}{
\begin{tabular}{lrrr} % p{0.15\linewidth}
\toprule
\textbf{Sex} &  & Female &    Male \\
\cmidrule{3-4}
0\%\\
& Female &       - &  0.9426 \\
& Male   &  0.9573 &       - \\
10\% \\
& Female &       - &  0.9372 \\
& Male   &  0.9643 &       - \\
80\%\\
&   - &    0.96 &  0.9487 \\
\bottomrule
\end{tabular}
}

\vspace*{2em}
\resizebox{0.45\linewidth}{!}{
\begin{tabular}{lrrrrrr}
\toprule
\textbf{Age} & & 0-65 & 65-70 & 70-75 & 75-80 & >80 \\
\cmidrule{3-7}
0\%\\
& 0-65    &      - &    0.9226 &    0.9320 &    0.8855 & 0.8506 \\
& 65-70 & 0.9396 &         - &    0.9718 &    0.9526 & 0.9435 \\
& 70-75 & 0.9403 &    0.9485 &         - &    0.9555 & 0.9381 \\
& 75-80 & 0.9380 &    0.9428 &    0.9734 &         - & 0.9392 \\
& >80    & 0.9244 &    0.9341 &    0.9604 &    0.9377 &      - \\
10\% \\
& 0-65    &      - &    0.9373 &    0.9511 &    0.9191 & 0.9047 \\
& 65-70 & 0.9496 &         - &    0.9707 &    0.9531 & 0.9390 \\
& 70-75 & 0.9537 &    0.9626 &         - &    0.9564 & 0.9346 \\
& 75-80 & 0.9523 &    0.9485 &    0.9706 &         - & 0.9342 \\
& >80    & 0.9571 &    0.9493 &    0.9614 &    0.9488 &      - \\
80\%\\
& -         & 0.9633 &    0.9236 &    0.9638 &    0.9569 & 0.9080 \\
\bottomrule
\end{tabular}
}

\vspace*{2em}
\resizebox{0.3\linewidth}{!}{
\begin{tabular}{lrrrr}
\toprule
\textbf{Race} & &  White &  Black &  Asian \\
\cmidrule{3-5}
0\%\\
& White &      - & 0.9653 & 0.9588 \\
& Black & 0.9251 &      - & 0.9358 \\
& Asian & 0.7930 & 0.7437 &      - \\
10\%\\
& White &      - & 0.9646 & 0.9486 \\
& Black & 0.9528 &      - & 0.9257 \\
& Asian & 0.9259 & 0.8985 &      - \\
80\%\\
& -     & 0.9594 & 0.9673 & 0.8880 \\
\bottomrule
\end{tabular}
}

\vspace*{2em}
\resizebox{0.45\linewidth}{!}{
\begin{tabular}{lrrrrr}
\toprule
\textbf{Study} & & ADNI-1 & ADNI-2/3 &   PENN &   AIBL \\
\cmidrule{3-6}
0\%\\
& ADNI-1   &      - &   0.9567 & 0.9814 & 0.9561 \\
& ADNI-2/3 & 0.9496 &        - & 0.9770 & 0.9655 \\
& PENN     & 0.9255 &   0.9266 &      - & 0.8858 \\
& AIBL     & 0.8910 &   0.8832 & 0.9075 &      - \\
10\%\\
& ADNI-1   &      - &   0.9621 & 0.9871 & 0.9866 \\
& ADNI-2/3 & 0.9641 &        - & 0.9914 & 0.9811 \\
& PENN     & 0.9536 &   0.9542 &      - & 0.9928 \\
& AIBL     & 0.9454 &   0.9331 & 0.9604 &      - \\
80\%\\
& -        & 0.9748 &   0.9526 & 0.9903 & 0.9990 \\
\bottomrule
\end{tabular}
}
\end{footnotesize}
\end{center}
\end{table}

\begin{table}
\caption{
%\newtext{
\textbf{Schizophrenia classification results.}
We report the average AUC on the target group (columns in the tables) computed using five-fold nested cross-validation for models trained only on data from the target group (denoted by 80\% in the tables), only on data from the source group (denoted by 0\% in the tables), and trained on all data from the source group and 10\% data from the target group (denoted by 10\% in the tables). Panels denote groups stratified by one of the four attributes, namely sex, age group, race and clinical study. Bar plots denote the proportion of subjects in these groups in our study. All models are ensembles trained using features derived from structural measures, demographic and clinical variables, genetic factors, and cognitive scores.}
%}
\label{tab:scz_table}
\begin{center}
\begin{footnotesize}

\resizebox{0.25\linewidth}{!}{
\begin{tabular}{lrrr}
\toprule
\textbf{Sex} & & Female &   Male \\
\cmidrule{3-4}
0\%\\
& Female &      - & 0.7355 \\
& Male   & 0.7597 &      - \\
10\%\\
& Female &      - & 0.8084 \\
& Male   & 0.7896 &      - \\
80\%\\
& -      & 0.7398 & 0.8109 \\
\bottomrule
\end{tabular}
}

\vspace*{2em}
\resizebox{0.35\linewidth}{!}{
\begin{tabular}{lrrrrr}
\toprule
\textbf{Age} & & 0-25 & 25-30 & 30-35 & >35 \\
\cmidrule{3-6}
0\%\\
& 0-25    &      - &    0.7956 &    0.7523 & 0.7056 \\
& 25-30 & 0.7114 &         - &    0.8469 & 0.7453 \\
& 30-35 & 0.6571 &    0.7279 &         - & 0.7084 \\
& >35    & 0.6838 &    0.8184 &    0.8318 &      - \\
10\%\\
& 0-25    &      - &    0.8123 &    0.8237 & 0.7662 \\
& 25-30 & 0.7613 &         - &    0.8445 & 0.7469 \\
& 30-35 & 0.7739 &    0.7845 &         - & 0.7518 \\
& >35    & 0.7706 &    0.8122 &    0.8281 &      - \\
80\%\\
& -         & 0.8112 &    0.8288 &    0.8117 & 0.7649 \\
\bottomrule
\end{tabular}
}

\vspace*{2em}
\resizebox{0.25\linewidth}{!}{
\begin{tabular}{lrrr}
\toprule
\textbf{Race} & & Native &  Asian \\
\cmidrule{3-4}
0\%\\
& Native &      - & 0.6979 \\
& Asian  & 0.5686 &      - \\
10\%\\
& Native &      - & 0.7309 \\
& Asian  & 0.6316 &      - \\
80\%\\
& -      & 0.6020 & 0.6738 \\
\bottomrule
\end{tabular}
}

\vspace*{2em}
\resizebox{0.5\linewidth}{!}{
\begin{tabular}{lrrrrrr}
\toprule
\textbf{Study} & &   Penn &  China & Munich & Utrecht & Melbourne \\
\cmidrule{3-7}
0\%\\
& Penn      &      - & 0.7842 & 0.6920 &  0.6734 &    0.8296 \\
& China     & 0.6763 &      - & 0.6829 &  0.7195 &    0.8714 \\
& Munich    & 0.7279 & 0.8010 &      - &  0.7418 &    0.8146 \\
& Utrecht   & 0.7070 & 0.8016 & 0.7164 &       - &    0.8583 \\
& Melbourne & 0.6771 & 0.8196 & 0.6907 &  0.7342 &         - \\
10\%\\
& Penn      &      - & 0.8073 & 0.7082 &  0.9464 &    0.8909 \\
& China     & 0.7007 &      - & 0.7267 &  0.9679 &    0.8737 \\
& Munich    & 0.7354 & 0.8211 &      - &  0.9516 &    0.8244 \\
& Utrecht   & 0.7226 & 0.8243 & 0.7334 &       - &    0.8759 \\
& Melbourne & 0.7088 & 0.8273 & 0.7191 &  0.9548 &         - \\
80\%\\
& -         & 0.7987 & 0.8640 & 0.7444 &  0.9599 &    0.8684 \\
\bottomrule
\end{tabular}
}
\end{footnotesize}
\end{center}
\end{table}

\begin{table}
\caption{
%\newtext{
\textbf{Brain age prediction results.}
We report the average mean absolute error (MAE) in years on the target group (columns in the tables) computed using five-fold nested cross-validation for models trained only on data from the target group (denoted by 80\% in the tables), only on data from the source group (denoted by 0\% in the tables), and trained on all data from the source group and 10\% data from the target group (denoted by 10\% in the tables). Panels denote groups stratified by one of the four attributes, namely sex, race and clinical study. Bar plots denote the proportion of subjects in these groups in our study. All models are ensembles trained using features derived from structural measures, demographic and clinical variables, genetic factors, and cognitive scores.}
%}
\label{tab:age_table}
\begin{center}
\begin{footnotesize}

\resizebox{0.25\linewidth}{!}{
\begin{tabular}{lrrr}
\toprule
\textbf{Sex} & & Female &   Male \\
\cmidrule{3-4}
0\%\\
& Female &      - & 4.6902 \\
& Male   & 4.6643 &      - \\
10\%\\
& Female &      - & 4.6166 \\
& Male   & 4.3896 &      - \\
80\%\\
& -      & 4.1181 & 4.3549 \\
\bottomrule
\end{tabular}
}

\vspace*{2em}
\resizebox{0.3\linewidth}{!}{
\begin{tabular}{lrrrr}
\toprule
\textbf{Race} & &  White &  Black &  Asian \\
\cmidrule{3-5}
0\%\\
& White &      - & 4.8421 & 4.7436 \\
& Black & 6.1148 &      - & 6.2676 \\
& Asian & 6.2358 & 6.8857 &      - \\
10\%\\
& White &      - & 4.8474 & 4.5357 \\
& Black & 4.5049 &      - & 6.0057 \\
& Asian & 4.5894 & 5.9918 &      - \\
80\%\\
& -     & 3.9901 & 4.8967 & 5.1489 \\
\bottomrule
\end{tabular}
}

\vspace*{2em}
\resizebox{1.\linewidth}{!}{
\begin{tabular}{lrrrrrrrrrrr}
\toprule
\textbf{Study} & & BIOCARD & BLSA-1.5T & BLSA-3T &  CARDIA &    SHIP &  SPRINT & UKBB &   WHIMS &    WRAP & lookAHEAD \\
\cmidrule{3-12}
0\%\\
& BIOCARD   &       - &    5.1597 &  8.1243 &  5.1326 &  8.8638 &  5.0915 &    4.9134 &  3.4510 &  4.2352 &    6.4431 \\
& BLSA-1.5T &  7.0434 &         - &  7.9681 &  9.1985 & 11.7253 &  5.3157 &    4.9993 &  3.1509 &  4.4632 &    9.0969 \\
& BLSA-3T   &  7.5188 &    5.4739 &       - &  9.7144 & 11.1708 &  6.0576 &    5.6380 &  6.5140 &  4.8916 &    9.2493 \\
& CARDIA    &  8.0467 &   14.1256 & 15.3413 &       - & 10.3559 & 12.8710 &   10.2514 & 15.9061 & 10.4533 &    5.2601 \\
& SHIP      & 10.3542 &   15.3256 & 13.8427 &  6.5199 &       - &  7.7349 &   14.1402 &  5.0092 &  8.9724 &    5.8628 \\
& SPRINT    &  7.5668 &    6.9445 &  9.8383 &  8.5229 & 12.4118 &       - &    5.8606 &  5.6735 &  4.9719 &    7.8257 \\
& UKBB &  6.1263 &    5.2981 &  7.5216 &  7.0385 & 11.0483 &  5.3614 &         - &  3.1166 &  4.2923 &    8.1095 \\
& WHIMS     & 10.3377 &    6.0249 & 10.7618 & 13.6363 & 15.3092 &  6.4792 &    6.1463 &       - &  5.1581 &    7.0639 \\
& WRAP      &  7.9669 &    5.9605 &  9.4942 & 10.1443 & 12.9613 &  5.6444 &    5.0151 &  2.8126 &       - &    7.6354 \\
& lookAHEAD &  7.5399 &   12.3101 & 13.3466 &  4.2895 &  9.5694 & 11.3139 &    8.7010 &  9.1397 &  8.4743 &         - \\
10\%\\
& BIOCARD   &       - &    4.7087 &  6.4790 &  3.0771 &  5.6160 &  4.9782 &    4.4169 &  2.7704 &  4.0595 &    4.7564 \\
& BLSA-1.5T &  5.9926 &         - &  6.4443 &  3.4137 &  5.7878 &  5.0184 &    4.4336 &  2.7778 &  4.3133 &    4.7999 \\
& BLSA-3T   &  5.6211 &    4.7634 &       - &  3.5383 &  5.7111 &  4.6509 &    4.2806 &  2.8641 &  4.1537 &    4.5702 \\
& CARDIA    &  6.7340 &    6.6454 &  7.4902 &       - &  5.7652 &  5.2600 &    4.6932 &  2.8337 &  4.7503 &    4.4487 \\
& SHIP      &  5.7043 &    5.5634 &  6.5381 &  2.9623 &       - &  4.7351 &    4.4113 &  2.7875 &  4.2589 &    4.2730 \\
& SPRINT    &  5.9473 &    4.8374 &  6.4243 &  3.0688 &  5.5828 &       - &    4.2567 &  2.8009 &  4.1908 &    4.7320 \\
& UKBB &  5.3467 &    4.9048 &  5.9838 &  3.7585 &  5.6768 &  4.6116 &         - &  2.7193 &  3.7259 &    4.6744 \\
& WHIMS     &  6.4178 &    5.7135 &  6.7564 &  3.1224 &  5.7796 &  5.0562 &    4.4045 &       - &  4.5165 &    4.6987 \\
& WRAP      &  6.1384 &    5.5298 &  6.5214 &  3.1077 &  5.6488 &  5.1082 &    4.3612 &  2.7497 &       - &    4.6183 \\
& lookAHEAD &  6.6080 &    6.3420 &  7.1366 &  3.0568 &  5.6129 &  5.3508 &    4.4653 &  2.7729 &  4.6745 &         - \\
80\%\\
& -         &  5.4879 &    4.2219 &  5.0418 &  2.7186 &  4.6912 &  4.3610 &    3.8118 &  2.6655 &  3.7032 &    3.9595 \\
\bottomrule
\end{tabular}
}
\end{footnotesize}
\end{center}
\end{table}

\begin{table}
\caption{
%\newtext{
\textbf{Mild cognitive impairment progression prediction results.}
Linear discriminant analysis on the output probabilities (that determines AD vs. cognitively normal CN) of the ensemble models trained for Alzheimer’s disease diagnosis is used to study whether subjects with mild cognitive impairment (MCI) progress to AD (known as pMCI) or remain stable MCI (known as sMCI) using only the baseline scans. The average AUC of pMCI vs. sMCI on the target group (columns in the tables) is shown for three different attributes (sex, age group and race) when models are trained only on data from the source group (denoted by 0\% in the tables), using $\alpha$-weighted ERM using all data from the source and 10\% data from the target group (denoted by 10\% in the tables) and with access to only all data from the target group (denoted by 80\% in the tables).}
%}
\label{tab:mci_table}
\begin{center}
\begin{footnotesize}

\resizebox{0.25\linewidth}{!}{
\begin{tabular}{lrrr}
\toprule
\textbf{Sex} & & Female &   Male \\
\cmidrule{3-4}
0\%\\
& Female &      - & 0.7312 \\
& Male   & 0.8082 &      - \\
10\%\\
& Female &      - & 0.7317 \\
& Male   & 0.7967 &      - \\
80\%\\
& -      & 0.7997 & 0.7265 \\
\bottomrule
\end{tabular}
}

\vspace*{2em}
\resizebox{0.45\linewidth}{!}{
\begin{tabular}{lrrrrrr}
\toprule
\textbf{Age} & & 0-65 & 65-70 & 70-75 & 75-80 & >80 \\
\cmidrule{3-7}
0\%\\
& 0-65    &      - &    0.7423 &    0.6771 &    0.6570 & 0.5976 \\
& 65-70 & 0.8098 &         - &    0.8016 &    0.7402 & 0.6260 \\
& 70-75 & 0.8018 &    0.7749 &         - &    0.6809 & 0.6399 \\
& 75-80 & 0.8306 &    0.7661 &    0.7594 &         - & 0.6473 \\
& >80    & 0.7337 &    0.7355 &    0.7562 &    0.7460 &      - \\
10\%\\
& 0-65    &      - &    0.7551 &    0.7012 &    0.6538 & 0.5597 \\
& 65-70 & 0.7662 &         - &    0.7622 &    0.7216 & 0.6489 \\
& 70-75 & 0.8230 &    0.7844 &         - &    0.7152 & 0.6250 \\
& 75-80 & 0.8371 &    0.7639 &    0.7438 &         - & 0.6558 \\
& >80    & 0.8042 &    0.7549 &    0.7315 &    0.7099 &      - \\
80\%\\
& -         & 0.7919 &    0.7786 &    0.7594 &    0.7073 & 0.6348 \\
\bottomrule
\end{tabular}
}

\vspace*{2em}
\resizebox{0.35\linewidth}{!}{
\begin{tabular}{lrrrr}
\toprule
\textbf{Race} & &  White &  Black &  Asian \\
\cmidrule{3-5}
0\%\\
& White &      - & 0.9550 & 1.0000 \\
& Black & 0.7118 &      - & 0.9208 \\
& Asian & 0.5171 & 0.4217 &      - \\
10\%\\
& White &      - & 0.9167 & 1.0000 \\
& Black & 0.7257 &      - & 0.8729 \\
& Asian & 0.6271 & 0.5058 &      - \\
80\%\\
& -     & 0.7759 & 0.9100 & 0.6208 \\
\bottomrule
\end{tabular}
}
\end{footnotesize}
\end{center}
\end{table}

\begin{table}
\caption{
%\newtext{
\textbf{Pearson’s correlation between the brain age residual and neuropsychological tests.}
For two different attributes (sex and race), the models are trained only on source data (denoted by 0\% in the tables), using $\alpha$-weighted ERM on all source data and 10\% target data (denoted by 10\% in the tables) and only on all target data (denoted by 80\% in the tables). Mini-mental state examination (MMSE) is a questionnaire test that measures global cognitive impairment. Digit span forward (DSF) test is a way of measuring the storage capacity of a person’s working memory. Trail making test part A (TMT A) measures a person’s executive functioning.}
%}
\label{tab:cog_table_1}
\begin{center}
\begin{footnotesize}

\text{(a) Mini-mental state examination (MMSE).}

\vspace*{0.5em}
\resizebox{0.32\linewidth}{!}{
\begin{tabular}{lrrr}
\toprule
\textbf{Sex} & & Female &    Male \\
\cmidrule{3-4}
0\%\\
& Female &       - & -0.1197 \\
& Male   & -0.2865 &       - \\
10\%\\
& Female &       - & -0.1458 \\
& Male   & -0.2674 &       - \\
80\%\\
& -      & -0.1614 & -0.0489 \\
\bottomrule
\end{tabular}
}
\quad
\resizebox{0.4\linewidth}{!}{
\begin{tabular}{lrrrr}
\toprule
\textbf{Race} & &  White &   Black &   Asian \\
\cmidrule{3-5}
0\%\\
& White &       - & -0.0829 & -0.4625 \\
& Black & -0.3033 &       - & -0.4917 \\
& Asian & -0.2434 & -0.1232 &       - \\
10\%\\
& White &       - & -0.0777 & -0.4097 \\
& Black & -0.2974 &       - & -0.4609 \\
& Asian & -0.2698 & -0.1508 &       - \\
80\%\\
& -     & -0.1447 & -0.0714 & -0.1495 \\
\bottomrule
\end{tabular}
}

\vspace*{2em}
\text{(b) Digit span forward (DSF).}

\vspace*{0.5em}
\resizebox{0.32\linewidth}{!}{
\begin{tabular}{lrrr}
\toprule
\textbf{Sex} & & Female &    Male \\
\cmidrule{3-4}
0\%\\
& Female &       - & -0.0032 \\
& Male   & -0.0865 &       - \\
10\%\\
& Female &       - & -0.0232 \\
& Male   & -0.0815 &       - \\
80\%\\
& -      &  0.0178 & -0.0379 \\
\bottomrule
\end{tabular}
}
\quad
\resizebox{0.4\linewidth}{!}{
\begin{tabular}{lrrrr}
\toprule
\textbf{Race} & &  White &   Black &   Asian \\
\cmidrule{3-5}
0\%\\
& White &       - &  0.0783 & -0.3364 \\
& Black & -0.1708 &       - & -0.3114 \\
& Asian & -0.0113 & -0.0750 &       - \\
10\%\\
& White &       - &  0.0894 & -0.3134 \\
& Black & -0.1541 &       - & -0.2931 \\
& Asian & -0.0256 & -0.1149 &       - \\
80\%\\
& -     & -0.0292 & -0.0780 & -0.1136 \\
\bottomrule
\end{tabular}
}

\vspace*{2em}
\text{(c) Trail making test part A (TMT A).}

\vspace*{0.5em}
\resizebox{0.32\linewidth}{!}{
\begin{tabular}{lrrr}
\toprule
\textbf{Sex} & & Female &   Male \\
\cmidrule{3-4}
0\%\\
& Female &      - & 0.1960 \\
& Male   & 0.1835 &      - \\
10\%\\
& Female &      - & 0.1940 \\
& Male   & 0.1578 &      - \\
80\%\\
& -      & 0.1572 & 0.0476 \\
\bottomrule
\end{tabular}
}
\quad
\resizebox{0.4\linewidth}{!}{
\begin{tabular}{lrrrr}
\toprule
\textbf{Race} & &  White &   Black &   Asian \\
\cmidrule{3-5}
0\%\\
& White &      - & -0.0026 & 0.4336 \\
& Black & 0.1740 &       - & 0.5911 \\
& Asian & 0.1279 &  0.0679 &      - \\
10\%\\
& White &      - & -0.0025 & 0.4019 \\
& Black & 0.2101 &       - & 0.5814 \\
& Asian & 0.1906 &  0.0825 &      - \\
80\%\\
& -     & 0.1157 &  0.2251 & 0.3113 \\
\bottomrule
\end{tabular}
}
\end{footnotesize}
\end{center}
\end{table}

\begin{table}
\caption{
%\newtext{
\textbf{Pearson’s correlation between the brain age residual and neuropsychological tests.}
For two different attributes (sex and race), the models are trained only on source data (denoted by 0\% in the tables), using $\alpha$-weighted ERM on all source data and 10\% target data (denoted by 10\% in the tables) and only on all target data (denoted by 80\% in the tables). Digit symbol substitution test (DSST) is another global measure of cognitive ability, requiring multiple cognitive domains to complete effectively. Trail making test part B (TMT B) measures a person’s executive functioning. Digit span backward (DSB) test is a way of measuring the storage capacity of a person’s working memory.}
%}
\label{tab:cog_table_2}
\begin{center}
\begin{footnotesize}

\text{(a) Digit symbol substitution test (DSST).}

\vspace*{0.5em}
\resizebox{0.32\linewidth}{!}{
\begin{tabular}{lrrr}
\toprule
\textbf{Sex} & &  Female &    Male \\
\cmidrule{3-4}
0\%\\
& Female &       - & -0.3963 \\
& Male   & -0.2232 &       - \\
10\%\\
& Female &       - & -0.4017 \\
& Male   & -0.2001 &       - \\
80\%\\
& -      & -0.1342 & -0.1540 \\
\bottomrule
\end{tabular}
}
\quad
\resizebox{0.4\linewidth}{!}{
\begin{tabular}{lrrrr}
\toprule
\textbf{Race} & &  White &   Black &   Asian \\
\cmidrule{3-5}
0\%\\
& White &       - & -0.0248 & -0.1814 \\
& Black & -0.4085 &       - & -0.7416 \\
& Asian & -0.4533 & -0.1100 &       - \\
10\%\\
& White &       - & -0.0227 & -0.2029 \\
& Black & -0.4266 &       - & -0.5764 \\
& Asian & -0.5560 & -0.1144 &       - \\
80\%\\
& -     & -0.1191 & -0.1980 & -0.5151 \\
\bottomrule
\end{tabular}
}

\vspace*{2em}
\text{(b) Trail making test part B (TMT B).}

\vspace*{0.5em}
\resizebox{0.32\linewidth}{!}{
\begin{tabular}{lrrr}
\toprule
\textbf{Sex} & & Female &   Male \\
\cmidrule{3-4}
0\%\\
& Female &      - & 0.1979 \\
& Male   & 0.2385 &      - \\
10\%\\
& Female &      - & 0.1939 \\
& Male   & 0.2087 &      - \\
80\%\\
& -      & 0.1908 & 0.0639 \\
\bottomrule
\end{tabular}
}
\quad
\resizebox{0.4\linewidth}{!}{
\begin{tabular}{lrrrr}
\toprule
\textbf{Race} & &  White &   Black &   Asian \\
\cmidrule{3-5}
0\%\\
& White &      - & 0.0296 & 0.3429 \\
& Black & 0.2377 &      - & 0.5779 \\
& Asian & 0.1324 & 0.1402 &      - \\
10\%\\
& White &      - & 0.0279 & 0.3293 \\
& Black & 0.2808 &      - & 0.5557 \\
& Asian & 0.2122 & 0.1610 &      - \\
80\%\\
& -     & 0.1191 & 0.1861 & 0.4670 \\
\bottomrule
\end{tabular}
}

\vspace*{2em}
\text{(c) Digit span backward (DSB).}

\vspace*{0.5em}
\resizebox{0.32\linewidth}{!}{
\begin{tabular}{lrrr}
\toprule
\textbf{Sex} & &  Female &    Male \\
\cmidrule{3-4}
0\%\\
& Female &       - & -0.0693 \\
& Male   & -0.0958 &       - \\
10\%\\
& Female &       - & -0.0898 \\
& Male   & -0.0710 &       - \\
80\%\\
& -      &  0.0239 &  0.0148 \\
\bottomrule
\end{tabular}
}
\quad
\resizebox{0.4\linewidth}{!}{
\begin{tabular}{lrrrr}
\toprule
\textbf{Race} & &  White &   Black &   Asian \\
\cmidrule{3-5}
0\%\\
& White &       - & 0.1262 & -0.2682 \\
& Black & -0.1285 &      - & -0.2715 \\
& Asian & -0.0612 & 0.0972 &       - \\
10\%\\
& White &       - & 0.1339 & -0.2306 \\
& Black & -0.1417 &      - & -0.2376 \\
& Asian & -0.0857 & 0.0506 &       - \\
80\%\\
& -     & -0.0349 & 0.0591 &  0.0491 \\
\bottomrule
\end{tabular}
}
\end{footnotesize}
\end{center}
\end{table}

\end{appendix}

\end{document}